\documentclass[journal,onecolumn]{IEEEtran}
\usepackage{acronym}
\usepackage{amssymb}
\usepackage[numbers,sort&compress]{natbib}
\usepackage{textcomp}
\usepackage{graphics}
\usepackage{epstopdf}
\usepackage{float}
\usepackage{color}
\usepackage[cmex10]{amsmath}
\usepackage{latexsym,amsfonts}
\usepackage{url}
\usepackage{longtable}
\usepackage[figuresright]{rotating}
\usepackage{listings}
\usepackage{etoolbox}
\usepackage{bm}
\usepackage{bbm}
\usepackage{euscript}
\usepackage{mathtools}
\usepackage{subfigure}
\usepackage{enumerate}
\usepackage{tikz}
\usepackage{relsize}
\usepackage{easyReview}
\mathtoolsset{showonlyrefs,showmanualtags}

\begin{document}

\title{Asymptotic Analysis of an Ensemble of Randomly Projected Linear Discriminants}

\author{Lama~B.~Niyazi,
        Abla~Kammoun,
        Hayssam~Dahrouj,
        Mohamed-Slim~Alouini,
        and Tareq~Y.~Al-Naffouri
        
\thanks{L. B. Niyazi, A. Kammoun, M.-S. Alouni, and T. Y. Al-Naffouri  are with the Electrical Engineering Program, King Abdullah University of Science and Technology, Thuwal, Saudi Arabia; emails: \{lama.niyazi, abla.kammoun, slim.alouini, tareq.alnaffouri\}@kaust.edu.sa}
\thanks{H. Dahrouj is with the Department of Electrical and Computer Engineering, Effat University, Jeddah, Saudi Arabia; email: hayssam.dahrouj@gmail.com}}
\maketitle

\begin{abstract}
Datasets from the fields of bioinformatics, chemometrics, and face recognition are typically characterized by small samples of high-dimensional data. Among the many variants of linear discriminant analysis that have been proposed in order to rectify the issues associated with classification in such a setting, the classifier in \cite{durrant2013random}, composed of an ensemble of randomly projected linear discriminants, seems especially promising; it is  computationally efficient and, with the optimal projection dimension parameter setting, is competitive with the state-of-the-art. In this work, we seek to further understand the behavior of this classifier through asymptotic analysis. Under the assumption of a growth regime in which the dataset and projection dimensions grow at constant rates to each other, we use random matrix theory to derive asymptotic misclassification probabilities showing the effect of the ensemble as a regularization of the data sample covariance matrix. The asymptotic errors further help to identify situations in which the ensemble offers a performance advantage. We also develop a consistent estimator of the misclassification probability as an alternative to the computationally-costly cross-validation estimator, which is conventionally used for parameter tuning. Finally, we demonstrate the use of our estimator for tuning the projection dimension on both real and synthetic data.
\end{abstract}


%
\IEEEpeerreviewmaketitle

\section{Introduction}
\IEEEPARstart{L}{}inear Discriminant Analysis (LDA) is a classical method, which under relatively strong assumptions, is equivalent to the Bayes optimal classifier. In spite of these assumptions, it has been shown to perform robustly on a variety of datasets \cite{lim2000comparison}. Consequently, LDA and its variants are a popular choice for classification in many applications including chemometrics \cite{giansante2003classification,skrobot2007use,azcarate2017chemometric,melucci2019checking}, face recognition \cite{zhao2003face,lu2003face,kramer2018understanding,portillo2018view}, and cancer identification through gene expression microarray data \cite{sharma2008cancer,paliwal2010improved,huerta2010hybrid,sharma2012filter,li2018fisher}. Such applications, however, often suffer from high-dimensionality of data and a relative scarcity of samples. For example, it is common for microarray datasets to contain less than $100$ samples of $5,000$ to $10,000$ features each \cite{dudoit2002comparison}. This gives rise to what is called the `small sample issue'.

For an LDA classifier, the constraint of a small sample is particularly problematic as the LDA discriminant depends on the inversion of a now singular covariance estimate. Many variants of LDA are designed to overcome this. Popular approaches involve utilizing the pseudoinverse to invert the covariance (with poor performance), restricting the covariance structure to a diagonal matrix \cite{mai2013review}, or making use of a ridge estimate of the covariance, otherwise known as Regularized Linear Discriminant Analysis (RLDA) \cite{RAUDYS1998385}. Other strategies reduce the dimensionality of the data so that the covariance estimate is no longer singular through a preliminary stage of feature selection or dimensionality reduction of the data. The Fisherface technique is a popular example of the latter that is widely used in face recognition. Here, the data is preprocessed by Principal Components Analysis (PCA) before applying LDA \cite{Sharma2015}. Besides PCA, other methods of dimensionality reduction include Singular Value Decomposition (SVD), the Discrete Cosine Transform (DCT) (for images), and random projection, a technique which projects the data onto a randomly selected lower-dimensional subspace. In their work on random projection in dimensionality reduction, the authors of \cite{Bingham:2001:RPD:502512.502546} studied the effects of these techniques on image and text data and observed that random projection introduces relatively little distortion in comparison. This finding corroborates existing theory; the Johnson-Lindenstrauss lemma states that, with high probability, the distances between points in a vector space versus the points projected onto a randomly selected subspace of sufficiently high dimension are preserved. At the same time, random projection is significantly more efficient than traditional methods of dimensionality reduction. In applications where data is high-dimensional and computational efficiency is of concern, it seems a promising approach.

\begin{figure}[h]
    \begin{center}
        \includegraphics[width=0.5\linewidth]{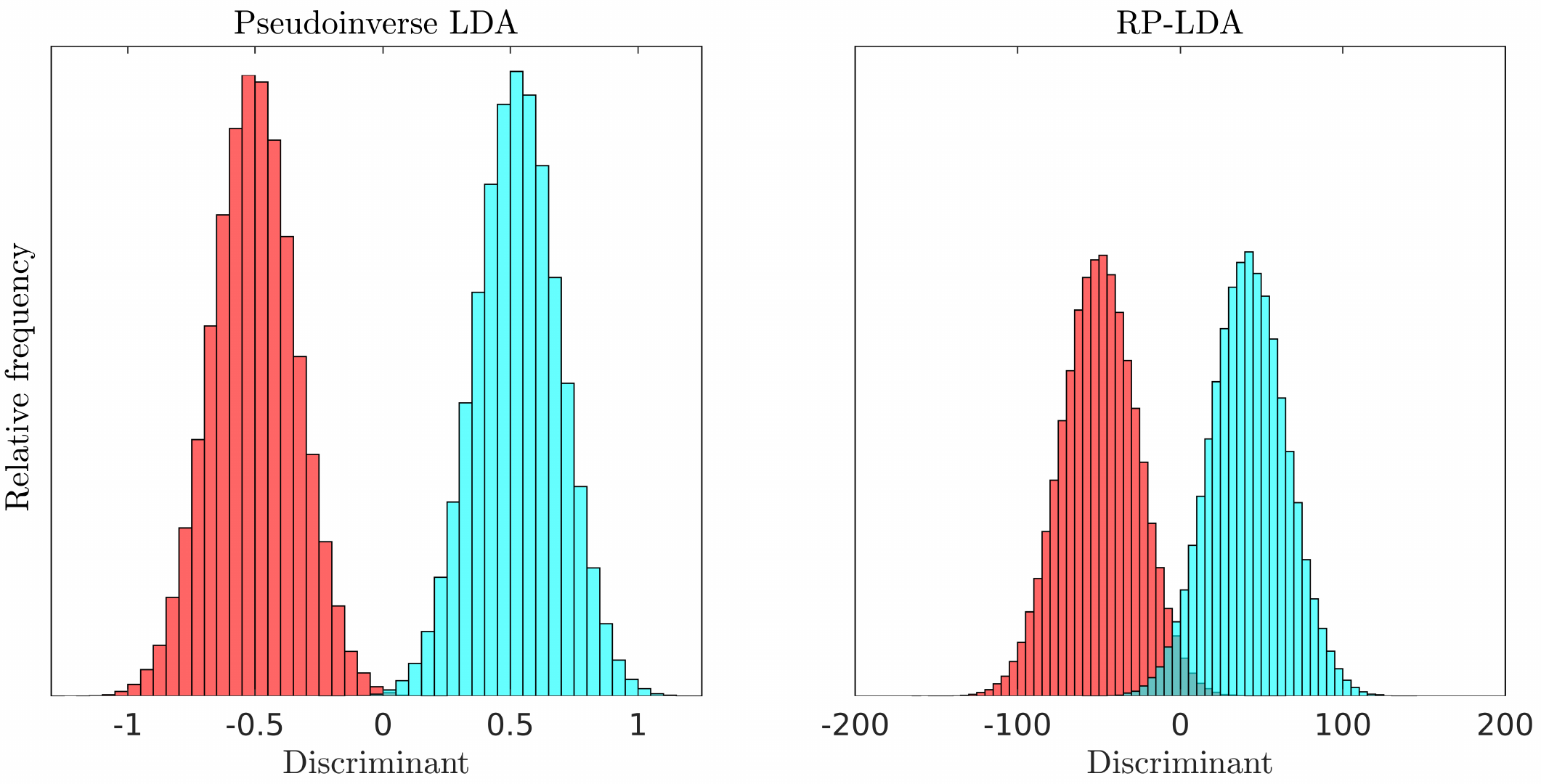}
    \end{center}
     \caption{Estimated class-conditional discriminant distributions for the pseudoinverse LDA and RP-LDA classifiers}
    \label{fig1}
\end{figure}

Our interest in random projection lies in overcoming the small sample issue encountered with the LDA classifier. The idea of an LDA classifier operating in a randomly projected subspace was first proposed in \cite{5597392}. We will refer to this scheme as the \textit{RP-LDA classifier}. It is often the case that a single random projection destroys discriminating class structure contained within the original data space \cite{cannings2017random}. As a result, a single random projection with LDA as the base classifier performs worse than LDA alone, even in a small sample regime for which LDA is pseudoinverse-modified. Figure \ref{fig1} demonstrates this effect. It plots estimated probability distributions of the class-conditional discriminants (relative frequency of many realizations of testing points of a given class for each class) of each of these classifiers in a two-class scenario. Although RP-LDA increases the mean-separation of the distributions, the accompanying increase in variance offsets this gain and results in a greater overlap between the distributions than occurs with pseudoinverse LDA. This prompts us to look into classifier ensembles comprised of multiple projections.

\begin{figure}[h]
    \begin{center}
        \includegraphics[width=0.7\linewidth]{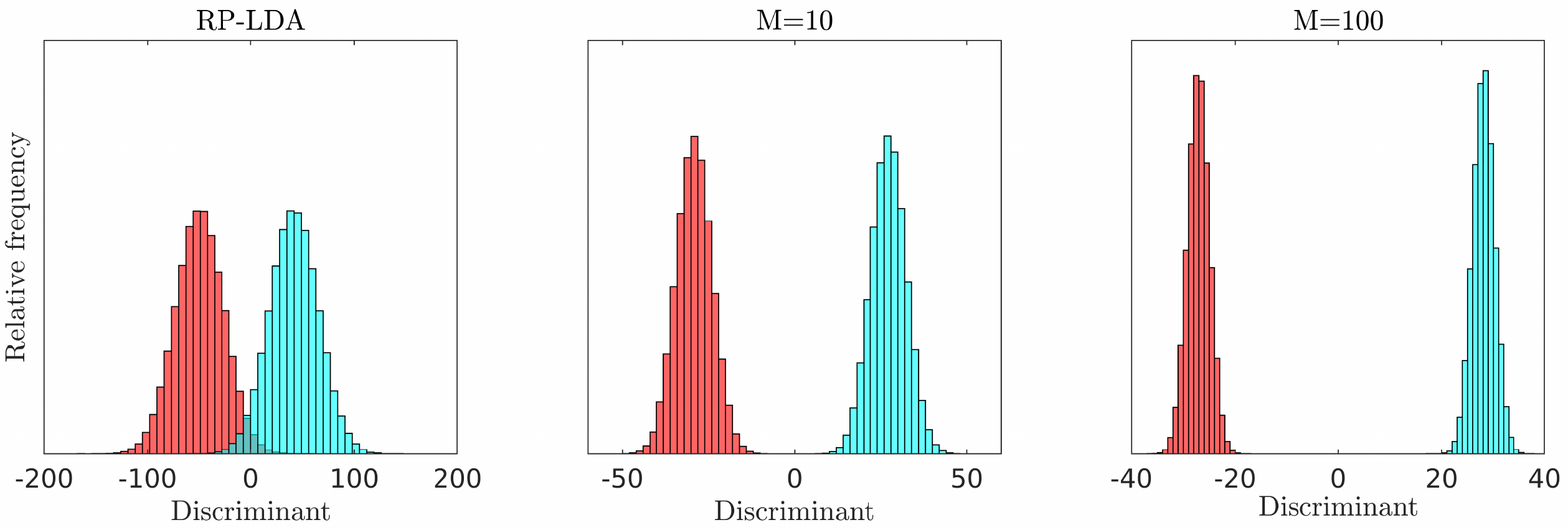}
    \end{center}
     \caption{Estimated class-conditional discriminant distributions for the RP-LDA, RP-LDA ensemble with $M=10$, and RP-LDA ensemble with $M=100$}
    \label{fig2}
\end{figure}

An ensemble of classifiers consists of multiple classifiers combined in some way as to reach a net decision rule. The literature has shown, both empirically and theoretically, that such strategies achieve better accuracy than single models \cite{Sammut:2017:EML:3153490}. The authors of \cite{durrant2013random} proposed an RP-LDA ensemble classifier in which $M$ individual RP-LDA classifiers each learned using a different random projection are combined by averaging their discriminants. The decision is then made based on this aggregated discriminant. The behavior of this classifier as $M$ increases is demonstrated in Figure \ref{fig2}. While the estimated distributions of the class-conditional discriminants for RP-LDA ($M=1$), RP-LDA ensemble with $M=10$, and RP-LDA ensemble with $M=100$ show a decrease in the mean separation with increasing $M$, Figure \ref{fig2} shows that this is accompanied by an overwhelming decrease in variance of the class-conditioned distributions. The net result is less overlap between the distributions with increasing $M$, suggesting better classification.

 In addition to demonstrating the RP-LDA ensemble's comparable performance to the state-of-the-art Support Vector Machine (SVM) for large enough $M$, the authors of \cite{durrant2013random} show that the condition number of the projected covariance estimate is bounded, and that the bound is a function of the projection dimension, that is, the projection dimension acts as a regularization parameter. They also empirically show that the classifier's accuracy is sensitive to the selection of projected dimension. Consequently, proper tuning of the projection dimension is important for ensuring good performance of the RP-LDA ensemble. Based on experimentation with various microarray datasets, \cite{durrant2013random} suggest a rule of thumb: set the projection dimension to about half of the the rank of the covariance estimate of the original unprojected data. Seeing as the RP-LDA ensemble classifier of \cite{durrant2013random} performs so well at relatively low complexity, it is worthwhile to develop a more refined and general method of tuning this parameter, rather than relying on empirical observations on a few datasets. More generally, it is worth studying the performance of the RP-LDA ensemble as a learning algorithm. That is what we attempt to do in this work through the asymptotic analysis of the classifier's misclassification probability. 

A distinction should be made between a trained classifier's expected misclassification rate on unseen data known as the \textit{prediction error (PE)} and the expected error of a learning algorithm, which is averaged over both training sets and testing examples and is referred to as the \textit{expected prediction error (EPE)} or \textit{generalization error} \cite{bengio2004no}. In application-based problems where a particular dataset is involved, PE is usually the metric of choice, whereas in algorithm evaluation, a measure of general performance irrespective of any particular dataset is needed, and EPE is the metric of choice \cite{bengio2004no}. In the absence of an exact knowledge of training set distributions, both of these quantities can only be estimated from data. The quality of the estimates are not only important for their reliability as indicators of classifier/algorithm performance, but also to be able to tune classifier parameters, which are selected on the basis of minimizing PE. Obtaining reliable estimators of these quantities is therefore an essential part of classifier design.

In practice, a part of the data is partitioned for testing from which an unbiased hold-out estimate of the classifier's PE may be computed \cite{dougherty2001small}. If several independent training and test sets are available, a hold-out estimate of the EPE of the classifier is also possible \cite{bengio2004no}. However, error estimation becomes problematic when data is scarce. This is especially true of high dimensional data, as the curse of dimensionality necessitates an exponential increase in samples with increasing dimensions of the data  for proper estimation \cite{domingos2012few}. Lacking this sheer quantity of samples, small sample approaches are resorted to, in which the training data is recycled for error estimation. The most popular of these approaches is cross-validation. Although cross-validation yields an unbiased estimate of the EPE, it has its disadvantages, mainly high variance of the resulting estimator and high computational cost \cite{bengio2004no}.

The main contribution of this paper is to derive a single consistent estimator of both the RP-LDA ensemble PE and EPE. The estimator is constructed so that it is consistent in high-dimensional data settings through the assumption of appropriate growth conditions. In this work, we assume a growth regime where the number of training data samples $n$ and the data dimensionality $p$, as well as the projection dimension $d$, grow at constant rates to each other, in contrast to the classical regime where $n$ grows to infinity while $p$ is fixed, which does not represent the finite scenario in this case. Random matrix theory facilitates the derivation of limits of expressions involving random matrices whose dimensions are subject to this regime. This yields a \textit{generalized estimator} (G-estimator) of the target quantity. The first application of random matrix theory to the family of discriminant analysis classifiers appears in \cite{zollanvari2015generalized} in which the authors construct a G-estimator of LDA error. They show that it performs favorably compared to traditional estimators such as bolstered resubstitution, bootstrap, and cross-validation. Following this work, the approach has been applied systematically to a number of classifiers. For example, reference \cite{elkhalil2017large} considers the asymptotic analysis of the RLDA and Regularized Quadratic Discriminant Analysis (RQDA) classifiers. Similarly, reference \cite{yang2018regularized} studies the Regularized Discriminant Analysis (RDA) classifier. A somewhat related work to the current one is \cite{8683386} which considers the asymptotic analysis of a singly projected LDA classifier which has knowledge of the true statistics of the data. For the analysis pursued in the current paper, the first step is to derive a closed form expression of the RP-LDA ensemble error. This is a function of random matrices whose dimensions are growing in the assumed growth regime. Through manipulation by random matrix theory tools, a corresponding G-estimator is obtained. After constructing the G-estimator, we demonstrate its use in tuning the projection dimension of the RP-LDA ensemble on specific datasets: synthetic data, generated so that the optimal Bayes classifier assumptions hold, as well as three real datasets including a microarray dataset.

Besides constructing the G-estimator, we use random matrix theory to derive the asymptotic misclassification probability of the RP-LDA ensemble classifier. Although this quantity is a prerequisite to constructing the G-estimator, it is useful in its own right. We first present the asymptotic error with respect to the random projection ensemble. This gives an understanding of how the ensemble affects the classification error as compared to classical LDA. From this, we observe that, asymptotically, the RP-LDA ensemble is a special case of RLDA where the regularization parameter is a function of the projection dimension. From this, we deduce that an RP-LDA ensemble classifier cannot outperform an RLDA classifier for which the regularization parameter has been properly tuned. We also derive the asymptotic errors under various combinations of known and unknown class-conditional data distribution statistics. Under certain assumptions, these quantities have explicit forms which we compare with existing results for LDA and RP-LDA. This yields further insights into the ensemble behavior and when and how it can outperform LDA.

To summarize, the main contributions of this paper are
\begin{itemize}
    \item An asymptotic characterization of the effect of the randomly projected ensemble, which shows that it is a special case of RLDA.
    \item The derivation of asymptotic probabilities of misclassification under various combinations of knowledge of the  class-conditional data distribution statistics and comparison, under certain assumptions, with the corresponding expressions for LDA.
    \item The construction of a G-estimator of the misclassification probability of the RP-LDA ensemble operating under unknown  class-conditional data distribution  statistics.
    \item A demonstration of the efficacy of the G-estimator for tuning of the RP-LDA ensemble projection dimension, on both synthetic and real data.
\end{itemize}

The structure of the rest of this paper is as follows: We present the classification setting, assumptions on the data, and the RP-LDA ensemble classifier decision rule in Section \ref{section_classifier}. In Section \ref{section_analysis}, we derive the error expression and define the growth regime for the asymptotic analysis. We then present the asymptotic misclassification probabilities and G-estimator in succession. In Section \ref{section_sims}, we demonstrate the tuning of the classifier projection dimensions using the G-estimator on a number of datasets, both synthetic and real. Finally, we conclude this paper in Section \ref{conclusion} with a summary of the findings.

Throughout the paper, scalars are denoted by plain lower-case letters, vectors by bold lower-case letters, and matrices by bold upper-case letters. The symbol $\textbf{I}_p$ is used to represent the $p\times p$ identity matrix, the symbol $\textbf{1}_p$ represents the all-ones $p\times 1$ vector, and the symbol $\textbf{0}_p$ represents the all-zeros $p\times 1$ vector. The notation $||\cdot||$ is used to symbolize the Euclidean norm when its argument is a vector and the spectral norm when its argument is a matrix. The operator $\lceil\cdot\rceil$ rounds its argument up to the nearest integer. Almost-sure convergence is denoted by $\xrightarrow{\text{a.s.}}$ or $a\asymp b$ which means $a-b\xrightarrow{\text{a.s.}}0$ . The function $\Phi(\cdot)$ denotes the standard Gaussian CDF. The following is a list of acronyms that occur throughout the paper.
\\

\begin{acronym}[RP-LDA] 
\acro{LDA}{Linear Discriminant Analysis}
\acro{RLDA}{Regularized Linear Discriminant Analysis}
\acro{QDA}{Quadratic Discriminant Analysis}
\acro{RP-LDA}{Randomly Projected Linear Discriminant Analysis}
\acro{RQDA}{Regularized Quadratic Discriminant Analysis}
\acro{RDA}{Regularized Discriminant Analysis}
\acro{PE}{Prediction Error}
\acro{EPE}{Expected Prediction Error}
\acro{DE}{Deterministic Equivalent}
\end{acronym}

\section{The RP-LDA Ensemble Classifier}\label{section_classifier}
For the current work, we consider binary classification under a supervised setting. We assume the following setup for which we state the decision rules of LDA, RP-LDA, and the RP-LDA ensemble in succession for known and unknown statistics.

A data point $\textbf{x}\in\mathbb{R}^p$ belongs to one of two classes $\mathcal{C}_0$ and $\mathcal{C}_1$ having prior probabilities $\pi_0$ and $\pi_1$, respectively. Conditioned on its class, we assume that $\textbf{x}$ is Gaussian distributed, with distinct means and a common covariance between the two classes, as follows:
 \begin{align}
&\textbf{x}|\textbf{x}\in\mathcal{C}_i\sim \mathcal{N}\left(\bm{\mu}_i,\bm{\Sigma}\right), \ i=0,1 \label{dist}
 \end{align}
 We have a training set of $n$ instances of training data distributed as \eqref{dist}. The training set consists of pairs of data points and their labels. More formally, the set of $n$ training data points is $\mathcal{T}=\{(\textbf{x}_i,{y}_i)\}_{i=1}^n$, where $\textbf{x}_i$ is a data point and ${y}_i\in\{0,1\}$ is its corresponding label. This set contains $n_0$ and $n_1$ sample points from $\mathcal{C}_0$ and $\mathcal{C}_1$, respectively. In what follows, we denote the decision rule on point $\textbf{x}$ of each classifier by  $h(\textbf{x},\tilde{\bm{\mu}}_0,\tilde{\bm{\mu}}_1,\tilde{\bm{\Sigma}},\tilde{{\pi}}_0,\tilde{{\pi}}_1)$ and an identifying subscript. The parameters $\tilde{\bm{\mu}}_0$, $\tilde{\bm{\mu}}_1$, $\tilde{\bm{\Sigma}}$, $\tilde{\bm{\pi}}_0$, and $\tilde{\bm{\pi}}_1$ take on different values depending on which of the statistics are known by the classifier, if any. This is elaborated on later. Additionally, we denote by $\textbf{X}_0\in \mathbb{R}^{p\times n_0}$ the matrix having the vectors in the set $\{\textbf{x}_i\in\mathcal{C}_0\}$ as its successive columns, and similarly denote by $\textbf{X}_1\in \mathbb{R}^{p\times n_1}$ the matrix having the vectors in the set $\{\textbf{x}_i\in\mathcal{C}_1\}$ as its successive columns.

 In this setting, maximizing the posterior probability $\mathbb{P}[\mathcal{C}_i|\textbf{x}_q]$, given the class-conditional data distribution statistics, yields LDA. This is the optimal Bayes classifier when applied to data for which the class-conditional distributions are indeed Gaussian with common covariance and when the statistics are known. Defining a general decision rule
 \begin{equation}
   {h}_{\text{LDA}}(\textbf{x},\tilde{\bm{\mu}}_0,\tilde{\bm{\mu}}_1,\tilde{\bm{\Sigma}},\tilde{\pi}_0,\tilde{\pi}_1):=\mathbbm{1}\left\{({\tilde{\bm{\mu}}}_1-\tilde{\bm{\mu}}_0)^T\tilde{\bf{\Sigma}}^{-1}\left(\textbf{x}-\frac{\tilde{\bm{\mu}}_0+\tilde{\bm{\mu}}_1}{2}\right)+\text{ln}\frac{\tilde{\pi}_1}{\tilde{\pi}_0}>0\right\},
 \end{equation}
  where $\mathbbm{1}\{\cdot\}$ is the indicator function, the optimal Bayes classifier LDA is then the special case \\ ${h}_{\text{LDA}}(\textbf{x},\bm{\mu}_0,\bm{\mu}_1,\bm{\Sigma},\pi_0,\pi_1)$. In practice, the true statistics of the data are unknown and the LDA classifier is learned on $\textbf{X}_0$ and $\textbf{X}_1$ by computing the maximum likelihood estimates $\hat{\bm{\mu}}_0$, $\hat{\bm{\mu}}_1$, $\hat{\bm{\Sigma}}$, $\hat{\pi}_0$, and $\hat{\pi}_1$ of the true statistics $\bm{\mu}_0$, $\bm{\mu}_1$, $\bf{\Sigma}$, and prior probabilities $\pi_0$ and $\pi_1$. These estimates are the sample means $\hat{\bm{\mu}}_0=\frac{1}{n_0}\textbf{X}_0\textbf{1}_{n_0}$ and $\hat{\bm{\mu}}_1=\frac{1}{n_1}\textbf{X}_1\textbf{1}_{n_1}$, pooled sample covariance matrix $\hat{\bm{\Sigma}}=\frac{(n_0-1)\hat{\bm{\Sigma}}_0+(n_1-1)\hat{\bm{\Sigma}}_1}{n_0+n_1-2}$, and the prior probability estimates $\hat{\pi}_0=\frac{n_0}{n}$ and $\hat{\pi}_1=\frac{n_1}{n}$, respectively,
  where $\hat{\bm{\Sigma}}_0=\frac{1}{n_0-1}\left(\textbf{X}_0-\hat{\bm{\mu}}_0\textbf{1}^T\right)\left(\textbf{X}_0-\hat{\bm{\mu}}_0\textbf{1}^T\right)^T$ and $\hat{\bm{\Sigma}}_1=\frac{1}{n_1-1}\left(\textbf{X}_1-\hat{\bm{\mu}}_1\textbf{1}^T\right)\left(\textbf{X}_1-\hat{\bm{\mu}}_1\textbf{1}^T\right)^T$. In this case, the LDA decision rule is given by ${h}_{\text{LDA}}(\textbf{x},\hat{\bm{\mu}}_0,\hat{\bm{\mu}}_1,\hat{\bm{\Sigma}},\hat{\pi}_0,\hat{\pi}_1)$.
  As mentioned in the introduction, when $n< p$, $\hat{\bm{\Sigma}}$ is singular. To deal with this issue, we consider reducing the dimensionality of the data by random projection.

  Random projection is a non-adaptive dimensionality reduction technique. In random projection, a matrix $\textbf{R}\in \mathbb{R}^{d\times p}$, with $d<p$, whose entries are generated i.i.d. from a zero-mean Gaussian distribution \cite{durrant2013learning}, multiplies each data point so that it is projected onto a lower-dimensional random subspace. The term \textit{randomly projected classifier} refers to the classifier being trained on data that has been projected onto the random column space of $\textbf{R}$; instead of being learned on $\textbf{X}_0$ and $\textbf{X}_1$, the classifier is learned on $\textbf{R}\textbf{X}_0$ and $\textbf{R}\textbf{X}_1$. Projecting the training data as $\textbf{R}\textbf{X}_0$ and $\textbf{R}\textbf{X}_1$ results in the following statistic estimates as a function of the old estimates
  \begin{eqnarray}
  &\hat{\bm{\mu}}_0^{\text{RP}}=\textbf{R}\hat{\bm{\mu}}_0 \nonumber , \
  \hat{\bm{\mu}}_1^{\text{RP}}=\textbf{R}\hat{\bm{\mu}}_1 \nonumber , \ \text{and }
  \hat{\bm{\Sigma}}^{\text{RP}}=\textbf{R}\hat{\bm{\Sigma}}\textbf{R}^T
  \end{eqnarray}
When the projection dimension $d$ is chosen such that $d\le\text{rank}(\hat{\bm{\Sigma}})$ (noting that $\text{rank}(\hat{\bm{\Sigma}})\le n-2$), the resulting $\hat{\bm{\Sigma}}^{\text{RP}}$ must be invertible since $\textbf{R}$ is almost surely of rank $d$ \cite{durrant2013learning}. The decision rule for the RP-LDA classifier, obtained by simply plugging the new estimates into the LDA decision rule, is ${h}_{\text{RP-LDA}}(\textbf{x},\hat{\bm{\mu}}_0,\hat{\bm{\mu}}_1,\hat{\bm{\Sigma}},\hat{\pi}_0,\hat{\pi}_1)$, where more generally
\begin{align}
    &{h}_{\text{RP-LDA}}(\textbf{x},\tilde{\bm{\mu}}_0,\tilde{\bm{\mu}}_1,\tilde{\bm{\Sigma}},\tilde{\pi}_0,\tilde{\pi}_1):=
     \mathbbm{1}\left\{(\tilde{\bm{\mu}}_1-\tilde{\bm{\mu}}_0)^T\textbf{\textbf{R}}^T(\textbf{R}\tilde{\bm{\Sigma}}\textbf{R}^T)^{-1}\textbf{R}\left(\textbf{x}-\frac{\tilde{\bm{\mu}}_0+\tilde{\bm{\mu}}_1}{2}\right)+\text{ln}\frac{\tilde{\pi}_1}{\tilde{\pi}_0}>0\right\}.
      \label{RPLDA}
\end{align}
The decision rule of this classifier in the special case when the statistics are known is \\ ${h}_{\text{RP-LDA}}(\textbf{x},\bm{\mu}_0,\bm{\mu}_1,\bm{\Sigma},\pi_0,\pi_1)$. In this case, the quantities $\textbf{R}{\bm{\mu}}_0$, $\textbf{R}{\bm{\mu}}_1$, and $\textbf{R}{\bm{\Sigma}}\textbf{R}^T$ represent the statistics of the class-conditional distributions of the data in the projected space.

As LDA trained on a single random projection typically performs poorly in practice, we look into ensembles of randomly-projected classifiers, particularly the ensemble of RP-LDA classifiers formulated in \cite{durrant2013random}. A more recent random projection ensemble classification framework based on majority voting is proposed in \cite{cannings2017random}, however this classifier is much more difficult to implement in practice and to analyze. In \cite{durrant2013random}, the discriminant is an average of $M$ individual RP-LDA discriminants, each corresponding to a different projection applied to the same training data. The decision rule for this particular RP-LDA ensemble is ${h}_{\text{RP-LDA}}^{\text{ens}}(\textbf{x},\hat{\bm{\mu}}_0,\hat{\bm{\mu}}_1,\hat{\bm{\Sigma}},\hat{\pi}_0,\hat{\pi}_1)$, where more generally
\begin{align}
     &{h}_{\text{RP-LDA}}^{\text{ens}}(\textbf{x},\tilde{\bm{\mu}}_0,\tilde{\bm{\mu}}_1,\tilde{\bm{\Sigma}},\tilde{\pi}_0,\tilde{\pi}_1):= 
     \mathbbm{1}\left\{\frac{1}{M}\sum_{i=1}^M(\tilde{\bm{\mu}}_1-\tilde{\bm{\mu}}_0)^T\textbf{R}_i^T(\textbf{R}_i\tilde{\Sigma}\textbf{R}_i^T)^{-1}\textbf{R}_i\left(\textbf{x}-\frac{\tilde{\bm{\mu}}_0+\tilde{\bm{\mu}}_1}{2}\right)+\text{ln}\frac{\tilde{\pi}_1}{\tilde{\pi}_0}>0\right\}.\label{LDAens}
 \end{align}
 When the statistics are known, the decision rule takes the form ${h}_{\text{RP-LDA}}^{\text{ens}}(\textbf{x},{\bm{\mu}}_0,{\bm{\mu}}_1,{\bm{\Sigma}},{\pi}_0,{\pi}_1)$. 
 
 From \eqref{LDAens}, we see that the choice of variance of the entries of each random projection matrix in the ensemble has no effect on the behavior of the RP-LDA ensemble classifier, as scaling $\textbf{R}_i$ by any real constant yields the same decision rule. In this work, all random projection matrices are specified as having i.i.d. entries $R_{i,j}\sim \mathcal{N}(0,\frac{1}{d}), \ \forall i,j$, where the choice of variance is purely to facilitate the application of random matrix theory results later.
 
 \section{Asymptotic Performance Analysis of the RP-LDA Ensemble Classifier}\label{section_analysis}
In this section, we pursue the asymptotic analysis of the RP-LDA ensemble classifier misclassification probability. To begin with, we construct the RP-LDA infinite ensemble and derive an expression for its probability of misclassification. We then define the asymptotic growth regime and present the asymptotic misclassification probabilities and the generalized consistent estimator of the classification error.
\subsection{The RP-LDA Infinite Ensemble}\label{subsection_error}
We take the limit as $M$, the number of random projections in the ensemble, goes to infinity. This yields an expectation in the discriminant rather than a sum and facilitates the analysis by random matrix theory. We call this mathematical object the \textit{RP-LDA infinite ensemble}. Although it cannot be realized as a classifier, it can be approximated by a finite ensemble for $M$ large enough. In this section, we construct the RP-LDA infinite ensemble classifier and derive its probability of misclassification.
\subsubsection{Construction}
Let ${W}_{\text{RP-RLDA}}^{\text{ens}}(\textbf{x},\tilde{\bm{\mu}}_0,\tilde{\bm{\mu}}_1,\tilde{\bm{\Sigma}},\tilde{\pi}_0,\tilde{\pi}_1)$ denote the discriminant of \eqref{LDAens}. In the limit as $M\rightarrow\infty$, for fixed $d$, $p$, and $n$, the discriminant becomes
 \begin{align}
     \lim_{M\rightarrow\infty} {W}_{\text{RP-RLDA}}^{\text{ens}}(\textbf{x},\tilde{\bm{\mu}}_0,\tilde{\bm{\mu}}_1,\tilde{\bm{\Sigma}},\tilde{\pi}_0,\tilde{\pi}_1)&=(\tilde{\bm{\mu}}_1-\tilde{\bm{\mu}}_0)^T\left(\lim_{M\rightarrow\infty}\frac{1}{M}\sum_{i=1}^M\textbf{R}_i^T(\textbf{R}_i\tilde{\bm{\Sigma}}\textbf{R}_i^T)^{-1}\textbf{R}_i\right)\left(\textbf{x}-\frac{\tilde{\bm{\mu}}_0+\tilde{\bm{\mu}}_1}{2}\right)+\text{ln}\frac{\tilde{\pi}_1}{\tilde{\pi}_0}& \nonumber \\
    &=(\tilde{\bm{\mu}}_1-\tilde{\bm{\mu}}_0)^T\mathbb{E}_{\textbf{R}}\left[\textbf{R}^T(\textbf{R}\tilde{\bm{\Sigma}}\textbf{R}^T)^{-1}\textbf{R}\right]\left(\textbf{x}-\frac{\tilde{\bm{\mu}}_0+\tilde{\bm{\mu}}_1}{2}\right)+\text{ln}\frac{\tilde{\pi}_1}{\tilde{\pi}_0},&
    \label{asymptotic}
 \end{align}
where the second step follows from the law of large numbers and the expectation is with respect to $\textbf{R}$, conditioned on the parameters $\tilde{\bm{\mu}}_0$, $\tilde{\bm{\mu}}_1,\tilde{\bm{\Sigma}}$, $\tilde{\pi}_0,\tilde{\pi}_1$. The expectation exists for $d\in\left\{1,\ldots,\text{rank}(\hat{\bm{\Sigma}})-2\right\}\cup\left\{\text{rank}(\hat{\bm{\Sigma}})+2,\ldots,p\right\}$ \cite{durrant2013random}, however, since we need $\hat{\bm{\Sigma}}^{\text{RP}}$ to be invertible, we restrict $d\le\text{rank}(\hat{\bm{\Sigma}})-2$. Let
 \begin{align}
     {W}_{\text{RP-LDA}}^{\infty-\text{ens}}(\textbf{x},\tilde{\bm{\mu}}_0,\tilde{\bm{\mu}}_1,\tilde{\bm{\Sigma}},\tilde{\pi}_0,\tilde{\pi}_1)
     &=(\tilde{\bm{\mu}}_1-\tilde{\bm{\mu}}_0)^T\mathbb{E}_{\textbf{R}}\left[\textbf{R}^T(\textbf{R}\tilde{\bm{\Sigma}}\textbf{R}^T)^{-1}\textbf{R}\right]\left(\textbf{x}-\frac{\tilde{\bm{\mu}}_0+\tilde{\bm{\mu}}_1}{2}\right)+\text{ln}\frac{\tilde{\pi}_1}{\tilde{\pi}_0}.
 \end{align}
 We define the RP-LDA infinite ensemble as the classifier with the decision rule
 \begin{equation}
     {h}_{\text{RP-LDA}}^{\infty-\text{ens}}(\textbf{x},\tilde{\bm{\mu}}_0,\tilde{\bm{\mu}}_1,\tilde{\bm{\Sigma}},\tilde{\pi}_0,\tilde{\pi}_1):=\mathbbm{1}\{ {W}_{\text{RP-LDA}}^{\infty-\text{ens}}(\textbf{x},\tilde{\bm{\mu}}_0,\tilde{\bm{\mu}}_1,\tilde{\bm{\Sigma}},\tilde{\pi}_0,\tilde{\pi}_1)>0\}.
 \end{equation}
 
 \subsubsection{Derivation of the Error}
 We now characterize the probability of misclassification of a point $\textbf{x}$ by the randomly projected LDA infinite ensemble. Denote by $\varepsilon(\tilde{\bm{\mu}}_0,\tilde{\bm{\mu}}_1,\tilde{\bm{\Sigma}},\tilde{\pi}_0,\tilde{\pi}_1)$ the probability of misclassification given the parameters $\tilde{\bm{\mu}}_0$, $\tilde{\bm{\mu}}_1$,$\tilde{\bm{\Sigma}}$, $\tilde{\pi}_0$, and $\tilde{\pi}_1$ which, as before, take values according to the knowledge of the class-conditional distribution statistics. Accordingly, $\varepsilon({\bm{\mu}}_0,{\bm{\mu}}_1,{\bm{\Sigma}},{\pi}_0,{\pi}_1)$ corresponds to the error of the RP-LDA ensemble operating under known statistics and $\varepsilon(\hat{\bm{\mu}}_0,\hat{\bm{\mu}}_1,\hat{\bm{\Sigma}},\hat{\pi}_0,\hat{\pi}_1)$ corresponds to the error of the RP-LDA infinite ensemble operating under unknown statistics conditioned on the training set $\mathcal{T}$ (and therefore given the parameters). Lemma $1$ presents the expressions for the probability of misclassification given the parameters and the corresponding expected probability of misclassification over $\mathcal{T}$.

 {\bf Lemma 1}
\textit{Let $m_0(\tilde{\bm{\mu}}_0,\tilde{\bm{\mu}}_1,\tilde{\bm{\Sigma}},\tilde{\pi}_0,\tilde{\pi}_1)$, $m_1(\tilde{\bm{\mu}}_0,\tilde{\bm{\mu}}_1,\tilde{\bm{\Sigma}},\tilde{\pi}_0,\tilde{\pi}_1)$, and $\sigma^2(\tilde{\bm{\mu}}_0,\tilde{\bm{\mu}}_1,\tilde{\bm{\Sigma}})$ be defined as} 
 \begin{equation}
     m_0(\tilde{\bm{\mu}}_0,\tilde{\bm{\mu}}_1,\tilde{\bm{\Sigma}},\tilde{\pi}_0,\tilde{\pi}_1)=(\tilde{\bm{\mu}}_1-\tilde{\bm{\mu}}_0)^T\mathbb{E}_{\textbf{R}}\left[\textbf{R}^T(\textbf{R}\tilde{\Sigma}\textbf{R}^T)^{-1}\textbf{R}\right]\left(\bm{\mu}_0-\frac{\tilde{\bm{\mu}}_0+\tilde{\bm{\mu}}_1}{2}\right)+\text{ln}\frac{\tilde{\pi}_1}{\tilde{\pi}_0}
    \label{m0}
 \end{equation}
 \begin{equation}
     m_1(\tilde{\bm{\mu}}_0,\tilde{\bm{\mu}}_1,\tilde{\bm{\Sigma}},\tilde{\pi}_0,\tilde{\pi}_1)=(\tilde{\bm{\mu}}_1-\tilde{\bm{\mu}}_0)^T\mathbb{E}_{\textbf{R}}\left[\textbf{R}^T(\textbf{R}\tilde{\Sigma}\textbf{R}^T)^{-1}\textbf{R}\right]\left(\bm{\mu}_1-\frac{\tilde{\bm{\mu}}_0+\tilde{\bm{\mu}}_1}{2}\right)+\text{ln}\frac{\tilde{\pi}_1}{\tilde{\pi}_0}\label{m1}
 \end{equation}
 and
 \begin{align}
     \sigma^2(\tilde{\bm{\mu}}_0,\tilde{\bm{\mu}}_1,\tilde{\bm{\Sigma}})&=(\tilde{\bm{\mu}}_1-\tilde{\bm{\mu}}_0)^T\mathbb{E}_{\textbf{R}}\left[\textbf{R}^T(\textbf{R}\tilde{\Sigma}\textbf{R}^T)^{-1}\textbf{R}\right]\bm{\Sigma}\mathbb{E}_{\textbf{R}}\left[\textbf{R}^T(\textbf{R}\tilde{\Sigma}\textbf{R}^T)^{-1}\textbf{R}\right](\tilde{\bm{\mu}}_1-\tilde{\bm{\mu}}_0)\label{sigma}.
 \end{align} \textit{The exact probability of misclassification of the RP-LDA infinite ensemble classifier conditioned on the parameters $\tilde{\bm{\mu}}_0$, $\tilde{\bm{\mu}}_1$,$\tilde{\bm{\Sigma}}$, $\tilde{\pi}_0$, also known as the PE, is}
 \begin{equation}
     \varepsilon(\tilde{\bm{\mu}}_0,\tilde{\bm{\mu}}_1,\tilde{\bm{\Sigma}},\tilde{\pi}_0,\tilde{\pi}_1)={\pi}_0\Phi\left(\frac{m_0(\tilde{\bm{\mu}}_0,\tilde{\bm{\mu}}_1,\tilde{\bm{\Sigma}},\tilde{\pi}_0,\tilde{\pi}_1)}{\sqrt{\sigma^2(\tilde{\bm{\mu}}_0,\tilde{\bm{\mu}}_1,\tilde{\bm{\Sigma}})}}\right)+{\pi}_1\Phi\left(-\frac{m_1(\tilde{\bm{\mu}}_0,\tilde{\bm{\mu}}_1,\tilde{\bm{\Sigma}},\tilde{\pi}_0,\tilde{\pi}_1)}{\sqrt{\sigma^2(\tilde{\bm{\mu}}_0,\tilde{\bm{\mu}}_1,\tilde{\bm{\Sigma}})}}\right),
     \label{final}
 \end{equation}
\textit{and the corresponding expected misclassification probability over the training set $\mathcal{T}$, also known as the generalization error, or the EPE is}
 \begin{equation}
 \mathbb{E}_{\tilde{\bm{\mu}}_0,\tilde{\bm{\mu}}_1,\tilde{\bm{\Sigma}},\tilde{\pi}_0,{\pi}_1}\left[{\pi}_0\Phi\left(\frac{m_0(\tilde{\bm{\mu}}_0,\tilde{\bm{\mu}}_1,\tilde{\bm{\Sigma}},\tilde{\pi}_0,\tilde{\pi}_1)}{\sqrt{\sigma^2(\tilde{\bm{\mu}}_0,\tilde{\bm{\mu}}_1,\tilde{\bm{\Sigma}})}}\right)+{\pi}_1\Phi\left(-\frac{m_1(\tilde{\bm{\mu}}_0,\tilde{\bm{\mu}}_1,\tilde{\bm{\Sigma}},\tilde{\pi}_0,\tilde{\pi}_1)}{\sqrt{\sigma^2(\tilde{\bm{\mu}}_0,\tilde{\bm{\mu}}_1,\tilde{\bm{\Sigma}})}}\right)\right].\label{finalerrorEPE}
 \end{equation}

 \textbf{Proof:} By the law of total probability
 \begin{equation}
 \varepsilon(\tilde{\bm{\mu}}_0,\tilde{\bm{\mu}}_1,\tilde{\bm{\Sigma}},\tilde{\pi}_0,\tilde{\pi}_1)={\pi}_0\varepsilon_0(\tilde{\bm{\mu}}_0,\tilde{\bm{\mu}}_1,\tilde{\bm{\Sigma}},\tilde{\pi}_0,\tilde{\pi}_1)+{\pi}_1\varepsilon_1(\tilde{\bm{\mu}}_0,\tilde{\bm{\mu}}_1,\tilde{\bm{\Sigma}},\tilde{\pi}_0,\tilde{\pi}_1)
 \end{equation}
 where $\varepsilon_0(\tilde{\bm{\mu}}_0,\tilde{\bm{\mu}}_1,\tilde{\bm{\Sigma}},\tilde{\pi}_0,\tilde{\pi}_1)$ is the probability of misclassification given the parameters and given that the point belongs to $\mathcal{C}_0$. Similarly, $\varepsilon_1(\tilde{\bm{\mu}}_0,\tilde{\bm{\mu}}_1,\tilde{\bm{\Sigma}},\tilde{\pi}_0,\tilde{\pi}_1)$ is the probability of misclassification given the parameters and given that the point belongs to $\mathcal{C}_1$. In terms of the discriminant ${W}_{\text{RP-LDA}}^{\infty-\text{ens}}(\textbf{x},\tilde{\bm{\mu}}_0,\tilde{\bm{\mu}}_1,\tilde{\bm{\Sigma}},\tilde{\pi}_0,\tilde{\pi}_1)$,
 \begin{equation}
     \varepsilon_0(\tilde{\bm{\mu}}_0,\tilde{\bm{\mu}}_1,\tilde{\bm{\Sigma}},\tilde{\pi}_0,\tilde{\pi}_1)=\mathbb{P}[{W}_{\text{RP-LDA}}^{\infty-\text{ens}}(\textbf{x},\tilde{\bm{\mu}}_0,\tilde{\bm{\mu}}_1,\tilde{\bm{\Sigma}},\tilde{\pi}_0,\tilde{\pi}_1)>0|\textbf{x}\in \mathcal{C}_0,\tilde{\bm{\mu}}_0,\tilde{\bm{\mu}}_1,\tilde{\bm{\Sigma}},\tilde{\pi}_0,\tilde{\pi}_1]
 \end{equation}
 and
 \begin{equation}
     \varepsilon_1(\tilde{\bm{\mu}}_0,\tilde{\bm{\mu}}_1,\tilde{\bm{\Sigma}},\tilde{\pi}_0,\tilde{\pi}_1)=\mathbb{P}[{W}_{\text{RP-LDA}}^{\infty-\text{ens}}(\textbf{x},\tilde{\bm{\mu}}_0,\tilde{\bm{\mu}}_1,\tilde{\bm{\Sigma}},\tilde{\pi}_0,\tilde{\pi}_1)<0|\textbf{x}\in \mathcal{C}_1,\tilde{\bm{\mu}}_0,\tilde{\bm{\mu}}_1,\tilde{\bm{\Sigma}},\tilde{\pi}_0,\tilde{\pi}_1].
 \end{equation}
 Conditioned on the classes and the parameters as such, the discriminant is a Gaussian random variable. More specifically,
  \begin{align}
     {W}_{\text{RP-LDA}}^{\infty-\text{ens}}(\textbf{x},\tilde{\bm{\mu}}_0,\tilde{\bm{\mu}}_1,\tilde{\bm{\Sigma}},\tilde{\pi}_0,\tilde{\pi}_1)|\textbf{x}\in \mathcal{C}_0,\tilde{\bm{\mu}}_0,\tilde{\bm{\mu}}_1,\tilde{\bm{\Sigma}},\tilde{\pi}_0,\tilde{\pi}_1 
     &\sim\mathcal{N}(m_0(\tilde{\bm{\mu}}_0,\tilde{\bm{\mu}}_1,\tilde{\bm{\Sigma}},\tilde{\pi}_0,\tilde{\pi}_1),\sigma^2(\tilde{\bm{\mu}}_0,\tilde{\bm{\mu}}_1,\tilde{\bm{\Sigma}}))
 \end{align}
 and
  \begin{align}
     {W}_{\text{RP-LDA}}^{\infty-\text{ens}}(\textbf{x},\tilde{\bm{\mu}}_0,\tilde{\bm{\mu}}_1,\tilde{\bm{\Sigma}},\tilde{\pi}_0,\tilde{\pi}_1)|\textbf{x}\in \mathcal{C}_1,\tilde{\bm{\mu}}_0,\tilde{\bm{\mu}}_1,\tilde{\bm{\Sigma}},\tilde{\pi}_0,\tilde{\pi}_1 
     &\sim\mathcal{N}( m_1(\tilde{\bm{\mu}}_0,\tilde{\bm{\mu}}_1,\tilde{\bm{\Sigma}},\tilde{\pi}_0,\tilde{\pi}_1),\sigma^2(\tilde{\bm{\mu}}_0,\tilde{\bm{\mu}}_1,\tilde{\bm{\Sigma}})),
 \end{align}
 where $m_0(\tilde{\bm{\mu}}_0,\tilde{\bm{\mu}}_1,\tilde{\bm{\Sigma}},\tilde{\pi}_0,\tilde{\pi}_1)$, $m_1(\tilde{\bm{\mu}}_0,\tilde{\bm{\mu}}_1,\tilde{\bm{\Sigma}},\tilde{\pi}_0,\tilde{\pi}_1)$, and $\sigma^2(\tilde{\bm{\mu}}_0,\tilde{\bm{\mu}}_1,\tilde{\bm{\Sigma}})$ are as defined by \eqref{m0}, \eqref{m1}, and \eqref{sigma} respectively. From this it can be shown that
 \begin{equation}
     \varepsilon_0(\tilde{\bm{\mu}}_0,\tilde{\bm{\mu}}_1,\tilde{\bm{\Sigma}},\tilde{\pi}_0,\tilde{\pi}_1)=\Phi\left(\frac{m_0(\tilde{\bm{\mu}}_0,\tilde{\bm{\mu}}_1,\tilde{\bm{\Sigma}},\tilde{\pi}_0,\tilde{\pi}_1)}{\sqrt{\sigma^2(\tilde{\bm{\mu}}_0,\tilde{\bm{\mu}}_1,\tilde{\bm{\Sigma}})}}\right)
 \end{equation}
 and
  \begin{equation}
     \varepsilon_1(\tilde{\bm{\mu}}_0,\tilde{\bm{\mu}}_1,\tilde{\bm{\Sigma}},\tilde{\pi}_0,\tilde{\pi}_1)=\Phi\left(-\frac{m_1(\tilde{\bm{\mu}}_0,\tilde{\bm{\mu}}_1,\tilde{\bm{\Sigma}},\tilde{\pi}_0,\tilde{\pi}_1)}{\sqrt{\sigma^2(\tilde{\bm{\mu}}_0,\tilde{\bm{\mu}}_1,\tilde{\bm{\Sigma}})}}\right)
 \end{equation}
and the expression in \eqref{final} follows. Taking the expectation of \eqref{final} over the parameters yields the expression in \eqref{finalerrorEPE}.
\subsection{Main Results}\label{subsection_results}
\subsubsection{Deterministic Equivalents}\label{DEs}
In this section, we first present an asymptotic characterization of the effect of the infinite ensemble on classification error. This shows that the ensemble regularizes $\tilde{\bm{\Sigma}}$ by a function of the projection dimension $d$. We then present asymptotic misclassification probabilities of the RP-LDA infinite ensemble under various combinations of known and unknown statistics. We consider special cases of these under which the expressions are closed form and compare them to existing analogous results for the LDA and RP-LDA classifiers. Through this we develop a deeper understanding of the RP-LDA ensemble classifier's behavior and identify situations in which it may be especially advantageous. Derivation of the the asymptotic errors is also the first step to developing the error G-estimator in Section \ref{gest}.

Formally, the asymptotic misclassification probabilities $\bar{\varepsilon}_{\tilde{\bm{\mu}}_0,\tilde{\bm{\mu}}_1,\tilde{\bm{\Sigma}}}$ are deterministic sequences   of $n$, $p$, and $d$, called deterministic equivalents (DEs), which satisfy
\begin{equation}
     \varepsilon(\tilde{\bm{\mu}}_0,\tilde{\bm{\mu}}_1,\tilde{\bm{\Sigma}},\tilde{\pi}_0,\tilde{\pi}_1)-\bar{\varepsilon}_{\tilde{\bm{\mu}}_0,\tilde{\bm{\mu}}_1,\tilde{\bm{\Sigma}}}\xrightarrow{\text{a.s.}}0\label{DEdef}
\end{equation}
under the conditions
\begin{enumerate}[{\hspace{20px}(a)}]
    \item $0<\liminf\frac{p}{n}<\limsup\frac{p}{n}<\infty$
    \item $0<\liminf\frac{d}{n}<\limsup\frac{d}{n}<1$
    \item $0<\liminf\frac{d}{p}<\limsup\frac{d}{p}<1$
    \item $\frac{n_i}{n}\rightarrow c_i\in(0,1), \ i=0,1$
    \item  $\limsup\limits_{p} \Vert\bm{\mu}_0-\bm{\mu}_1\Vert_2<\infty$
    \item  $\limsup\limits_{p} \Vert\bm{\Sigma}\Vert_2<\infty$
    \item $\liminf\limits_{p} \lambda_{\text{min}}\left(\bm{\Sigma}\right)>0$
\end{enumerate}
The subscript in $\bar{\varepsilon}_{\tilde{\bm{\mu}}_0,\tilde{\bm{\mu}}_1,\tilde{\bm{\Sigma}}}$ indicates the parameters of the corresponding classifier. The parameters $\tilde{\pi}_0$ and $\tilde{\pi}_1$ are dropped, because whether they are set to the true prior probabilities or their estimates makes no difference asymptotically. The conditions (a), (b), (c), and (d) specify the growth regime we wish to study, where the dimensions grow at constant rates to each other. We consider the general case where $p$ may be greater or less than $n$. This is represented in (a). The involved derivations require that $d$ be less than $n$, specified in (b), and, furthermore, less than $\text{rank}(\hat{\bm{\Sigma}})-2$ for the expectation in \eqref{asymptotic} to exist. The third condition, (c), imposes that $d<p$. This comes from the definition of random projection as a dimensionality reduction but is also imposed by the derivations. The fourth condition, (d), likewise follows by definition since $n=n_0+n_1$. Note that here the constants $c_0$ and $c_1$ are defined distinctly from the prior probabilities $\pi_0$ and $\pi_1$. This is to account for cases when the ratio of classes in the sample may not reflect the true distribution of the classes within the population. The two conditions (e) and (f) are technicalities stemming from the use of random matrix theory tools. The context in which they become necessary is detailed in the appendix. Finally, condition (g) is necessary so that all members of the sequence, $\hat{\bm{\Sigma}}^{\text{RP}}=\textbf{R}\hat{\bm{\Sigma}}\textbf{R}^T$ as $d$, $p$ and $n$ grow, are invertible.

The DE $\bar{\varepsilon}_{\tilde{\bm{\mu}}_0,\tilde{\bm{\mu}}_1,\tilde{\bm{\Sigma}}}$ can be constructed out of corresponding DEs for each of the statistics of the class-conditional discriminant. This is presented in Lemma 2. Note that it follows from \eqref{DEdef} that
\begin{equation}
     \mathbb{E}_{\tilde{\bm{\mu}}_0,\tilde{\bm{\mu}}_1,\tilde{\bm{\Sigma}},\tilde{\pi}_0,\tilde{\pi}_1}\left[\varepsilon(\tilde{\bm{\mu}}_0,\tilde{\bm{\mu}}_1,\tilde{\bm{\Sigma}},\tilde{\pi}_0,\tilde{\pi}_1)\right]-\bar{\varepsilon}_{\tilde{\bm{\mu}}_0,\tilde{\bm{\mu}}_1,\tilde{\bm{\Sigma}}}\xrightarrow{\text{a.s.}}0\label{epe}
\end{equation}
and so $\bar{\varepsilon}_{\tilde{\bm{\mu}}_0,\tilde{\bm{\mu}}_1,\tilde{\bm{\Sigma}}}$ doubles as a deterministic equivalent of the generalization error defined in \eqref{finalerrorEPE}.

{\bf Lemma 2}
\textit{Let $\bar{m}_{0,\tilde{\bm{\mu}}_0,\tilde{\bm{\mu}}_1,\tilde{\bm{\Sigma}}}$, $\bar{m}_{1,\tilde{\bm{\mu}}_0,\tilde{\bm{\mu}}_1,\tilde{\bm{\Sigma}}}$, and $\bar{\sigma}^2_{\tilde{\bm{\mu}}_0,\tilde{\bm{\mu}}_1,\tilde{\bm{\Sigma}}}$ be deterministic sequences of $d$, $p$, and $n$ such that}
 \begin{align*}
     m_0(\tilde{\bm{\mu}}_0,\tilde{\bm{\mu}}_1,\tilde{\bm{\Sigma}},\tilde{\pi}_0,\tilde{\pi}_1)-\bar{m}_{0,\tilde{\bm{\mu}}_0,\tilde{\bm{\mu}}_1,\tilde{\bm{\Sigma}}}&\xrightarrow{\text{a.s.}}0\\
     m_1(\tilde{\bm{\mu}}_0,\tilde{\bm{\mu}}_1,\tilde{\bm{\Sigma}},\tilde{\pi}_0,\tilde{\pi}_1)-\bar{m}_{1,\tilde{\bm{\mu}}_0,\tilde{\bm{\mu}}_1,\tilde{\bm{\Sigma}}}&\xrightarrow{\text{a.s.}}0\\
     \sigma^2(\tilde{\bm{\mu}}_0,\tilde{\bm{\mu}}_1,\tilde{\bm{\Sigma}})-\bar{\sigma}^2_{\tilde{\bm{\mu}}_0,\tilde{\bm{\mu}}_1,\tilde{\bm{\Sigma}}}&\xrightarrow{\text{a.s.}}0
 \end{align*}
 \textit{for $n$, $p$ and $d$ growing subject to (a)-(g).
 By the continuous mapping theorem and other properties of almost sure convergence, \eqref{DEdef} and \eqref{epe} hold with}
 \begin{equation}
     \bar{\varepsilon}_{\tilde{\bm{\mu}}_0,\tilde{\bm{\mu}}_1,\tilde{\bm{\Sigma}}}={\pi}_0\Phi\left(\frac{\bar{m}_{0,\tilde{\bm{\mu}}_0,\tilde{\bm{\mu}}_1,\tilde{\bm{\Sigma}}}}{\sqrt{\bar{\sigma}^2_{\tilde{\bm{\mu}}_0,\tilde{\bm{\mu}}_1,\tilde{\bm{\Sigma}}}}}\right)+{\pi}_1\Phi\left(-\frac{\bar{m}_{1,\tilde{\bm{\mu}}_0,\tilde{\bm{\mu}}_1,\tilde{\bm{\Sigma}}}}{\sqrt{\bar{\sigma}^2_{\tilde{\bm{\mu}}_0,\tilde{\bm{\mu}}_1,\tilde{\bm{\Sigma}}}}}\right)\label{ultimateDE}.
 \end{equation}

Now we consider the deterministic equivalent of the misclassification probability with respect to the random projection, that is, given the parameters $\tilde{\bm{\mu}}_0$, $\tilde{\bm{\mu}}_1$, $\tilde{\bm{\Sigma}}$, $\tilde{\pi}_0$, and $\tilde{\pi}_1$. This yields an asymptotic, deterministic expression of the random effect of the infinite ensemble which approximates the random effect of the finite ensemble in the finite regime. As per Lemma 2, we derive the deterministic equivalents of each of the class-conditional discriminant statistics with respect to the random projection ensemble.  First we define the quantities $\zeta_{\bm{\Sigma}}\left(\bm{\Sigma}\right)$ and  $\zeta_{\hat{\bm{\Sigma}}}\left(\bm{\Sigma}\right)$ and then present the DEs in Theorem 1.

 Define $\zeta_{\bm{\Sigma}}\left(\bm{\Sigma}\right)$ as the unique root of the monotonically decreasing function
 \begin{equation}
     g(x)=1-\frac{1}{d}\text{tr}\left\{\bm{\Sigma}{\left(\bm{\Sigma}+\frac{1}{x}\textbf{I}_p\right)^{-1}}\right\}\label{zetaSigma}
 \end{equation}
over $x>0$, that is,
\begin{equation}
    g(\zeta_{\bm{\Sigma}}\left(\bm{\Sigma}\right))=0. \label{zetaSigma2}
\end{equation}
Also let $\zeta_{\hat{\bm{\Sigma}}}\left(\bm{\Sigma}\right)$ be defined as
 \begin{equation}
     \zeta_{\hat{\bm{\Sigma}}}\left(\bm{\Sigma}\right)=\frac{x^*}{1-x^*\frac{1}{n}\text{tr}\left\{\bm{\Sigma}\left(x^*\bm{\Sigma}+\textbf{I}_p\right)^{-1}\right\}},\label{zetaSigmaHat2}
 \end{equation}
 where $x^*$ is the root of the monotonically decreasing function
 \begin{equation}
     h(x)=1-\frac{p}{d}+\frac{1}{d}\text{tr}\left\{\left(x\bm{\Sigma}+\textbf{I}_p\right)^{-1}\right\}\label{zetaSigmaHat1}
 \end{equation}
over $x>0$.
 The deterministic equivalents of the class-conditional discriminant statistics of the RP-LDA infinite ensemble given the parameters  are as presented in the following theorem.

\textbf{Theorem 1}\textit{ (DEs with respect to random projection)}
\textit{Under the growth regime defined by the conditions (c)-(g), the following asymptotic convergences  hold}
 \begin{align*}
     m_0(\tilde{\bm{\mu}}_0,\tilde{\bm{\mu}}_1,\tilde{\bm{\Sigma}},\tilde{\pi}_0,\tilde{\pi}_1)-(\tilde{\bm{\mu}}_1-\tilde{\bm{\mu}}_0)^T\left(\tilde{\bm{\Sigma}}+\frac{1}{\zeta_{\tilde{\bm{\Sigma}}}\left(\bm{\Sigma}\right)}\textbf{I}_p\right)^{-1}\left(\bm{\mu}_0-\frac{\tilde{\bm{\mu}}_0+\tilde{\bm{\mu}}_1}{2}\right)-\text{ln}\frac{\pi_1}{\pi_0}&\xrightarrow{\text{a.s.}}0\\
     m_1(\tilde{\bm{\mu}}_0,\tilde{\bm{\mu}}_1,\tilde{\bm{\Sigma}},\tilde{\pi}_0,\tilde{\pi}_1)-(\tilde{\bm{\mu}}_1-\tilde{\bm{\mu}}_0)^T\left(\tilde{\bm{\Sigma}}+\frac{1}{\zeta_{\tilde{\bm{\Sigma}}}\left(\bm{\Sigma}\right)}\textbf{I}_p\right)^{-1}\left(\bm{\mu}_1-\frac{\tilde{\bm{\mu}}_0+\tilde{\bm{\mu}}_1}{2}\right)-\text{ln}\frac{\pi_1}{\pi_0}&\xrightarrow{\text{a.s.}}0\\
     \sigma^2(\tilde{\bm{\mu}}_0,\tilde{\bm{\mu}}_1,\tilde{\bm{\Sigma}})-(\tilde{\bm{\mu}}_1-\tilde{\bm{\mu}}_0)^T\left(\tilde{\bm{\Sigma}}+\frac{1}{\zeta_{\tilde{\bm{\Sigma}}}\left(\bm{\Sigma}\right)}\textbf{I}_p\right)^{-1}\bm{\Sigma}\left(\tilde{\bm{\Sigma}}+\frac{1}{\zeta_{\tilde{\bm{\Sigma}}}\left(\bm{\Sigma}\right)}\textbf{I}_p\right)^{-1}(\tilde{\bm{\mu}}_1-\tilde{\bm{\mu}}_0)&\xrightarrow{\text{a.s.}}0,
 \end{align*}
\textit{where $\zeta_{\tilde{\bm{\Sigma}}}\left(\bm{\Sigma}\right)$ is as defined by \eqref{zetaSigma} and \eqref{zetaSigma2}, if $\tilde{\bm{\Sigma}}=\bm{\Sigma}$, or \eqref{zetaSigmaHat1} and \eqref{zetaSigmaHat2}, if $\tilde{\bm{\Sigma}}=\hat{\bm{\Sigma}}$. The error DE with respect to the random projection is then given by \eqref{ultimateDE} of Lemma 2.}

\textbf{Proof:} See Appendix \ref{firstDE}

By comparing the asymptotic expressions of the class-conditional discriminant statistics in Theorem 1 to their exact expressions \eqref{m0}, \eqref{m1} and \eqref{sigma}, it follows that the effect of the infinite ensemble projection is a regularization of $\tilde{\bm{\Sigma}}$ by the quantity $\frac{1}{\zeta_{\tilde{\bm{\Sigma}}}\left(\bm{\Sigma}\right)}$. Since $\zeta_{\tilde{\bm{\Sigma}}}\left(\bm{\Sigma}\right)$ is a function of $d$ through either \eqref{zetaSigma} and \eqref{zetaSigma2} or \eqref{zetaSigmaHat1} and \eqref{zetaSigmaHat2}, depending on whether $\tilde{\bm{\Sigma}}=\bm{\Sigma}$ or $\tilde{\bm{\Sigma}}=\hat{\bm{\Sigma}}$ respectively, the effect of $d$ is to control this regularization parameter. Specifically when $\tilde{\bm{\Sigma}}=\hat{\bm{\Sigma}}$, this result is consistent with the work of \cite{durrant2013random}, which shows that the condition number of the covariance estimate in the projected space, $\hat{\bm{\Sigma}}^{\text{RP}}$, is bounded, and that the bound is a function of the projection dimension, that is, the projection dimension acts as a regularization parameter. The advantage of expressing this effect as in Theorem 1 is that the DEs exhibit the same form as the exact (non-asymptotic) RLDA class-conditional discriminant statistics with regularization parameter set to $\frac{1}{\zeta_{\tilde{\bm{\Sigma}}}\left(\bm{\Sigma}\right)}$. From this, we deduce that though an RP-LDA infinite ensemble classifier may be more computationally efficient than an RLDA classifier due to working with data of reduced dimension, it can never surpass the accuracy of an RLDA classifier for which the regularization parameter has been properly tuned; the possible values to which $d$ can be set restrict the possible values of $\frac{1}{\zeta_{\tilde{\bm{\Sigma}}}\left(\bm{\Sigma}\right)}$ to a subset of $(0,\infty)$, whereas an RLDA classifier's regularization parameter can vary over all of $(0,\infty)$.

Now, in a similar fashion, we present the DEs for different combinations of known and unknown statistics.

\textbf{Theorem 2}{ \textit{(Known means and covariance)}
\textit{ Let $\zeta_{\bm{\Sigma}}\left(\bm{\Sigma}\right)$ be as defined by \eqref{zetaSigma} and \eqref{zetaSigma2}. When the class-conditional data distribution means $\bm{\mu}_0$, $\bm{\mu}_1$ and covariance $\bm{\Sigma}$ are known, the DEs of the class-conditional discriminant statistics $ m_0({\bm{\mu}}_0,{\bm{\mu}}_1,{\bm{\Sigma}},\tilde{\pi}_0,\tilde{\pi}_1)$, $ m_1({\bm{\mu}}_0,{\bm{\mu}}_1,{\bm{\Sigma}},\tilde{\pi}_0,\tilde{\pi}_1)$, and $ \sigma^2({\bm{\mu}}_0,{\bm{\mu}}_1,{\bm{\Sigma}})$  are given by}
\begin{align*}
     \bar{m}_{0,{\bm{\mu}}_0,{\bm{\mu}}_1,{\bm{\Sigma}}}&=-\frac{1}{2}({\bm{\mu}}_1-{\bm{\mu}}_0)^T\left({\bm{\Sigma}}+\frac{1}{\zeta_{\bm{\Sigma}}\left(\bm{\Sigma}\right)}\textbf{I}_p\right)^{-1}\left({\bm{\mu}}_1-{\bm{\mu}}_0\right)+\text{ln}\frac{\pi_1}{\pi_0}\\
    \bar{m}_{1,{\bm{\mu}}_0,{\bm{\mu}}_1,{\bm{\Sigma}}}&=\frac{1}{2}({\bm{\mu}}_1-{\bm{\mu}}_0)^T\left({\bm{\Sigma}}+\frac{1}{\zeta_{\bm{\Sigma}}\left(\bm{\Sigma}\right)}\textbf{I}_p\right)^{-1}\left({\bm{\mu}}_1-{\bm{\mu}}_0\right)+\text{ln}\frac{\pi_1}{\pi_0}\\
     \bar{\sigma}^2_{{\bm{\mu}}_0,{\bm{\mu}}_1,{\bm{\Sigma}}}&=({\bm{\mu}}_1-{\bm{\mu}}_0)^T\left({\bm{\Sigma}}+\frac{1}{\zeta_{\bm{\Sigma}}\left(\bm{\Sigma}\right)}\textbf{I}_p\right)^{-1}\bm{\Sigma}\left({\bm{\Sigma}}+\frac{1}{\zeta_{\bm{\Sigma}}\left(\bm{\Sigma}\right)}\textbf{I}_p\right)^{-1}({\bm{\mu}}_1-{\bm{\mu}}_0).
 \end{align*}
 \textit{The error DE is then given by $\bar{\varepsilon}_{{\bm{\mu}}_0,{\bm{\mu}}_1,{\bm{\Sigma}}}$ as defined in \eqref{ultimateDE} of Lemma 2.}}

\textbf{Proof:} See Appendix \ref{appA1}.

\textbf{Corollary 1}\textit{ Following from the setting of Theorem 2 where the class-conditional data distribution statistics are known, and additionally assuming equal priors $\pi_0=\pi_1$ and covariance $\bm{\Sigma}=\textbf{I}_p$, the error DE is given by}
\begin{equation*}
    \bar{\varepsilon}_{{\bm{\mu}}_0,{\bm{\mu}}_1,{\bm{\Sigma}}}=\Phi\left(-\frac{\Vert\bm{\mu}_0-\bm{\mu}_1\Vert_2}{2}\right).
\end{equation*}

The asymptotic misclassification probability of the RP-LDA infinite ensemble in Corollary 1 is exactly equal to that of LDA with known statistics under identical conditions (see \cite{wang2018dimension}). For comparison, the asymptotic misclassification probability of an RP-LDA classifier with known statistics, that is, with decision rule ${h}_{\text{RP-LDA}}(\textbf{x},\bm{\mu}_0,\bm{\mu}_1,\bm{\Sigma},\pi_0,\pi_1)$, operating under identical conditions is given by $\Phi\left(-\frac{\Vert\bm{\mu}_0-\bm{\mu}_1\Vert_2}{2}\sqrt{d/p}\right)$ (see \cite{8683386}), which indicates an increasing error with decreasing $d$. When $d=p$, the asymptotic misclassification probability is the same as for LDA operating in the full data space. Thus asymptotically, while a single projection incurs a loss in performance due to dimensionality reduction compared to LDA when $p<n$ in this setting, an infinite ensemble of projections incurs no such loss. Additionally, random projection is more computationally efficient than LDA due to working with reduced dimensions. As mentioned previously, when $p>n$, direct application of LDA is not possible anyway.

\textbf{Theorem 3}\textit{ (Unknown means and known covariance)}
\textit{Let $\zeta_{\bm{\Sigma}}\left(\bm{\Sigma}\right)$ be as defined by \eqref{zetaSigma} and \eqref{zetaSigma2}. When the class-conditional data distribution means $\bm{\mu}_0$, $\bm{\mu}_1$ are unknown and the covariance $\bm{\Sigma}$ is known, the DEs of the class-conditional discriminant statistics} $m_0(\hat{\bm{\mu}}_0,\hat{\bm{\mu}}_1,{\bm{\Sigma}},\tilde{\pi}_0,\tilde{\pi}_1)$, $m_1(\hat{\bm{\mu}}_0,\hat{\bm{\mu}}_1,{\bm{\Sigma}},\tilde{\pi}_0,\tilde{\pi}_1)$, \textit{and} $\sigma^2(\hat{\bm{\mu}}_0,\hat{\bm{\mu}}_1,{\bm{\Sigma}})$ \textit{are given by}
\begin{align}
     \bar{m}_{0,\hat{\bm{\mu}}_0,\hat{\bm{\mu}}_1,{\bm{\Sigma}}}&=-\frac{1}{2}\left(\bm{\mu}_1-\bm{\mu}_0\right)^T\left({\bm{\Sigma}}+\frac{1}{\zeta_{\bm{\Sigma}}\left(\bm{\Sigma}\right)}\textbf{I}_p\right)^{-1}\left(\bm{\mu}_1-\bm{\mu}_0\right)\\
     &\hspace{30pt}+\frac{1}{2}\left(\frac{1}{n_0}-\frac{1}{n_1}\right)\text{tr}\left\{\bm{\Sigma}\left({\bm{\Sigma}}+\frac{1}{\zeta_{\bm{\Sigma}}\left(\bm{\Sigma}\right)}\textbf{I}_p\right)^{-1}\right\}+\text{ln}\frac{\pi_1}{\pi_0}\\
    \bar{m}_{1,\hat{\bm{\mu}}_0,\hat{\bm{\mu}}_1,{\bm{\Sigma}}}&=\frac{1}{2}\left(\bm{\mu}_1-\bm{\mu}_0\right)^T\left({\bm{\Sigma}}+\frac{1}{\zeta_{\bm{\Sigma}}\left(\bm{\Sigma}\right)}\textbf{I}_p\right)^{-1}\left(\bm{\mu}_1-\bm{\mu}_0\right)\\
     &\hspace{30pt}+\frac{1}{2}\left(\frac{1}{n_0}-\frac{1}{n_1}\right)\text{tr}\left\{\bm{\Sigma}\left({\bm{\Sigma}}+\frac{1}{\zeta_{\bm{\Sigma}}\left(\bm{\Sigma}\right)}\textbf{I}_p\right)^{-1}\right\}+\text{ln}\frac{\pi_1}{\pi_0}\\
     \bar{\sigma}^2_{\hat{\bm{\mu}}_0,\hat{\bm{\mu}}_1,{\bm{\Sigma}}}&=\left(\bm{\mu}_1-\bm{\mu}_0\right)^T\left({\bm{\Sigma}}+\frac{1}{\zeta_{\bm{\Sigma}}\left(\bm{\Sigma}\right)}\textbf{I}_p\right)^{-1}\bm{\Sigma}\left({\bm{\Sigma}}+\frac{1}{\zeta_{\bm{\Sigma}}\left(\bm{\Sigma}\right)}\textbf{I}_p\right)^{-1}\left(\bm{\mu}_1-\bm{\mu}_0\right)\\
     &\hspace{30pt}+\left(\frac{1}{n_0}+\frac{1}{n_1}\right)\text{tr}\left\{\bm{\Sigma}\left({\bm{\Sigma}}+\frac{1}{\zeta_{\bm{\Sigma}}\left(\bm{\Sigma}\right)}\textbf{I}_p\right)^{-1}\bm{\Sigma}\left({\bm{\Sigma}}+\frac{1}{\zeta_{\bm{\Sigma}}\left(\bm{\Sigma}\right)}\textbf{I}_p\right)^{-1}\right\}
 \end{align}
 \textit{The error DE is then given by $\bar{\varepsilon}_{\hat{\bm{\mu}}_0,\hat{\bm{\mu}}_1,{\bm{\Sigma}}}$ as defined in \eqref{ultimateDE} of Lemma 2.}

\textbf{Proof:} See Appendix \ref{appA2}.

\textbf{Corollary 2}\textit{ Following from the setting of Theorem 3 where the class-conditional data distribution means are unknown and covariance is known, and additionally assuming equal priors $\pi_0=\pi_1$ and covariance $\bm{\Sigma}=\textbf{I}_p$, the error DE is given by}
\begin{equation}
    \bar{\varepsilon}_{\hat{\bm{\mu}}_0,\hat{\bm{\mu}}_1,{\bm{\Sigma}}}=\frac{1}{2}\Phi\left(-\frac{1}{2}\frac{\left\Vert\bm{\mu}_0-\bm{\mu}_1\right\Vert_2^2+\frac{p}{n_1}-\frac{p}{n_0}}{\sqrt{\left\Vert\bm{\mu}_0-\bm{\mu}_1\right\Vert_2^2+\frac{p}{n_0}+\frac{p}{n_1}}}\right)+\frac{1}{2}\Phi\left(-\frac{1}{2}\frac{\left\Vert\bm{\mu}_0-\bm{\mu}_1\right\Vert_2^2+\frac{p}{n_0}-\frac{p}{n_1}}{\sqrt{\left\Vert\bm{\mu}_0-\bm{\mu}_1\right\Vert_2^2+\frac{p}{n_0}+\frac{p}{n_1}}}\right).\label{unknownMEans}
\end{equation}

The expression for the asymptotic misclassification probability of the RP-LDA infinite ensemble with unknown means and known covariance in Corollary 2 is exactly the same as for LDA when $p<n$ under identical conditions (see \cite{wang2018dimension}). The advantage of the RP-LDA infinite ensemble over LDA is that it is more computationally efficient while being applicable to the regime $p
>n$.

Before presenting Theorems 4 and 5, we define a few more quantities.
Let
\begin{equation}\kappa=\left(1-\frac{p}{n}\frac{\zeta_{\hat{\bm{\Sigma}}}^2\left(\bm{\Sigma}\right)\frac{1}{p}\text{tr}\left\{{\bm{\Sigma}}\left(\tilde{e}\bm{\Sigma}+\textbf{I}_p\right)^{-1}\bm{\Sigma}\left(\tilde{e}\bm{\Sigma}+\textbf{I}_p\right)^{-1}\right\}}{(1+{e})^2}\right)^{-1}\label{solvekappa}
\end{equation}
and let $\tilde{e}$ be the quantity obtained by solving the following system of equations in $e$ and $\tilde{e}$
 \begin{align}
    &e=\zeta_{\hat{\bm{\Sigma}}}\left(\bm{\Sigma}\right)\frac{1}{n}\text{tr}\left\{\bm{\Sigma}\left(\tilde{e}\bm{\Sigma}+\textbf{I}_p\right)^{-1}\right\}\\
    &\tilde{e}=\frac{\zeta_{\hat{\bm{\Sigma}}}\left(\bm{\Sigma}\right)}{1+e}\label{systemE}
\end{align}
We are now ready to present Theorems 4 and 5 which assume unknown covariance.

\textbf{Theorem 4}\textit{ (Known means and unknown covariance)}
\textit{Let $\zeta_{\hat{\bm{\Sigma}}}\left(\bm{\Sigma}\right)$ be as defined by \eqref{zetaSigmaHat1} and \eqref{zetaSigmaHat2}, $\kappa$ be as defined by \eqref{solvekappa}, and $\tilde{e}$ be as defined by the system \eqref{systemE}. When the class-conditional data distribution means $\bm{\mu}_0$, $\bm{\mu}_1$ are known and the covariance $\bm{\Sigma}$ is unknown, the DEs of the class-conditional discriminant statistics $ m_0({\bm{\mu}}_0,{\bm{\mu}}_1,\hat{\bm{\Sigma}},\tilde{\pi}_0,\tilde{\pi}_1)$, $ m_1({\bm{\mu}}_0,{\bm{\mu}}_1,\hat{\bm{\Sigma}},\tilde{\pi}_0,\tilde{\pi}_1)$, and $ \sigma^2({\bm{\mu}}_0,{\bm{\mu}}_1,\hat{\bm{\Sigma}})$  are given by}
\begin{align*}
     \bar{m}_{0,{\bm{\mu}}_0,{\bm{\mu}}_1,\hat{\bm{\Sigma}}}&=-\frac{1}{2}\zeta_{\hat{\bm{\Sigma}}}\left(\bm{\Sigma}\right)\left(\bm{\mu}_1-\bm{\mu}_0\right)^T\left(\tilde{e}\bm{\Sigma}+\textbf{I}_p\right)^{-1}\left(\bm{\mu}_1-\bm{\mu}_0\right)+\text{ln}\frac{\pi_1}{\pi_0}\\
    \bar{m}_{1,{\bm{\mu}}_0,{\bm{\mu}}_1,\hat{\bm{\Sigma}}}&=\frac{1}{2}\zeta_{\hat{\bm{\Sigma}}}\left(\bm{\Sigma}\right)\left(\bm{\mu}_1-\bm{\mu}_0\right)^T\left(\tilde{e}\bm{\Sigma}+\textbf{I}_p\right)^{-1}\left(\bm{\mu}_1-\bm{\mu}_0\right)+\text{ln}\frac{\pi_1}{\pi_0}\\
     \bar{\sigma}^2_{{\bm{\mu}}_0,{\bm{\mu}}_1,\hat{\bm{\Sigma}}}&=\kappa\zeta_{\hat{\bm{\Sigma}}}^2\left(\bm{\Sigma}\right)\left(\bm{\mu}_1-\bm{\mu}_0\right)^T\left(\tilde{e}\bm{\Sigma}+\textbf{I}_p\right)^{-1}\bm{\Sigma}\left(\tilde{e}\bm{\Sigma}+\textbf{I}_p\right)^{-1}\left(\bm{\mu}_1-\bm{\mu}_0\right).
 \end{align*}
 \textit{The error DE is then given by $\bar{\varepsilon}_{{\bm{\mu}}_0,{\bm{\mu}}_1,\hat{\bm{\Sigma}}}$ as defined in \eqref{ultimateDE} of Lemma 2.}

\textbf{Proof:} See Appendix \ref{appA3}.

\textbf{Corollary 3}\textit{ Following from the setting of Theorem 4 where the class-conditional data distribution means are known and the covariance is unknown, and additionally assuming equal priors $\pi_0=\pi_1$ and covariance $\bm{\Sigma}=\textbf{I}_p$, the error DE is given by}
\begin{equation*}
    \bar{\varepsilon}_{{\bm{\mu}}_0,{\bm{\mu}}_1,\tilde{\bm{\Sigma}}}=\Phi\left(-\frac{\Vert\bm{\mu}_0-\bm{\mu}_1\Vert_2}{2}\sqrt{1-\frac{d^2}{np}}\right).
\end{equation*}

When the covariance is unknown, the random projection ensemble introduces a multiplicative factor $\sqrt{1-\frac{d^2}{np}}$ which is a function of $d$. For comparison, the asymptotic misclassification probability of LDA when $p<n$ under an identical setting is $\Phi\left(-\frac{\Vert\bm{\mu}_0-\bm{\mu}_1\Vert_2}{2}\sqrt{1-\frac{p}{n}}\right)$ (see  \cite{wang2018dimension}). While $p$ and $n$ are constrained by the nature of the data available, $d$ can be tuned so that the misclassification probability is minimized. Corollary 3 suggests that, by choosing $d$ as small as possible, we can approach the asymptotic misclassification probability of LDA with complete knowledge of the statistics. This is at reduced computational cost and feasible even when $p>n$ in contrast to classical LDA.

\textbf{Theorem 5} \textit{(Unknown means and unknown covariance)}
\textit{Let $\zeta_{\hat{\bm{\Sigma}}}\left(\bm{\Sigma}\right)$ be as defined by \eqref{zetaSigmaHat1} and \eqref{zetaSigmaHat2}, $\kappa$ be as defined by \eqref{solvekappa}, and $\tilde{e}$ be as defined by the system \eqref{systemE}. When the class-conditional data distribution means $\bm{\mu}_0$, $\bm{\mu}_1$ and the covariance $\bm{\Sigma}$ are all unknown, the DEs of the class-conditional discriminant statistics $ m_0(\hat{\bm{\mu}}_0,\hat{\bm{\mu}}_1,\hat{\bm{\Sigma}},\tilde{\pi}_0,\tilde{\pi}_1)$, $ m_1(\hat{\bm{\mu}}_0,\hat{\bm{\mu}}_1,\hat{\bm{\Sigma}},\tilde{\pi}_0,\tilde{\pi}_1)$, and $ \sigma^2(\hat{\bm{\mu}}_0,\hat{\bm{\mu}}_1,\hat{\bm{\Sigma}})$  are given by}
\begin{align*}
     \bar{m}_{0,\hat{\bm{\mu}}_0,\hat{\bm{\mu}}_1,\hat{\bm{\Sigma}}}&=-\frac{1}{2}\zeta_{\hat{\bm{\Sigma}}}\left(\bm{\Sigma}\right)\left(\bm{\mu}_1-\bm{\mu}_0\right)^T\left(\tilde{e}\bm{\Sigma}+\textbf{I}_p\right)^{-1}\left(\bm{\mu}_1-\bm{\mu}_0\right)\\
            &\hspace{30pt}+\frac{1}{2}\zeta_{\hat{\bm{\Sigma}}}\left(\bm{\Sigma}\right)\left(\frac{1}{n_0}-\frac{1}{n_1}\right)\text{tr}\left\{\bm{\Sigma}\left(\tilde{e}\bm{\Sigma}+\textbf{I}_p\right)^{-1}\right\}+\text{ln}\frac{\pi_1}{\pi_0}\\
    \bar{m}_{1,\hat{\bm{\mu}}_0,\hat{\bm{\mu}}_1,\hat{\bm{\Sigma}}}&=\frac{1}{2}\zeta_{\hat{\bm{\Sigma}}}\left(\bm{\Sigma}\right)\left(\bm{\mu}_1-\bm{\mu}_0\right)^T\left(\tilde{e}\bm{\Sigma}+\textbf{I}_p\right)^{-1}\left(\bm{\mu}_1-\bm{\mu}_0\right)\\
    &\hspace{30pt}+\frac{1}{2}\zeta_{\hat{\bm{\Sigma}}}\left(\bm{\Sigma}\right)\left(\frac{1}{n_0}-\frac{1}{n_1}\right)\text{tr}\left\{\bm{\Sigma}\left(\tilde{e}\bm{\Sigma}+\textbf{I}_p\right)^{-1}\right\}+\text{ln}\frac{\pi_1}{\pi_0}\\
     \bar{\sigma}^2_{\hat{\bm{\mu}}_0,\hat{\bm{\mu}}_1,\hat{\bm{\Sigma}}}&=\kappa\zeta_{\hat{\bm{\Sigma}}}^2\left(\bm{\Sigma}\right)\left(\bm{\mu}_1-\bm{\mu}_0\right)^T\left(\tilde{e}\bm{\Sigma}+\textbf{I}_p\right)^{-1}\bm{\Sigma}\left(\tilde{e}\bm{\Sigma}+\textbf{I}_p\right)^{-1}\left(\bm{\mu}_1-\bm{\mu}_0\right)\\
       &\hspace{30pt}+\kappa\zeta_{\hat{\bm{\Sigma}}}^2\left(\bm{\Sigma}\right)\left(\frac{1}{n_0}+\frac{1}{n_1}\right)\text{tr}\left\{\bm{\Sigma}\left(\tilde{e}\bm{\Sigma}+\textbf{I}_p\right)^{-1}\bm{\Sigma}\left(\tilde{e}\bm{\Sigma}+\textbf{I}_p\right)^{-1}\right\}.
 \end{align*}
 \textit{The error DE is then given by $\bar{\varepsilon}_{\hat{\bm{\mu}}_0,\hat{\bm{\mu}}_1,\hat{\bm{\Sigma}}}$ as defined in \eqref{ultimateDE} of Lemma 2.}

\textbf{Proof:} See Appendix \ref{appA4}.

\textbf{Corollary 4}\textit{ Following from the setting of Theorem 5 where the class-conditional data distribution means and covariance are unknown, and additionally assuming equal priors $\pi_0=\pi_1$ and covariance $\bm{\Sigma}=\textbf{I}_p$, the error DE is given by}
\begin{eqnarray}
    \bar{\varepsilon}_{\hat{\bm{\mu}}_0,\hat{\bm{\mu}}_1,\hat{\bm{\Sigma}}}&=\frac{1}{2}\Phi\left(-\frac{1}{2}\frac{\left\Vert\bm{\mu}_0-\bm{\mu}_1\right\Vert_2^2+\frac{p}{n_1}-\frac{p}{n_0}}{\sqrt{\left\Vert\bm{\mu}_0-\bm{\mu}_1\right\Vert_2^2+\frac{p}{n_0}+\frac{p}{n_1}}}\sqrt{1-\frac{d^2}{np}}\right)+\frac{1}{2}\Phi\left(-\frac{1}{2}\frac{\left\Vert\bm{\mu}_0-\bm{\mu}_1\right\Vert_2^2+\frac{p}{n_0}-\frac{p}{n_1}}{\sqrt{\left\Vert\bm{\mu}_0-\bm{\mu}_1\right\Vert_2^2+\frac{p}{n_0}+\frac{p}{n_1}}}\sqrt{1-\frac{d^2}{np}}\right).
\end{eqnarray}

For comparison, the asymptotic misclassification probability of LDA when $p<n$ under an identical setting is $$\frac{1}{2}\Phi\left(-\frac{1}{2}\frac{\left\Vert\bm{\mu}_0-\bm{\mu}_1\right\Vert_2^2+\frac{p}{n_1}-\frac{p}{n_0}}{\sqrt{\left\Vert\bm{\mu}_0-\bm{\mu}_1\right\Vert_2^2+\frac{p}{n_0}+\frac{p}{n_1}}}\sqrt{1-\frac{p}{n}}\right)+\frac{1}{2}\Phi\left(-\frac{1}{2}\frac{\left\Vert\bm{\mu}_0-\bm{\mu}_1\right\Vert_2^2+\frac{p}{n_0}-\frac{p}{n_1}}{\sqrt{\left\Vert\bm{\mu}_0-\bm{\mu}_1\right\Vert_2^2+\frac{p}{n_0}+\frac{p}{n_1}}}\sqrt{1-\frac{p}{n}}\right)$$ (see \cite{wang2018dimension}). Corollary 4 suggests that, as $d$ gets small, the misclassification probability approaches that of LDA with unknown means and known covariance given in \eqref{unknownMEans}.

\begin{figure}[ht]
    \begin{center}
        \includegraphics[width=0.8\linewidth]{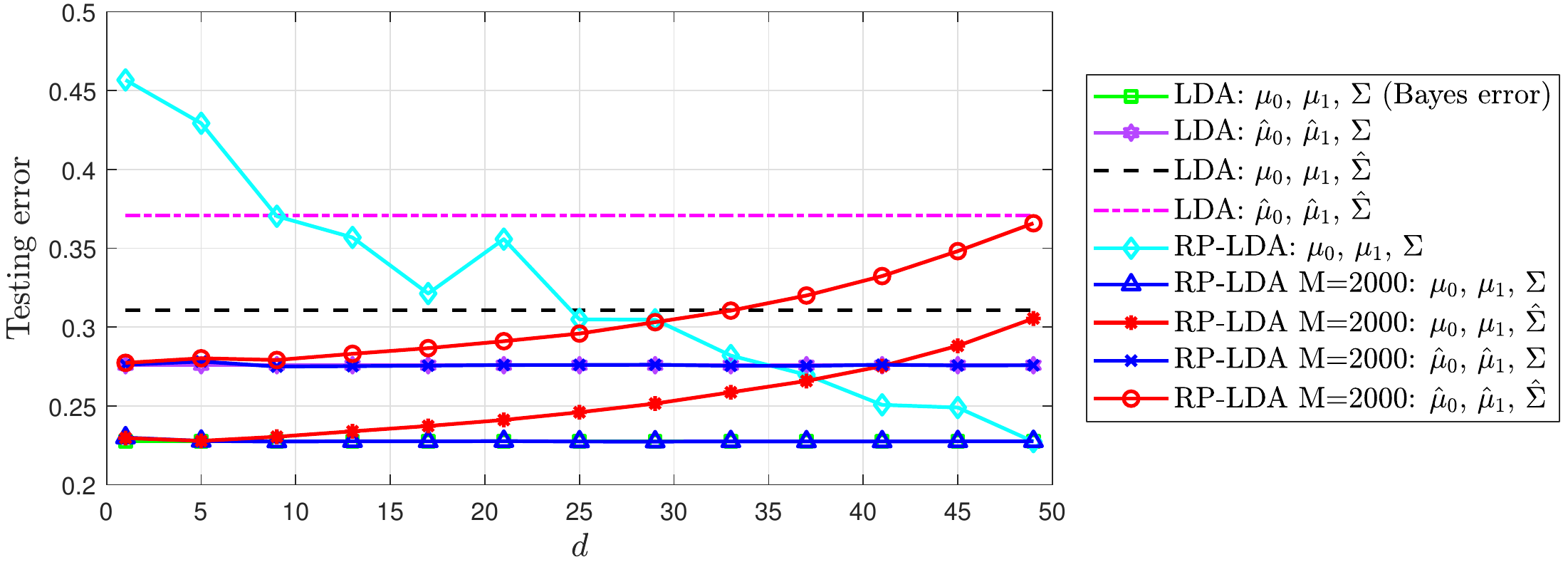}
    \end{center}
     \caption{A plot of the empirical errors over $10^5$ testing points of several classifiers under different combinations of known and unknown statistics. Here $p=50$ and $n=100$.}
     \label{fig3}
\end{figure}

To conclude this section, we note that the asymptotic effect of the random projection ensemble is to regularize the covariance parameter $\tilde{\bm{\Sigma}}$. Additionally, under the assumption of equal priors and that $\bm{\Sigma}=\textbf{I}_p$, the RP-LDA infinite ensemble asymptotically coincides with LDA for $p<n$ in two cases: when all statistics are known or when the means are unknown but the covariance is known. When the means are known and the covariance is unknown, the DE suggests that the RP-LDA ensemble error can asymptotically approach the error of LDA with known statistics as $d$ gets smaller. When both the means and the covariance are unknown, the DE suggests that the RP-LDA ensemble can approach LDA with unknown means and known covariance as $d$ gets smaller. These findings are demonstrated in the finite scenario by the simulation results illustrated in Figure \ref{fig3}. For this simulation, we generate a set of $100$ training samples with dimensionality $p=50$ having the class-conditional distribution statistics
\begin{equation}
\mu_0=\frac{1}{p^{1/4}}\left[\textbf{1}^T_{\lceil{\sqrt{p}}\rceil} \ \textbf{0}^T_{p-\lceil{\sqrt{p}}\rceil-2} \ 2 \ 2\right]^T, \ \mu_1=\textbf{0}_p, \text{ and }
    \bm{\Sigma}=\textbf{I}_p.
\end{equation}
We train LDA and the RP-LDA ensemble with $M=2000$ under different combinations of known and unknown means and covariance. We then evaluate the testing errors for each of these classifiers over an independently generated set of $10^{5}$ data samples as the projection dimension $d$ varies from 1 to $\text{rank}(\hat{\bm{\Sigma}})-2$.

The classifiers behave as predicted by the asymptotic probabilities: LDA with known statistics (which coincides with the Bayes error in this case) matches RP-LDA $M=2000$ with known statistics, RP-LDA $M=2000$ with unknown means matches LDA with unknown means, RP-LDA $M=2000$ with unknown covariance tends to LDA with known statistics, and RP-LDA $M=2000$ with unknown statistics tends to LDA with unknown means.

\subsubsection{Generalized Consistent Estimator of Error}\label{gest}
In this section, we derive a generalized consistent estimator $\hat{\varepsilon}$ of the error $\varepsilon(\hat{\bm{\mu}}_0,\hat{\bm{\mu}}_1,\hat{\bm{\Sigma}},\tilde{\pi}_0,\tilde{\pi}_1)$ of the RP-LDA infinite ensemble classifier under unknown statistics $\bm{\mu}_0$, $\bm{\mu}_1$, and $\bm{\Sigma}$. The conventional plugin estimator of $\varepsilon(\hat{\bm{\mu}}_0,\hat{\bm{\mu}}_1,\hat{\bm{\Sigma}},\tilde{\pi}_0,\tilde{\pi}_1)$ is not consistent in the regime where $n$, $p$, and $d$ grow at constant rates to each other. On the other hand, the G-estimator is by definition a function $\hat{\varepsilon}$ of $\hat{\bm{\mu}}_0$, $\hat{\bm{\mu}}_1$, and $\hat{\bm{\Sigma}}$ such that
\begin{equation}
     \hat{\varepsilon}-\varepsilon(\hat{\bm{\mu}}_0,\hat{\bm{\mu}}_1,\hat{\bm{\Sigma}},\tilde{\pi}_0,\tilde{\pi}_1)\xrightarrow{\text{a.s.}}0\label{finalerror}
\end{equation}
as $n$, $p$, and $d$ grow under the growth regime defined by conditions (a)-(e) in Section \ref{DEs}. It also satisfies
\begin{equation}
     \hat{\varepsilon}- \mathbb{E}_{\hat{\bm{\mu}}_0,\hat{\bm{\mu}}_1,\hat{\bm{\Sigma}},\tilde{\pi}_0,\tilde{\pi}_1}\left[\varepsilon(\hat{\bm{\mu}}_0,\hat{\bm{\mu}}_1,\hat{\bm{\Sigma}},\tilde{\pi}_0,\tilde{\pi}_1)\right]\xrightarrow{\text{a.s.}}0,\label{epeest}
\end{equation}
and so $\hat{\varepsilon}$ doubles as a G-estimator of the generalization error.

Similar to how we constructed the deterministic equivalent of the error out of individual deterministic equivalents of the class-conditional discriminant statistics (presented in Lemma 1), we can construct a G-estimator $\hat{\varepsilon}$ of the error out of individual G-estimators $\hat{m}_0$, $\hat{m}_1$, and $\hat{\sigma}^2$ of the class-conditional discriminant statistics. This is formally stated in Lemma 3.

\textbf{Lemma 3}
\textit{Let $\hat{m}_0$, $\hat{m}_1$ and $\hat{\sigma}^2$ be functions of $\hat{\bm{\mu}}_0$, $\hat{\bm{\mu}}_1$, and $\hat{\bm{\Sigma}}$ such that}
  \begin{align*}
     \hat{m}_0- m_0(\hat{\bm{\mu}}_0,\hat{\bm{\mu}}_1,\hat{\bm{\Sigma}},\tilde{\pi}_0,\tilde{\pi}_1)&\xrightarrow{\text{a.s.}}0\nonumber\\
     \hat{m}_1- m_1(\hat{\bm{\mu}}_0,\hat{\bm{\mu}}_1,\hat{\bm{\Sigma}},\tilde{\pi}_0,\tilde{\pi}_1)&\xrightarrow{\text{a.s.}}0\nonumber\\
     \hat{\sigma}^2-\sigma^2(\hat{\bm{\mu}}_0,\hat{\bm{\mu}}_1,\hat{\bm{\Sigma}})&\xrightarrow{\text{a.s.}}0
     \label{conv11}
 \end{align*}
 \textit{for $n$, $p$, and $d$ growing subject to (a)-(g). Then by the continuous mapping theorem and other properties of almost-sure convergence, we have that \eqref{finalerror} and \eqref{epeest} hold with}
\begin{equation*}
\hat{\varepsilon}={\hat{\pi}}_0\Phi\left(\frac{\hat{m}_0}{\sqrt{\hat{\sigma}^2}}\right)+{\hat{\pi}}_1\Phi\left(\frac{-\hat{m}_1}{\sqrt{\hat{\sigma}^2}}\right).
\label{yessme}
\end{equation*}

Thus, the problem breaks down into deriving the generalized consistent estimators $\hat{m}_0$, $\hat{m}_1$, and $\hat{\sigma}^2$. These are presented in Theorem 6. Each G-estimator can be constructed out of the corresponding intermediate convergence result presented in Theorem 1. This is fully detailed in the appendix.

{\bf Theorem 6} \textit{Let $\zeta_{\hat{\bm{\Sigma}}}(\hat{\bm{\Sigma}})$ be the root of the monotonically decreasing function}
\begin{equation*}
    f(x)=1-\frac{1}{d}\text{tr}\left\{\hat{\bm{\Sigma}}\left(\hat{\bm{\Sigma}}+\frac{1}{x}\textbf{I}_p\right)^{-1}\right\}
\end{equation*}
\textit{over $x>0$. When the class-conditional data distribution means $\bm{\mu}_0$, $\bm{\mu}_1$ and the covariance $\bm{\Sigma}$ are all unknown, the G-estimators of the class-conditional discriminant statistics $ m_0(\hat{\bm{\mu}}_0,\hat{\bm{\mu}}_1,\hat{\bm{\Sigma}},\tilde{\pi}_0,\tilde{\pi}_1)$, $ m_1(\hat{\bm{\mu}}_0,\hat{\bm{\mu}}_1,\hat{\bm{\Sigma}},\tilde{\pi}_0,\tilde{\pi}_1)$, and $ \sigma^2(\hat{\bm{\mu}}_0,\hat{\bm{\mu}}_1,\hat{\bm{\Sigma}})$  are given by}
\begin{align*}
     \hat{m}_{0}&=-\frac{1}{2}(\hat{\bm{\mu}}_1-\hat{\bm{\mu}}_0)^T\left(\hat{\bm{\Sigma}}+\frac{1}{\zeta_{\hat{\bm{\Sigma}}}(\hat{\bm{\Sigma}})}\textbf{I}_p\right)^{-1}(\hat{\bm{\mu}}_1-\hat{\bm{\mu}}_0)+\frac{\frac{1}{n_0}\text{tr}\left\{\hat{\bm{\Sigma}}\left(\hat{\bm{\Sigma}}+\frac{1}{\zeta_{\hat{\bm{\Sigma}}}(\hat{\bm{\Sigma}})}\textbf{I}_p\right)^{-1}\right\}}{1-\frac{1}{n}\text{tr}\left\{\hat{\bm{\Sigma}}\left(\hat{\bm{\Sigma}}+\frac{1}{\zeta_{\hat{\bm{\Sigma}}}(\hat{\bm{\Sigma}})}\textbf{I}_p\right)^{-1}\right\}}+\text{ln}\frac{\tilde{\pi}_1}{\tilde{\pi}_0}\\
    \hat{m}_{1}&=\frac{1}{2}(\hat{\bm{\mu}}_1-\hat{\bm{\mu}}_0)^T\left(\hat{\bm{\Sigma}}+\frac{1}{\zeta_{\hat{\bm{\Sigma}}}(\hat{\bm{\Sigma}})}\textbf{I}_p\right)^{-1}(\hat{\bm{\mu}}_1-\hat{\bm{\mu}}_0)-\frac{\frac{1}{n_1}\text{tr}\left\{\hat{\bm{\Sigma}}\left(\hat{\bm{\Sigma}}+\frac{1}{\zeta_{\hat{\bm{\Sigma}}}(\hat{\bm{\Sigma}})}\textbf{I}_p\right)^{-1}\right\}}{1-\frac{1}{n}\text{tr}\left\{\hat{\bm{\Sigma}}\left(\hat{\bm{\Sigma}}+\frac{1}{\zeta_{\hat{\bm{\Sigma}}}(\hat{\bm{\Sigma}})}\textbf{I}_p\right)^{-1}\right\}}+\text{ln}\frac{\tilde{\pi}_1}{\tilde{\pi}_0}\\
     \hat{\sigma}^2&=\left(1+\frac{\frac{1}{n}\text{tr}\left\{\hat{\bm{\Sigma}}\left(\hat{\bm{\Sigma}}+\frac{1}{\zeta_{\hat{\bm{\Sigma}}}(\hat{\bm{\Sigma}})}\textbf{I}_p\right)^{-1}\right\}}{1-\frac{1}{n}\text{tr}\left\{\hat{\bm{\Sigma}}\left(\hat{\bm{\Sigma}}+\frac{1}{\zeta_{\hat{\bm{\Sigma}}}(\hat{\bm{\Sigma}})}\textbf{I}_p\right)^{-1}\right\}}\right)^2(\hat{\bm{\mu}}_1-\hat{\bm{\mu}}_0)^T\left(\hat{\bm{\Sigma}}+\frac{1}{\zeta_{\hat{\bm{\Sigma}}}(\hat{\bm{\Sigma}})}\textbf{I}_p\right)^{-1}\hat{\bm{\Sigma}}\left(\hat{\bm{\Sigma}}+\frac{1}{\zeta_{\hat{\bm{\Sigma}}}(\hat{\bm{\Sigma}})}\textbf{I}_p\right)^{-1}(\hat{\bm{\mu}}_1-\hat{\bm{\mu}}_0).
 \end{align*}
 \textit{The G-estimator of error is then given by $\hat{\varepsilon}$ as defined in Lemma 3.}

\textbf{Proof:} See Appendix \ref{app:appB}

\section{Tuning the Projection Dimension}\label{section_sims}
The previous section derives a G-estimator of the RP-LDA infinite ensemble error. In this section, we show that the G-estimator can be as good an estimate of the error of a finite ensemble as the conventional hold-out/testing error or cross-validation estimates when varying the projection dimension $d$. This is demonstrated on synthetic data, generated in conformity with the Gaussian assumptions, in Section \ref{synthetic}, as well as on several real datasets, which do not necessarily conform to these assumptions, in Section \ref{real}.
\subsection{Synthetic Data}\label{synthetic}
In this section, we generate all data points synthetically according to the distribution specified by \eqref{dist}. In addition, as in \cite{zollanvari2015generalized}, we employ stratified sampling, meaning that the data points making up the two classes $\mathcal{C}_0$ and $\mathcal{C}_1$ are sampled independently of each other, and so $n_0$ and $n_1$ cannot be used to estimate their prior probabilities. We therefore assume the prior probabilities are known and use them directly. The data is then generated such that $\frac{n_0}{n}\approx\pi_0$ and $\frac{n_1}{n}\approx\pi_1$.

We consider two cases, one where $n>p$ so that $\hat{\bm{\Sigma}}$ is already invertible and one where $n<p$ so that $\hat{\bm{\Sigma}}$ is singular. In the first case, classification by LDA is possible, whereas in the second case, it is not. In either scenario, the G-estimator works well to tune $d$, as we shall demonstrate.

For the first experiment, we generate $n=400$ training points of dimension $p=200$ having class-conditional data distribution statistics

\begin{equation}
\mu_0=\frac{1}{p^{1/4}}\left[\textbf{1}^T_{\lceil{\sqrt{p}}\rceil} \ \textbf{0}^T_{p-\lceil{\sqrt{p}}\rceil-2} \ 2 \ 2\right]^T,\label{sim1}
\end{equation}
\begin{equation}
   \mu_1=\textbf{0}_p, \label{sim2}
\end{equation}
and
\begin{equation}
    \bm{\Sigma}=\frac{10}{p}\textbf{1}_p\textbf{1}_p^T+0.1\textbf{I}_p.\label{sim3}
\end{equation}
We want to estimate the error of an RP-LDA finite ensemble, with $M=100$ and unknown statistics, trained on this data, over a range of values of projection dimension $d$ and to tune $d$ accordingly.

To compute the error DE $\bar{\varepsilon}_{\hat{\bm{\mu}}_0,\hat{\bm{\mu}}_1,\hat{\bm{\Sigma}}}$ for a given $d$, we use Theorem 5. The root of \eqref{zetaSigmaHat1} is determined using the bisection method over $x>0$ with a tolerance of $10^{-6}$ as the stopping criterion. The quantity $\tilde{e}$ from the system of equations \eqref{systemE} is computed using a  fixed point iteration method with a tolerance of $10^{-6}$ as the stopping criterion. To compute the error G-estimator  $\hat{\varepsilon}$ for a given $d$, we use Theorem 6. The root of $f(x)$ is determined using the bisection method over $x>0$ with a tolerance of $10^{-6}$ as the stopping criterion.

We also compute the testing error of the RP-LDA ensemble with $M=100$ over an independently generated testing set of $10^{5}$ points. This serves as a benchmark for the accuracy of $\bar{\varepsilon}_{\hat{\bm{\mu}}_0,\hat{\bm{\mu}}_1,\hat{\bm{\Sigma}}}$ and $\hat{\varepsilon}$. As benchmarks for the performance of the RP-LDA ensemble as a classifier, we compute the testing error of an LDA classifier under unknown statistics and an LDA classifier under known statistics, which is the optimal Bayes error for this data, over the same testing set.

\begin{figure}[ht]
    \begin{center}
        \includegraphics[width=0.8\linewidth]{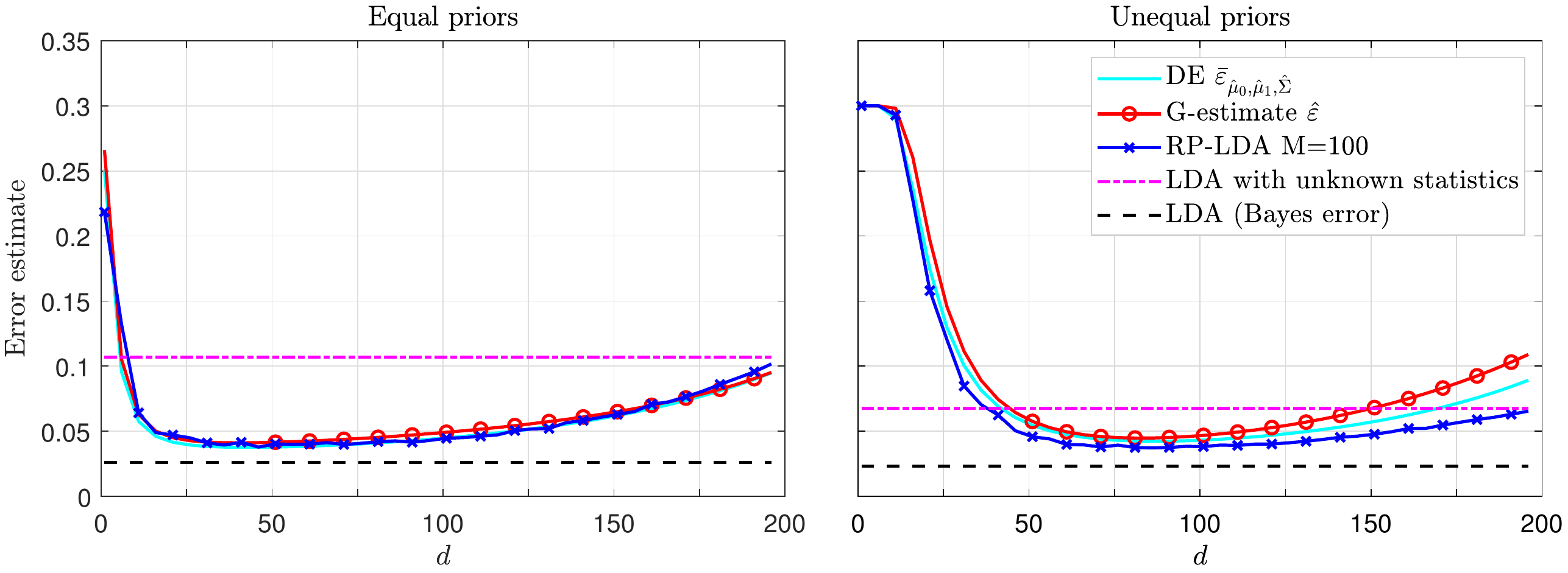}
    \end{center}
     \caption{The RP-LDA infinite ensemble error DE $\bar{\varepsilon}_{\hat{\bm{\mu}}_0,\hat{\bm{\mu}}_1,\hat{\bm{\Sigma}}}$, G-estimator $\hat{\varepsilon}$, and empirical errors of RP-LDA $M=100$, LDA with unknown statistics, and LDA with known statistics over $10^5$ testing points plotted against $d$ for fixed $p=200$ and $n=400$. The left-hand figure has $\pi_0=\pi_1=0.5$, while the right-hand figure has $\pi_0=0.7$ and $\pi_1=0.3$.}\label{fig4}
\end{figure}

Figure \ref{fig4} plots the aforementioned quantities against $d$ varying from $1$ up to $\text{rank}(\hat{\bm{\Sigma}})-2$ under the assumptions of equal priors $\pi_0=\pi_1=0.5$ and unequal priors $\pi_0=0.7$ and $\pi_1=0.3$, respectively.
In both cases, $\bar{\varepsilon}_{\hat{\bm{\mu}}_0,\hat{\bm{\mu}}_1,\hat{\bm{\Sigma}}}$ and $\hat{\varepsilon}$ follow the trend of the RP-LDA ensemble testing error and either quantity can be used to reliably select the projection dimension $d$ which minimizes the testing error. In the case of equal priors, the testing error estimate points to an optimal projection dimension of $d=46$ for which the testing error is $0.0378$, while $\bar{\varepsilon}_{\hat{\bm{\mu}}_0,\hat{\bm{\mu}}_1,\hat{\bm{\Sigma}}}$ and $\hat{\varepsilon}$ point to an optimal projection dimension of $d=41$ for which the testing error is $0.0415$. According to the testing error, using the G-estimator to tune $d$ would result in a loss of $0.0037$ in accuracy, but at a much lower computational cost. In the case of unequal priors, the testing error estimate and $\bar{\varepsilon}_{\hat{\bm{\mu}}_0,\hat{\bm{\mu}}_1,\hat{\bm{\Sigma}}}$ indicate an optimal projection dimension of $d=86$ for which the testing error is $0.0372$, while $\hat{\varepsilon}$ points to an optimal projection dimension of $d=81$ for which the testing error is $0.0375$. According to the testing error, using the G-estimator to tune $d$ would result in a negligible loss of $3\times 10^{-4}$ in accuracy.

Despite initially motivating a randomly projected variant of LDA by the small sample issue and the resulting singularity of $\hat{\bm{\Sigma}}$, Figure \ref{fig4} demonstrates that even when this is not an issue and LDA is possible, the RP-LDA ensemble may be a better choice in terms of accuracy. Note that this is shown analytically for $\bm{\Sigma}=\textbf{I}_p$ in Corollary $4$. Here we consider a general covariance. In the case of equal priors, the RP-LDA ensemble has a lower misclassification rate than an LDA classifier trained on the same data when $d>6$. At the optimal $d$, the RP-LDA ensemble error is $0.0378$  while the Bayes error is $0.0259$, a difference of $0.0119$. In the case of unequal priors, the RP-LDA ensemble has a lower misclassification rate than an LDA classifier trained on the same data when $d>36$. At the optimal $d$, the RP-LDA ensemble error is $0.0372$  while the Bayes error is $0.0230$, a difference of $0.0142$. The conclusion here is that for $\hat{\bm{\Sigma}}$ of full rank, an RP-LDA ensemble classifier can outperform an LDA classifier under unknown statistics and can even approach the Bayes error if $d$ is tuned properly.

\begin{figure}[ht]
    \begin{center}
        \includegraphics[width=0.8\linewidth]{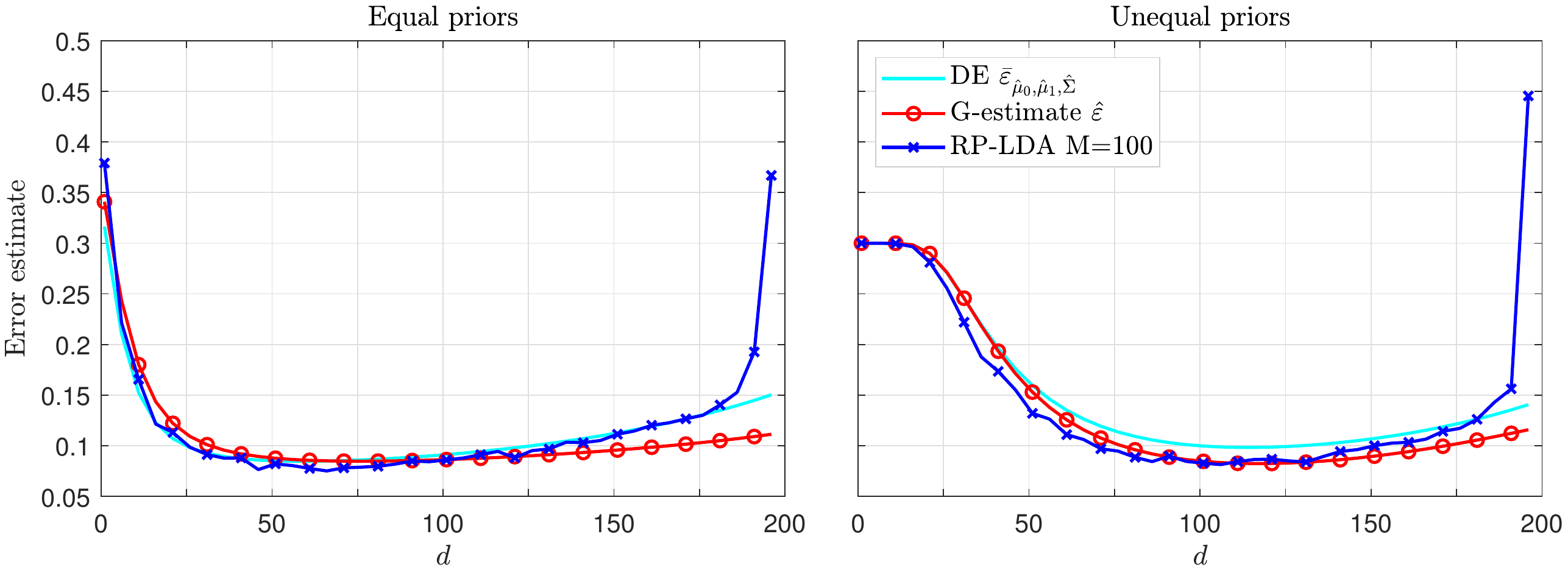}
    \end{center}
     \caption{The RP-LDA infinite ensemble error DE $\bar{\varepsilon}_{\hat{\bm{\mu}}_0,\hat{\bm{\mu}}_1,\hat{\bm{\Sigma}}}$, G-estimator $\hat{\varepsilon}$, and empirical error of RP-LDA $M=100$ over $10^5$ testing points plotted against $d$ for fixed $p=400$ and $n=200$. The left-hand figure has $\pi_0=\pi_1=0.5$, while the right-hand figure has $\pi_0=0.7$ and $\pi_1=0.3$.}\label{fig5}
\end{figure}

For the second experiment in this section, we generate $n=200$ training points of dimension $p=400$ having the same class-conditional distribution statistics \eqref{sim1}, \eqref{sim2} and \eqref{sim3}, as in the first simulation. For these problem dimensions, an LDA classifier is not possible. We plot the error DE $\bar{\varepsilon}_{\hat{\bm{\mu}}_0,\hat{\bm{\mu}}_1,\hat{\bm{\Sigma}}}$, error G-estimator $\hat{\varepsilon}$, and empirical error of an RP-LDA ensemble classifier with M=100 over an independently generated set of $10^{5}$ test points against $d$. This is shown in Figure \ref{fig5} for $d$ varying from $1$ up to $\text{rank}(\hat{\bm{\Sigma}})-2$ under the assumption of equal priors $\pi_0=\pi_1=0.5$ and unequal priors $\pi_0=0.7$ and $\pi_1=0.3$.

In both cases, $\bar{\varepsilon}_{\hat{\bm{\mu}}_0,\hat{\bm{\mu}}_1,\hat{\bm{\Sigma}}}$ and $\hat{\varepsilon}$ follow the trend set by the testing error estimate. In the case of equal priors, the optimal $d$ according to the testing error would be $d=66$ at an error of $0.0751$. Using the G-estimator to tune $d$ would result in a loss of $0.0037$ in accuracy at lower computational cost. In the case of unequal priors, the optimal $d$ according to the testing error would be $d=106$ at an error of $0.0814$. Using the G-estimator to tune $d$ would result in a loss of $0.005$ in accuracy.

These simulations demonstrate that for data distributed as \eqref{dist}, the RP-LDA infinite ensemble error G-estimator $\hat{\varepsilon}$ is sufficiently accurate to tune $d$ reliably when $M$ is large enough. This is true whether $n>p$ or $n<p$. The simulations also show that selecting  the projection dimension $d$ properly can lead to better performance than LDA under unknown statistics and even approaching that of the Bayes optimal classifier for such data. In the next section we show that these observations translate to real datasets as well.

\subsection{Real Data}\label{real}
In this set of experiments, we demonstrate the use of the G-estimator as a reliable estimate of the error of an RP-LDA finite ensemble with $M=100$ over several real datasets and show how this fact can be used to tune the projection dimension $d$. Our benchmark is an averaged 10-fold cross-validation. The training data is randomly shuffled and then partitioned into $10$ folds by which a single 10-fold cross-validation estimate is produced. The data is then shuffled again and the process is repeated. This is done $100$ times, yielding $100$ instances of the 10-fold cross-validation estimate. These are then averaged to produce the final estimate. The objective of this procedure is to reduce the variance of the benchmark, although in practice a single 10-fold cross-validation estimate is used.

We first consider the phoneme dataset from \cite{hastie1995penalized}. It consists of a total of $4509$ instances of digitized speech vectors of the five phonemes `aa', `ao', `dcl', `iy', and `sh', having $256$ features each. We extracted all $1717$ instances of the phonemes `ao' and `aa' (which are the closest in pronunciation) in order to construct a binary classification problem. We label `ao' as belonging to class $\mathcal{C}_0$ and `aa' as belonging to class $\mathcal{C}_1$, with $n_0=1022$ and $n_1=695$.

\begin{figure}[ht]
    \begin{center}
        \includegraphics[width=0.5\linewidth]{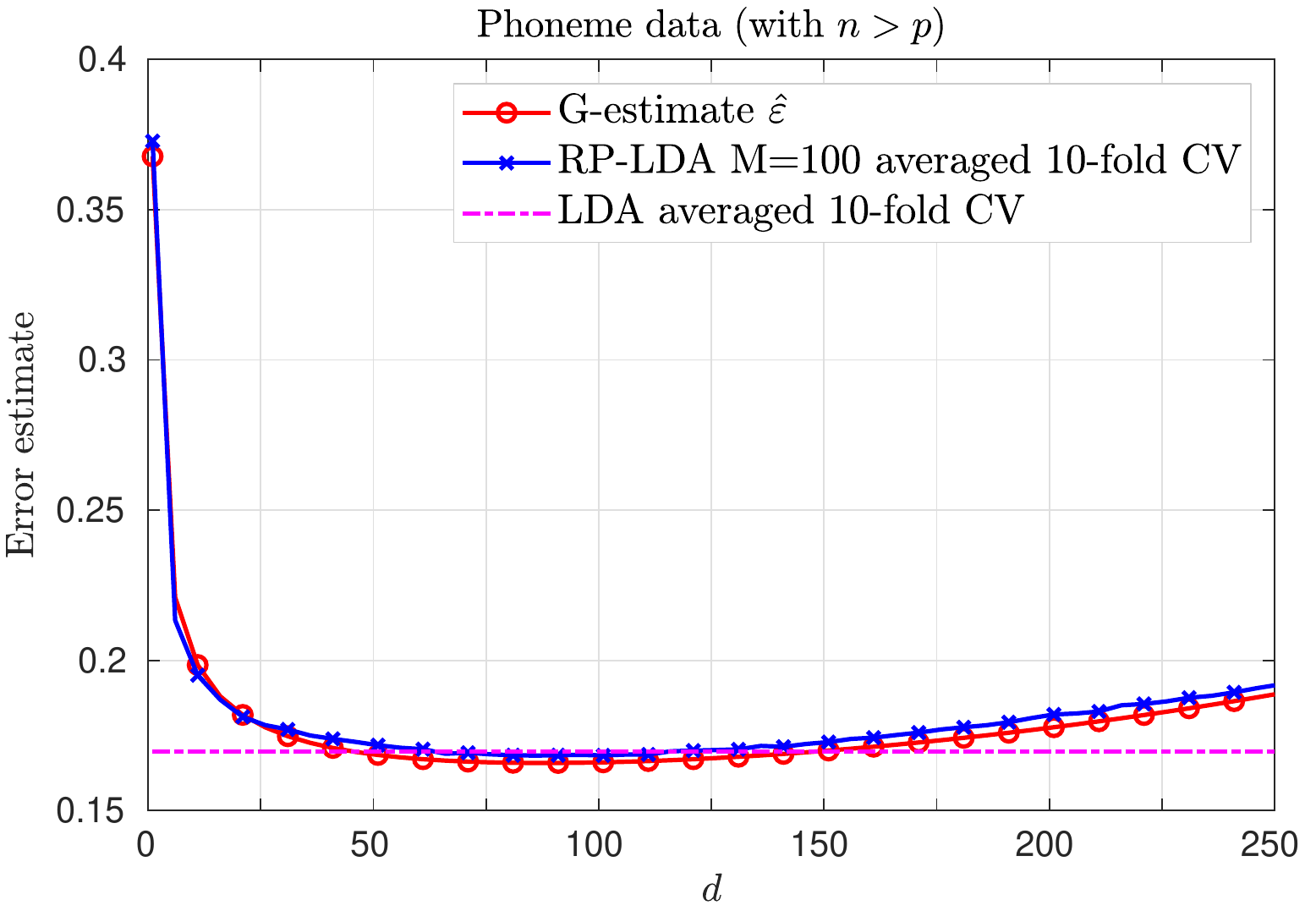}
    \end{center}
     \caption{The RP-LDA infinite ensemble error G-estimator $\hat{\varepsilon}$ and the averaged 10-fold cross-validation estimates of the errors of  RP-LDA $M=100$ and LDA plotted against $d$ for phoneme data with $p=256$, $n=1717$, $n_0=1022$, and $n_1=695$.}
     \label{fig6}
\end{figure}

For the first experiment on this dataset, we use all available data samples. The problem dimensions are $p=256$, $n=1717$, $n_0=1022$, and $n_1=695$. We train an RP-LDA ensemble classifier with $M=100$ as well as an LDA classifier on this data. Figure \ref{fig6} plots the G-estimate $\hat{\varepsilon}$ of RP-LDA infinite ensemble error, the averaged 10-fold cross-validation estimate of the error of the RP-LDA ensemble  with $M=100$, and the averaged 10-fold cross-validation estimate of the error of LDA with unknown statistics against $d$. As the cross-validation estimate is computed on a subset of the complete training set, it is the bottleneck for how large $d$ can be set without $\hat{\bm{\Sigma}}^{\text{RP}}$ becoming singular. We vary $d$ from 1 up to the smallest rank minus 2 among the sample covariances resulting from cross-validation.

Figure \ref{fig6} shows that for this dataset, $\hat{\varepsilon}$ follows the trend of the averaged 10-fold cross-validation estimate with $d$. Both $\hat{\varepsilon}$ and the cross-validation estimate indicate an optimal $d$ of $86$. Additionally, according to the cross-validation estimate, the RP-LDA ensemble with $M=100$ outperforms LDA over the range $61<d<116$.

\begin{figure}[ht]
    \begin{center}
        \includegraphics[width=0.5\linewidth]{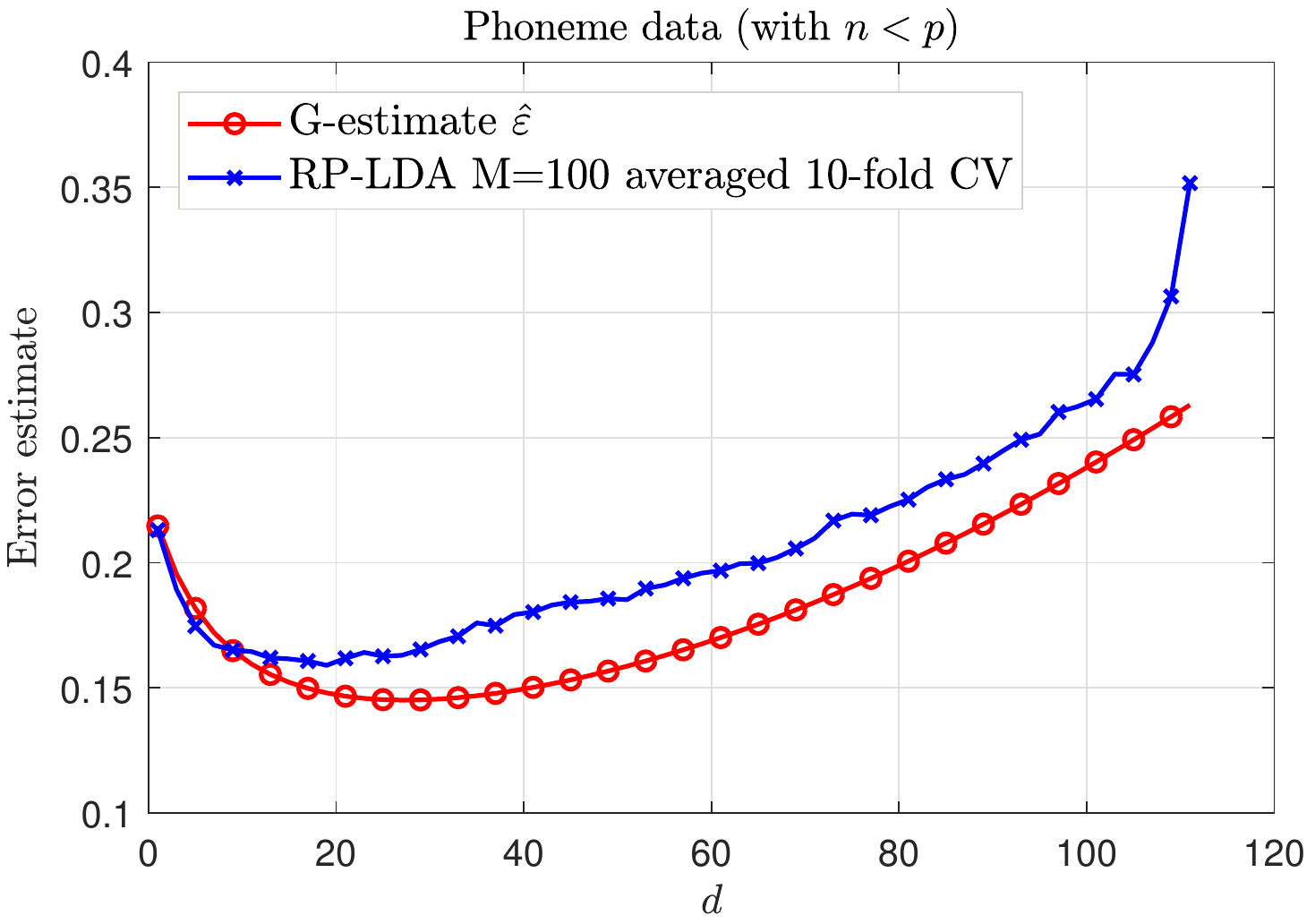}
    \end{center}
     \caption{The RP-LDA infinite ensemble error G-estimator $\hat{\varepsilon}$ and the averaged 10-fold cross-validation estimate of the error of  RP-LDA $M=100$ plotted against $d$ for phoneme data with $p=256$, $n=128$, $n_0=64$, and $n_1=64$.}\label{fig7}
\end{figure}

Next we consider a case where $n<p$. We construct such a dataset by extracting $64$ samples of the phoneme `ao' and 64 samples of the phoneme `aa'. The problem dimensions are $p=256$, $n=128$, $n_0=64$, and $n_1=64$. Figure \ref{fig7} plots the G-estimate $\hat{\varepsilon}$ of RP-LDA infinite ensemble error and the averaged 10-fold cross-validation estimate of the error of the RP-LDA ensemble  with $M=100$ against $d$.

 Once again, $\hat{\varepsilon}$ follows the trend of the cross-validation estimate. The cross-validation estimate points to an optimum of $d=19$ at an error of $ 0.1590$, while $\hat{\varepsilon}$ points to an optimum of $d=27$ at which the cross-validation estimate of error is $0.1630$. Using the G-estimator to tune $d$, we incur a loss in accuracy of $0.004$, but at a much reduced computational cost.

\begin{figure}[ht]
    \begin{center}
        \includegraphics[width=0.5\linewidth]{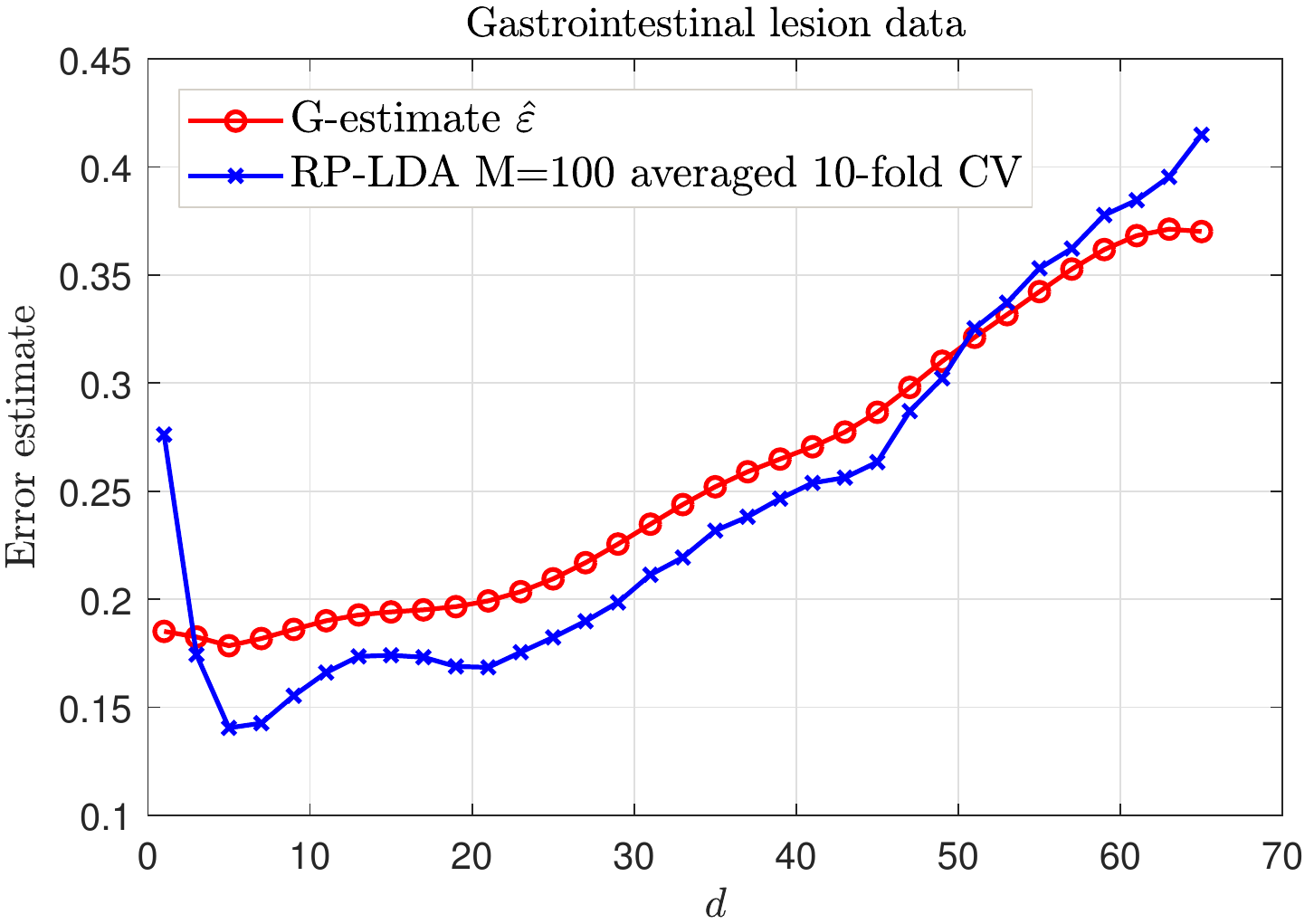}
    \end{center}
     \caption{The RP-LDA infinite ensemble error G-estimator $\hat{\varepsilon}$ and the averaged 10-fold cross-validation estimate of the error of  RP-LDA $M=100$ plotted against $d$ for gastrointestinal lesion data with $p=689$, $n=76$, $n_0=21$, and $n_1=55$.}\label{fig8}
\end{figure}

 The next dataset we consider is \cite{mesejo2016computer} which consists of $76$ samples of $689$ features extracted from colonoscopic videos of gastrointestinal lesions. The lesions are either benign or malignant, with $21$ occurrences of benign lesions and $55$ occurrences of malignant lesions. There are actually two separate datasets: one in which the videos are recorded under white light and the other under narrow band imaging. We make use of the former. We sort benign lesions into class $\mathcal{C}_0$ and malignant lesions into class $\mathcal{C}_1$. The problem dimensions are $p=689$, $n=76$, $n_0=21$, and $n_1=55$. Figure \ref{fig8} plots the G-estimate $\hat{\varepsilon}$ of RP-LDA infinite ensemble error and the averaged 10-fold cross-validation estimate of the error of the RP-LDA ensemble  with $M=100$ against $d$.

The G-estimator $\hat{\varepsilon}$ follows the trend of the cross-validation estimate and also agrees on the optimal $d=5$ at which the cross-validation estimate of error is $ 0.1405$.

\begin{figure}[ht]
    \begin{center}
        \includegraphics[width=0.5\linewidth]{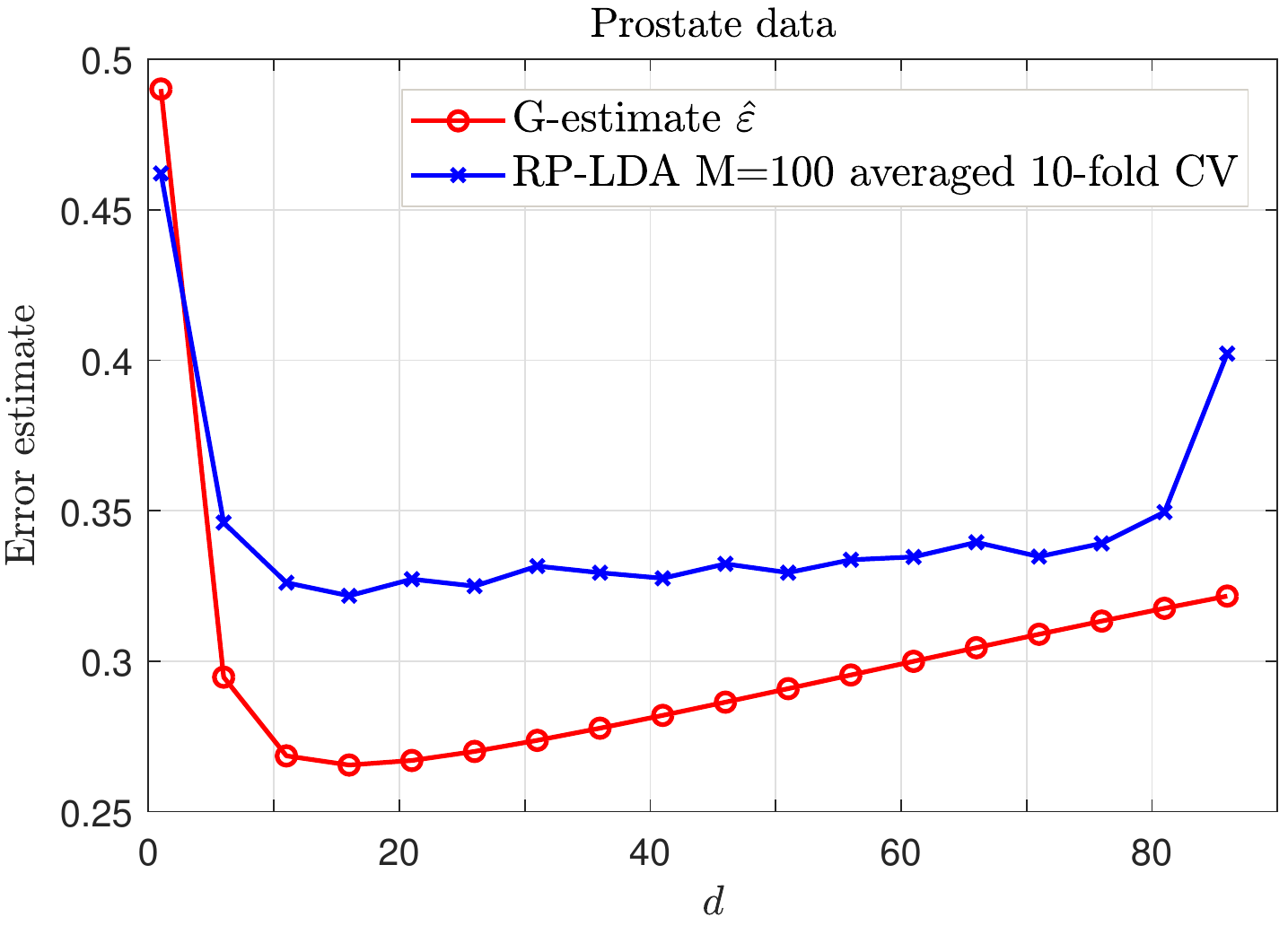}
    \end{center}
     \caption{The RP-LDA infinite ensemble error G-estimator $\hat{\varepsilon}$ and the averaged 10-fold cross-validation estimate of the error of  RP-LDA $M=100$ plotted against $d$ for prostate data with $p=6032$, $n=102$, $n_0=50$, and $n_1=52$.}\label{fig9}
\end{figure}

The final dataset we examine is \cite{singh2002gene}. It is a set of $102$ microarrays consisting of $6032$ gene expressions each, corresponding to $52$ men who have prostate cancer and $50$ men who do not have prostate cancer. We sort microarrays corresponding to healthy men into class $\mathcal{C}_0$ and microarrays corresponding to men who have prostate cancer into class $\mathcal{C}_1$. The problem dimensions are $p=6032$, $n=102$, $n_0=50$, and $n_1=52$. Figure \ref{fig9} plots the G-estimate $\hat{\varepsilon}$ of RP-LDA infinite ensemble error and the averaged 10-fold cross-validation estimate of the error of the RP-LDA ensemble  with $M=100$ against $d$.
The G-estimator $\hat{\varepsilon}$ follows the trend of the cross-validation estimate and also agrees on the optimal $d=14$ for which the cross-validation estimate of error is $0.3217$.

The above set of experiments show that the G-estimator $\hat{\varepsilon}$ can be relied on to tune the projection dimension $d$ in practical scenarios where datasets do not necessarily conform to the Gaussian assumptions under which $\hat{\varepsilon}$ was derived. They also show that the RP-LDA ensemble can outperform LDA with unknown statistics (when $n>p$) if $d$ is tuned properly.

\section{Conclusion}\label{conclusion}
In conclusion, we have conducted an asymptotic analysis of a classifier composed of an ensemble of randomly projected linear discriminants. This computationally inexpensive classifier was first introduced in \cite{durrant2013random} where it was shown to be competitive with the state-of-the-art SVM. It was also shown that the classifier's accuracy is sensitive to the projection dimension setting. This creates a need for a reliable and computationally inexpensive estimator of the classifier error in order to be able to tune the projection dimension correctly.

Through asymptotic analysis of the \textit{RP-LDA ensemble classifier}, we constructed a G-estimator of the classification error under a growth regime in which the data dimensions and projection dimension are assumed to grow at constant rates to each other. We demonstrated that the G-estimate is sufficiently accurate to reliably tune the projection dimension of a classifier trained on a given dataset with the goal of achieving the optimal misclassification rate. This provides an alternative to more computationally costly estimators such as cross-validation. We also showed that the Gaussian assumptions under which the G-estimator is derived are not too restrictive, as the estimate works well on several real datasets in addition to synthetic datasets.

In the process of deriving the G-estimator, we derived an asymptotic misclassification probability. This showed that the RP-LDA ensemble classifier behaves as a special case of an RLDA classifier, with the regularization parameter being a function of the projection dimension. This result indicates that an RP-LDA ensemble classifier cannot outperform an RLDA classifier for which the regularization parameter has been properly tuned. We also derived other asymptotic errors under different cases of known and unknown statistics. Assuming equal priors and isotropic population covariance, we obtained closed form expressions for these quantities. We compared these with existing analogous expressions for LDA and RP-LDA. Most notably, the results suggest that when the statistics are completely known, the RP-LDA ensemble classifier matches the performance of LDA with known statistics, which is the optimal classifier when the data is Gaussian. Furthermore, when the covariance is unknown, the RP-LDA ensemble classifier performance can actually be improved over LDA under identical conditions by setting $d$ appropriately. When the means are unknown and the covariance is known, the RP-LDA ensemble classifier offers no additional performance advantage over LDA under this setting besides lower computational cost due to working with reduced data dimensions.

\appendices
\section{Derivation of Deterministic Equivalents}\label{app:appA}
\subsection{Preliminaries}
Regardless of the values taken by the parameters $\tilde{\bm{\mu}}_0$, $\tilde{\bm{\mu}}_1$,$\tilde{\bm{\Sigma}}$, $\tilde{\pi}_0$ and $\tilde{\pi}_1$, the class-conditional discriminant statistics in \eqref{m0}, \eqref{m1}, and \eqref{sigma} are random in the projection matrix $\textbf{R}$. Thus, the first step in deriving DEs of these quantities, is to derive the DEs with respect to $\textbf{R}$, that is, with conditioning on the parameters $\tilde{\bm{\mu}}_0$, $\tilde{\bm{\mu}}_1$,$\tilde{\bm{\Sigma}}$, $\tilde{\pi}_0$ and $\tilde{\pi}_1$. This first stage yields the result of Theorem 1. From there, the various combinations of known and unknown statistics can be substituted for the parameters, and further work done to obtain DEs with respect to the training set $\mathcal{T}$, where applicable. The class-conditional discriminant statistics' DEs with respect to the random projection are derived in the first subsection of this appendix via Lemma 4 (to be presented in what follows). With the help of this result, the later subsections deal with the proofs of each of the special cases stated in Theorems 2 to 5.

\textbf{Lemma 4} \textit{Let $\textbf{a}$ and $\textbf{b}$ be $p\times 1$ vectors, independent of $\textbf{R}$, and $\gamma>0$}. \textit{If $\limsup\Vert\textbf{a}\Vert_2<
\infty$ and $\limsup\Vert\textbf{b}\Vert_2<
\infty$, then under assumptions (c), (d), and (f)}

\begin{equation*}
    \textbf{a}^T\left(\mathbb{E}_{\textbf{R}}\left[\textbf{R}^T\left(\textbf{R}\tilde{\bm{\Sigma}}\textbf{R}^T+\gamma\textbf{I}_p\right)^{-1}\textbf{R}\right]-\left(\tilde{\bm{\Sigma}}+\frac{1}{\delta({\tilde{\bm{\Sigma}}})}\textbf{I}_p\right)^{-1}\right)\textbf{b}\xrightarrow[]{\text{a.s.}}0
\end{equation*}
\textit{where $\delta({\tilde{\bm{\Sigma}}})$ satisfies the system of equations given by}
\begin{align*}
    &\delta({\tilde{\bm{\Sigma}}})=\frac{1}{\gamma\left(1+\frac{p}{d}\tilde{\delta}({\tilde{\bm{\Sigma}}})\right)}\\
    &\tilde{\delta}({\tilde{\bm{\Sigma}}})=\frac{1}{\gamma}\frac{1}{p}\text{tr}\left\{\tilde{\textbf{D}}\left(\textbf{I}_p+\delta({\tilde{\bm{\Sigma}}})\tilde{\textbf{D}}\right)^{-1}\right\}
\end{align*}
\textbf{Proof:}
Letting ${\textbf{R}}=[{\textbf{r}}_1\cdots {\textbf{r}}_p]$, $\tilde{\bm{\Sigma}}$ be eigendecomposed as $\tilde{\bm{\Sigma}}=\tilde{\textbf{U}}\tilde{\textbf{D}}\tilde{\textbf{U}}^T$, and noting that $\textbf{R}\tilde{\textbf{U}}$ has the same distribution as $\textbf{R}$ since orthogonal transformation preserves the Gaussian distribution,
\begin{align}
    \textbf{a}^T\mathbb{E}_{\textbf{R}}\left[\textbf{R}^T(\textbf{R}\tilde{\Sigma}\textbf{R}^T+\gamma
\textbf{I}_p)^{-1}\textbf{R}\right]\textbf{b}
&=\mathbb{E}_{\textbf{R}}\left[\textbf{a}^T\textbf{R}^T(\textbf{R}\tilde{\Sigma}\textbf{R}^T+\gamma
\textbf{I}_p)^{-1}\textbf{R}\textbf{b}\right]\nonumber \\
&=\mathbb{E}_{\textbf{R}}\left[\textbf{a}^T\tilde{\textbf{U}}\tilde{\textbf{U}}^T\textbf{R}^T(\textbf{R}\tilde{\textbf{U}}\tilde{\textbf{D}}\tilde{\textbf{U}}^T\textbf{R}^T+\gamma\textbf{I}_p)\textbf{R}\tilde{\textbf{U}}\tilde{\textbf{U}}^T\textbf{b}\right]\nonumber \\
&=\mathbb{E}_{\textbf{R}}\left[\textbf{a}^T\tilde{\textbf{U}}\textbf{R}^T(\textbf{R}\tilde{\textbf{D}}\textbf{R}^T+\gamma\textbf{I}_p)\textbf{R}\tilde{\textbf{U}}^T\textbf{b}\right]\nonumber \\
&=\mathbb{E}_{\textbf{R}}\left[\sum_{i,j}\left[\tilde{\textbf{U}}^T\textbf{a}\right]_i\left[\tilde{\textbf{U}}^T\textbf{b}\right]_j{\textbf{r}}_i^T({\textbf{R}}\tilde{\textbf{D}}{\textbf{R}}^T+\gamma\textbf{I}_p)^{-1}{\textbf{r}}_j\right]\label{a1}
\end{align}
Equation \eqref{a1} can be expressed as a sum of summations over indices $i=j$ and $i\ne j$. The latter term \\ $\mathbb{E}_{\textbf{R}}\left[\sum_{i\ne j}\left[\tilde{\textbf{U}}^T\textbf{a}\right]_i\left[\tilde{\textbf{U}}^T\textbf{b}\right]_j{\textbf{r}}_i^T({\textbf{R}}\tilde{\textbf{D}}{\textbf{R}}^T+\gamma\textbf{I}_p)^{-1}{\textbf{r}}_j\right]$ converges almost surely to zero and so we have
\begin{equation}
    \textbf{a}^T\mathbb{E}_{\textbf{R}}\left[\textbf{R}^T(\textbf{R}\tilde{\Sigma}\textbf{R}^T+\gamma
\textbf{I}_p)^{-1}\textbf{R}\right]\textbf{b}\asymp\mathbb{E}_{\textbf{R}}\left[\sum_{i}\left[\tilde{\textbf{U}}^T\textbf{a}\right]_i\left[\tilde{\textbf{U}}^T\textbf{b}\right]_i{\textbf{r}}_i^T({\textbf{R}}\tilde{\textbf{D}}{\textbf{R}}^T+\gamma\textbf{I}_p)^{-1}{\textbf{r}}_i\right]\label{rem}
\end{equation}
The proof of $\mathbb{E}_{\textbf{R}}\left[\sum_{i\ne j}\left[\tilde{\textbf{U}}^T\textbf{a}\right]_i\left[\tilde{\textbf{U}}^T\textbf{b}\right]_j{\textbf{r}}_i^T({\textbf{R}}\tilde{\textbf{D}}{\textbf{R}}^T+\gamma\textbf{I}_p)^{-1}{\textbf{r}}_j\right]\xrightarrow[]{\text{a.s.}}0$ is omitted but can be shown using techniques similar to those used to develop the asymptotic expression in \eqref{rem} in what follows.

Letting $\tilde{d}_k$ denote the $k^\text{th}$ entry of $\tilde{\textbf{D}}$ and by making use of the matrix inversion lemma (see \cite{muller2016random}),	
\begin{align}
    {\textbf{r}}_i^T({\textbf{R}}\tilde{\textbf{D}}{\textbf{R}}^T+\gamma\textbf{I}_p)^{-1}{\textbf{r}}_i&={\textbf{r}}_i^T\left(\sum_{k=1}^p\tilde{d}_k{\textbf{r}_k}{\textbf{r}_k}^T+\gamma\textbf{I}_p\right)^{-1}{\textbf{r}}_i\nonumber \\
    &=\frac{{\textbf{r}}_i^T\left(\sum_{k\ne i}\tilde{d}_k{\textbf{r}_k}{\textbf{r}_k}^T+\gamma\textbf{I}_p\right)^{-1}{\textbf{r}}_i}{1+\tilde{d}_i{\textbf{r}}_i^T\left(\sum_{k\ne i}\tilde{d}_k{\textbf{r}_k}{\textbf{r}_k}^T+\gamma\textbf{I}_p\right)^{-1}{\textbf{r}}_i} \label{a2}
\end{align}
Letting ${\alpha}_i:={\textbf{r}}_i^T\left(\sum_{k\ne i}\tilde{d}_k{\textbf{r}_k}{\textbf{r}_k}^T+\gamma\textbf{I}_p\right)^{-1}{\textbf{r}}_i$ and substituting \eqref{a2} into the asymptotic expression in \eqref{rem},
\begin{align}
    \mathbb{E}_{\textbf{R}}\left[\sum_{i}\left[\tilde{\textbf{U}}^T\textbf{a}\right]_i\left[\tilde{\textbf{U}}^T\textbf{b}\right]_i{\textbf{r}}_i^T({\textbf{R}}\tilde{\textbf{D}}{\textbf{R}}^T+\gamma\textbf{I}_p)^{-1}{\textbf{r}}_i\right]&=\sum_i\left[\tilde{\textbf{U}}^T\textbf{a}\right]_i\left[\tilde{\textbf{U}}^T\textbf{b}\right]_i\mathbb{E}_{{\textbf{R}}}\left[\frac{{\alpha}_i}{1+\tilde{d}_i{\alpha}_i}\right]\nonumber\\
    &=\sum_i\left[\tilde{\textbf{U}}^T\textbf{a}\right]_i\left[\tilde{\textbf{U}}^T\textbf{b}\right]_i\mathbb{E}_{{\textbf{R}}}\left[\frac{{\alpha}_i}{1+\tilde{d}_i\mathbb{E}_{\textbf{r}_i}[{\alpha}_i]}\right]+\epsilon
\end{align}
where $\mathbb{E}_{\textbf{r}_i}[\cdot]$ is with respect to $\textbf{r}_i$ conditioned on the rest of the columns of $\textbf{R}$ and
\begin{align}
        \epsilon&=\sum_i\left[\tilde{\textbf{U}}^T\textbf{a}\right]_i\left[\tilde{\textbf{U}}^T\textbf{b}\right]_i\mathbb{E}_{{\textbf{R}}}\left[\frac{{\alpha}_i}{1+\tilde{d}_i{\alpha}_i}\right]-\left[\tilde{\textbf{U}}^T\textbf{a}\right]_i\left[\tilde{\textbf{U}}^T\textbf{b}\right]_i\mathbb{E}_{{\textbf{R}}}\left[\frac{{\alpha}_i}{1+\tilde{d}_i\mathbb{E}_{{\textbf{r}}_i}[{\alpha}_i]}\right]\nonumber \\
    &=\sum_i\left[\tilde{\textbf{U}}^T\textbf{a}\right]_i\left[\tilde{\textbf{U}}^T\textbf{b}\right]_i\mathbb{E}_{{\textbf{R}}}\left[\frac{\tilde{d}_i{\alpha}_i(\mathbb{E}_{{\textbf{r}}_i}[{\alpha}_i]-{\alpha}_i)}{(1+\tilde{d}_i{\alpha}_i)(1+\tilde{d}_i\mathbb{E}_{{\textbf{r}}_i}[{\alpha}_i])}\right]
\end{align}
Here we have substituted $\alpha_i$ in the denominator by its expectation. The error in doing so is $\epsilon$. We can show that $\epsilon$ converges almost surely to zero by bounding it by a decaying function of $d$. First bound $\epsilon$ as follows, using the fact that $\alpha_i>0$ and $\tilde{d}_i \ge 0$ in the second line
\begin{align}
        |\epsilon|&\leq\sum_i\left|\left[\tilde{\textbf{U}}^T\textbf{a}\right]_i\right|\left|\left[\tilde{\textbf{U}}^T\textbf{b}\right]_i\right|\left|\mathbb{E}_{{\textbf{R}}}\left[\frac{\tilde{d}_i{\alpha}_i(\mathbb{E}_{{\textbf{r}}_i}[\alpha_i]-{\alpha}_i)}{(1+\tilde{d}_i{\alpha}_i)(1+\tilde{d}_i\mathbb{E}_{{\textbf{r}}_i}[{\alpha}_i])}\right]\right|\nonumber \\
        &\le\sum_i\left|\left[\tilde{\textbf{U}}^T\textbf{a}\right]_i\right|\left|\left[\tilde{\textbf{U}}^T\textbf{b}\right]_i\right|\mathbb{E}_{{\textbf{R}}}\left[{|\tilde{d}_i{\alpha}_i|\left|\mathbb{E}_{{\textbf{r}}_i}[\alpha_i]-{\alpha}_i\right|}\right]
\end{align}
By expressing $\mathbb{E}_{{\textbf{R}}}\left[|\tilde{d}_i{\alpha}_i||\mathbb{E}_{{\textbf{r}}_i}[{\alpha}_i]-{\alpha}_i|\right]=\mathbb{E}_{{\textbf{R}}}\left[\mathbb{E}_{{\textbf{r}}_i}\left[|\tilde{d}_i{\alpha}_i||\mathbb{E}_{{\textbf{r}}_i}[{\alpha}_i]-{\alpha}_i|\right]\right]$ and then applying the Cauchy-Schwarz inequality to the inner expectation, we have
\begin{equation}
    |\epsilon|\le\sum_i\left|\left[\tilde{\textbf{U}}^T\textbf{a}\right]_i\right|\left|\left[\tilde{\textbf{U}}^T\textbf{b}\right]_i\right|\mathbb{E}_{{\textbf{R}}}\left[ \sqrt{\mathbb{E}_{{\textbf{r}}_i}[(\tilde{d}_i{\alpha}_i)^2]}\sqrt{\mathbb{E}_{{\textbf{r}}_i}[(\mathbb{E}_{{\textbf{r}}_i}[\alpha_i]-\alpha_i)^2]}\right]\label{b1}
\end{equation}
Consider ${\mathbb{E}_{{\textbf{r}}_i}[(\mathbb{E}_{{\textbf{r}}_i}[\alpha_i]-\alpha_i)^2]}$ first. Using the fact that $\mathbb{E}_{{\textbf{r}}_i}\left[{\textbf{r}}_i{\textbf{r}}_i^T\right]=\frac{1}{d}\textbf{I}_p$, it can be shown that $\mathbb{E}_{{\textbf{r}}_i}[\alpha_i]=\frac{1}{d}\text{tr}\left\{\left(\sum_{k\ne i}\tilde{d}_k{\textbf{r}_k}{\textbf{r}_k}^T+\gamma\textbf{I}_p\right)^{-1}\right\}$. Applying the preliminary trace lemma  (see \cite{muller2016random}) to ${\mathbb{E}_{{\textbf{r}}_i}[(\mathbb{E}_{{\textbf{r}}_i}[\alpha_i]-\alpha_i)^2]}$, we have
\begin{equation}
    {\mathbb{E}_{{\textbf{r}}_i}[(\mathbb{E}_{{\textbf{r}}_i}[\alpha_i]-\alpha_i)^2]}\le\frac{C}{d}
\end{equation}
where $C$ is a constant. The bound on $\epsilon$ in \eqref{b1} then simplifies to
\begin{equation}
    |\epsilon|\le\frac{C'}{\sqrt{d}}\sum_i\left|\left[\tilde{\textbf{U}}^T\textbf{a}\right]_i\right|\left|\left[\tilde{\textbf{U}}^T\textbf{b}\right]_i\right|\mathbb{E}_{{\textbf{R}}}\left[ \sqrt{\mathbb{E}_{{\textbf{r}}_i}[{\alpha}_i^2]}\right]\label{b2}
\end{equation}
where $C'$ is a constant which incorporates the largest $\tilde{d}_i$ into the constant $C$. Now moving onto the term $\mathbb{E}_{{\textbf{R}}}\left[ \sqrt{\mathbb{E}_{{\textbf{r}}_i}[{\alpha}_i^2]}\right]$ and recalling that ${\alpha}_i={\textbf{r}}_i^T\left(\sum_{k\ne i}\tilde{d}_k{\textbf{r}_k}{\textbf{r}_k}^T+\gamma\textbf{I}_p\right)^{-1}{\textbf{r}}_i$
we have
\begin{align}
    \mathbb{E}_{{\textbf{R}}}\left[ \sqrt{\mathbb{E}_{{\textbf{r}}_i}[{\alpha}_i^2]}\right]&\le\sqrt{\mathbb{E}_{{\textbf{R}}}\left[ \mathbb{E}_{{\textbf{r}}_i}[{\alpha}_i^2]\right]}\nonumber \\
    &=\sqrt{\mathbb{E}_{{\textbf{R}}}\left[ {\alpha}_i^2\right]}\nonumber \\
    &\le\sqrt{\mathbb{E}_{{\textbf{R}}}\left[\left\Vert\left(\sum_{k\ne i}\tilde{d}_k{\textbf{r}_k}{\textbf{r}_k}^T+\gamma\textbf{I}_p\right)^{-1}\right\Vert_2^2\right]\mathbb{E}_{{\textbf{R}}}[||{\textbf{r}}_i||_2^4]}\label{a3}
\end{align}
where the first line uses Jensen's inequality and the last line uses the Cauchy-Schwarz inequality, the subordinance property of matrix norms, and finally the fact that \\ $\left(\sum_{k\ne i}\tilde{d}_k{\textbf{r}_k}{\textbf{r}_k}^T+\gamma\textbf{I}_p\right)^{-1}$ and ${\textbf{r}}_i$ are independent, in that order. It can be shown that
\begin{equation*}
    \left\Vert\left(\sum_{k\ne i}\tilde{d}_k{\textbf{r}_k}{\textbf{r}_k}^T+\gamma\textbf{I}_p\right)^{-1}\right\Vert_2^2\le\frac{1}{\gamma^2}
\end{equation*}
so that the term $\mathbb{E}_{{\textbf{R}}}\left[\left\Vert\left(\sum_{k\ne i}\tilde{d}_k{\textbf{r}_k}{\textbf{r}_k}^T+\gamma\textbf{I}_p\right)^{-1}\right\Vert_2^2\right]$ in \eqref{a3} bounded. The term $\mathbb{E}_{{\textbf{R}}}[||{\textbf{r}}_i||_2^4]$ is also bounded because all moments of a Gaussian vector are bounded. We now have
\begin{equation}
      |\varepsilon|<\frac{C''}{\sqrt{d}}\sum_i\left|\left[\tilde{\textbf{U}}^T\textbf{a}\right]_i\right|\left|\left[\tilde{\textbf{U}}^T\textbf{b}\right]_i\right|
      \le\frac{C''}{\sqrt{d}}\left\Vert \tilde{\textbf{U}}^T\textbf{a}\right\Vert_2\left\Vert \tilde{\textbf{U}}^T\textbf{b}\right\Vert_2
\end{equation}
where $C''$ is yet another constant and the second line follows by the Cauchy-Schwarz inequality. Since $\tilde{\textbf{U}}$ is an orthogonal matrix,  $||\tilde{\textbf{U}}||_2=1$, and therefore if $\limsup\limits_{p}||\textbf{a}||_2<
\infty$ and $\limsup\limits_{p}||\textbf{b}||_2<
\infty$ are satisfied then both $\left\Vert \tilde{\textbf{U}}^T\textbf{a}\right\Vert_2\le\left\Vert \tilde{\textbf{U}}\right\Vert_2\left\Vert{\textbf{a}}\right\Vert_2$ and $\left\Vert \tilde{\textbf{U}}^T\textbf{b}\right\Vert_2\le\left\Vert \tilde{\textbf{U}}\right\Vert_2\left\Vert{\textbf{b}}\right\Vert_2$ are bounded and we can claim, where $K$ is a constant,
\begin{equation}
    |\varepsilon|\le\frac{K}{\sqrt{d}}
\end{equation}
that is, $\varepsilon$ is almost-surely bounded by zero.

By showing that $\epsilon\xrightarrow{\text{a.s}}0$, we have
\begin{equation}
  \mathbb{E}_{\textbf{R}}\left[\sum_{i}\left[\tilde{\textbf{U}}^T\textbf{a}\right]_i\left[\tilde{\textbf{U}}^T\textbf{b}\right]_i{\textbf{r}}_i^T({\textbf{R}}\tilde{\textbf{D}}{\textbf{R}}^T+\gamma\textbf{I}_p)^{-1}{\textbf{r}}_i\right]\asymp\sum_i\left[\tilde{\textbf{U}}^T\textbf{a}\right]_i\left[\tilde{\textbf{U}}^T\textbf{b}\right]_i\mathbb{E}_{{\textbf{R}}}\left[\frac{{\alpha}_i}{1+\tilde{d}_i\mathbb{E}_{\textbf{r}_i}[{\alpha}_i]}\right]
\end{equation}
In conjunction with \eqref{rem}, this yields
\begin{equation}
\textbf{a}^T\mathbb{E}_{\textbf{R}}\left[\textbf{R}^T(\textbf{R}\tilde{\Sigma}\textbf{R}^T+\gamma
\textbf{I}_p)^{-1}\textbf{R}\right]\textbf{b}\asymp\sum_i\left[\tilde{\textbf{U}}^T\textbf{a}\right]_i\left[\tilde{\textbf{U}}^T\textbf{b}\right]_i\mathbb{E}_{{\textbf{R}}}\left[\frac{{\alpha}_i}{1+\tilde{d}_i\mathbb{E}_{\textbf{r}_i}[{\alpha}_i]}\right] \label{rem2}
\end{equation}
We further develop the asymptotic expression in \eqref{rem2} as follows
\begin{align}
   \sum_i\left[\tilde{\textbf{U}}^T\textbf{a}\right]_i\left[\tilde{\textbf{U}}^T\textbf{b}\right]_i\mathbb{E}_{{\textbf{R}}}\left[\frac{{\alpha}_i}{1+\tilde{d}_i\mathbb{E}_{\textbf{r}_i}[{\alpha}_i]}\right]
    &=\sum_i\left[\tilde{\textbf{U}}^T\textbf{a}\right]_i\left[\tilde{\textbf{U}}^T\textbf{b}\right]_i\mathbb{E}_{{\textbf{R}}}\left[\mathbb{E}_{{\textbf{r}_i}}\left[\frac{{\alpha}_i}{1+\tilde{d}_i\mathbb{E}_{{\textbf{r}}_i}[{\alpha}_i]}\right]\right]\nonumber \\
    &=\sum_i\left[\tilde{\textbf{U}}^T\textbf{a}\right]_i\left[\tilde{\textbf{U}}^T\textbf{b}\right]_i\mathbb{E}_{{\textbf{R}}}\left[\frac{\mathbb{E}_{{\textbf{r}_i}}[{\alpha}_i]}{1+\tilde{d}_i\mathbb{E}_{{\textbf{r}}_i}[{\alpha}_i]}\right]\nonumber \\
    &=\sum_i\left[\tilde{\textbf{U}}^T\textbf{a}\right]_i\left[\tilde{\textbf{U}}^T\textbf{b}\right]_i\mathbb{E}_{{\textbf{R}}}\left[\frac{\frac{1}{d}\text{tr}\left\{\left(\sum_{k\ne i}\tilde{d}_k{\textbf{r}_k}{\textbf{r}_k}^T+\gamma\textbf{I}_p\right)^{-1}\right\}}{1+\tilde{d}_i\frac{1}{d}\text{tr}\left\{\left(\sum_{k\ne i}\tilde{d}_k{\textbf{r}_k}{\textbf{r}_k}^T+\gamma\textbf{I}_p\right)^{-1}\right\}}\right]
\end{align}
By applying the rank-one perturbation lemma (see \cite{muller2016random}), the continuous mapping theorem, and the bounded convergence theorem, it can be shown that
\begin{align}
    &\sum_i\left[\tilde{\textbf{U}}^T\textbf{a}\right]_i\left[\tilde{\textbf{U}}^T\textbf{b}\right]_i\mathbb{E}_{{\textbf{R}}}\left[\frac{\frac{1}{d}\text{tr}\left\{\left(\sum_{k\ne i}\tilde{d}_k{\textbf{r}_k}{\textbf{r}_k}^T+\gamma\textbf{I}_p\right)^{-1}\right\}}{1+\tilde{d}_i\frac{1}{d}\text{tr}\left\{\left(\sum_{k\ne i}\tilde{d}_k{\textbf{r}_k}{\textbf{r}_k}^T+\gamma\textbf{I}_p\right)^{-1}\right\}}\right]\nonumber \\ &\hspace{30pt}\asymp \sum_i\left[\tilde{\textbf{U}}^T\textbf{a}\right]_i\left[\tilde{\textbf{U}}^T\textbf{b}\right]_i\mathbb{E}_{{\textbf{R}}}\left[\frac{\frac{1}{d}\text{tr}\left\{\left(\textbf{R}\tilde{\textbf{D}}\textbf{R}^T+\gamma\textbf{I}_p\right)^{-1}\right\}}{1+\tilde{d}_i\frac{1}{d}\text{tr}\left\{\left(\textbf{R}\tilde{\textbf{D}}\textbf{R}^T+\gamma\textbf{I}_p\right)^{-1}\right\}}\right]
\end{align}
The error in substituting  $\frac{1}{d}\text{tr}\left\{\left(\textbf{R}\tilde{\textbf{D}}\textbf{R}^T+\gamma\textbf{I}_p\right)^{-1}\right\}$ in the denominator by its expectation can be shown to converge almost surely to zero by a similar derivation to the preceding one and so we have
\begin{align}
    &\sum_i\left[\tilde{\textbf{U}}^T\textbf{a}\right]_i\left[\tilde{\textbf{U}}^T\textbf{b}\right]_i\mathbb{E}_{{\textbf{R}}}\left[\frac{\frac{1}{d}\text{tr}\left\{\left(\textbf{R}\tilde{\textbf{D}}\textbf{R}^T+\gamma\textbf{I}_p\right)^{-1}\right\}}{1+\tilde{d}_i\frac{1}{d}\text{tr}\left\{\left(\textbf{R}\tilde{\textbf{D}}\textbf{R}^T+\gamma\textbf{I}_p\right)^{-1}\right\}}\right]\nonumber \\ &\hspace{30pt}\asymp
    \sum_i\left[\tilde{\textbf{U}}^T\textbf{a}\right]_i\left[\tilde{\textbf{U}}^T\textbf{b}\right]_i\frac{\mathbb{E}_{{\textbf{R}}}\left[\frac{1}{d}\text{tr}\left\{\left(\textbf{R}\tilde{\textbf{D}}\textbf{R}^T+\gamma\textbf{I}_p\right)^{-1}\right\}\right]}{1+\tilde{d}_i\mathbb{E}_{{\textbf{R}}}\left[\frac{1}{d}\text{tr}\left\{\left(\textbf{R}\tilde{\textbf{D}}\textbf{R}^T+\gamma\textbf{I}_p\right)^{-1}\right\}\right]}\nonumber\\
    &\hspace{30pt}=\textbf{a}^T\tilde{\textbf{U}}\left(\tilde{\textbf{D}}+\frac{1}{\mathbb{E}_{\textbf{R}}\left[\frac{1}{d}\text{tr}\left\{\left(\textbf{R}\tilde{\textbf{D}}\textbf{R}^T+\gamma\textbf{I}_p\right)^{-1}\right\}\right]}\textbf{I}_p\right)^{-1}\tilde{\textbf{U}}^T\textbf{b}\nonumber\\
    &\hspace{30pt}=\textbf{a}^T\left(\tilde{\bm{\Sigma}}+\frac{1}{\mathbb{E}_{\textbf{R}}\left[\frac{1}{d}\text{tr}\left\{\left(\textbf{R}\tilde{\textbf{D}}\textbf{R}^T+\gamma\textbf{I}_p\right)^{-1}\right\}\right]}\textbf{I}_p\right)^{-1}\textbf{b}\label{a4}
\end{align}
It follows from \eqref{rem2}-\eqref{a4} that
\begin{equation}
\textbf{a}^T\mathbb{E}_{\textbf{R}}\left[\textbf{R}^T(\textbf{R}\tilde{\Sigma}\textbf{R}^T+\gamma
\textbf{I}_p)^{-1}\textbf{R}\right]\textbf{b}\asymp\textbf{a}^T\left(\tilde{\bm{\Sigma}}+\frac{1}{\mathbb{E}_{\textbf{R}}\left[\frac{1}{d}\text{tr}\left\{\left(\textbf{R}\tilde{\textbf{D}}\textbf{R}^T+\gamma\textbf{I}_p\right)^{-1}\right\}\right]}\textbf{I}_p\right)^{-1}\textbf{b} \label{rem3}
\end{equation}
Now it remains to find the deterministic equivalent of $\mathbb{E}_{\textbf{R}}\left[\frac{1}{d}\text{tr}\left\{\left(\textbf{R}\tilde{\textbf{D}}\textbf{R}^T+\gamma\textbf{I}_p\right)^{-1}\right\}\right]$ which can then be substituted directly into the asymptotic expression in \eqref{rem3} by the continuous mapping theorem.
By applying the result for the deterministic equivalent of the trace of the resolvent of a matrix with separable variance profile in \cite{hachem2013bilinear}, we have
\begin{equation}
    \frac{1}{d}\text{tr}\left\{\left(\textbf{R}\tilde{\textbf{D}}\textbf{R}^T+\gamma\textbf{I}_p\right)^{-1}\right\}\asymp\delta({\tilde{\bm{\Sigma}}})
\end{equation}
where $\delta({\tilde{\bm{\Sigma}}})$ satisfies
\begin{align}
    &\delta({\tilde{\bm{\Sigma}}})=\frac{1}{\gamma+\frac{1}{d}\text{tr}\left\{\tilde{\textbf{D}}\left(\textbf{I}_p+\delta({\tilde{\bm{\Sigma}}})\tilde{\textbf{D}}\right)^{-1}\right\}}\label{delta}
\end{align} assuming that  $\limsup\limits_{p} \Vert\bm{\Sigma}\Vert_2<\infty$. Since $\frac{1}{d}\text{tr}\left\{\left(\textbf{R}\tilde{\textbf{D}}\textbf{R}^T+\gamma\textbf{I}_p\right)^{-1}\right\}\le\frac{1}{\gamma}$, then by the bounded convergence theorem we have
\begin{equation}
    \mathbb{E}_{\textbf{R}}\left[\frac{1}{d}\text{tr}\left\{\left(\textbf{R}\tilde{\textbf{D}}\textbf{R}^T+\gamma\textbf{I}_p\right)^{-1}\right\}\right]\asymp\mathbb{E}_{\textbf{R}}\left[\delta({\tilde{\bm{\Sigma}}})\right]=\delta({\tilde{\bm{\Sigma}}})
\end{equation}
therefore
\begin{equation}
    \textbf{a}^T\left(\tilde{\bm{\Sigma}}+\frac{1}{\mathbb{E}_{\textbf{R}}\left[\frac{1}{d}\text{tr}\left\{\left(\textbf{R}\tilde{\textbf{D}}\textbf{R}^T+\gamma\textbf{I}_p\right)^{-1}\right\}\right]}\textbf{I}_p\right)^{-1}\textbf{b}\asymp\textbf{a}^T\left(\tilde{\bm{\Sigma}}+\frac{1}{\delta({\tilde{\bm{\Sigma}}})}\textbf{I}_p\right)^{-1}\textbf{b}\label{a5}
\end{equation}
and Lemma 4 follows from \eqref{rem3} and \eqref{a5}.

\subsection{Proof of Theorem 1}\label{firstDE}
To begin with, we derive the following intermediate convergence relations
\begin{align}
    &(\tilde{\bm{\mu}}_1-\tilde{\bm{\mu}}_0)^T\mathbb{E}_{\textbf{R}}\left[\textbf{R}^T(\textbf{R}\tilde{\Sigma}\textbf{R}^T+\gamma\textbf{I}_p)^{-1}\textbf{R}\right]\left(\bm{\mu}_0-\frac{\tilde{\bm{\mu}}_0+\tilde{\bm{\mu}}_1}{2}\right)+\text{ln}\frac{\tilde{\pi}_1}{\tilde{\pi}_0}\\
    &\hspace{80pt}\asymp(\tilde{\bm{\mu}}_1-\tilde{\bm{\mu}}_0)^T\left(\tilde{\bm{\Sigma}}+\frac{1}{\delta({\tilde{\bm{\Sigma}}})}\textbf{I}_p\right)^{-1}\left(\bm{\mu}_0-\frac{\tilde{\bm{\mu}}_0+\tilde{\bm{\mu}}_1}{2}\right)+\text{ln}\frac{\pi_1}{\pi_0}\\
     &(\tilde{\bm{\mu}}_1-\tilde{\bm{\mu}}_0)^T\mathbb{E}_{\textbf{R}}\left[\textbf{R}^T(\textbf{R}\tilde{\Sigma}\textbf{R}^T+\gamma\textbf{I}_p)^{-1}\textbf{R}\right]\left(\bm{\mu}_1-\frac{\tilde{\bm{\mu}}_0+\tilde{\bm{\mu}}_1}{2}\right)+\text{ln}\frac{\tilde{\pi}_1}{\tilde{\pi}_0}\\
      &\hspace{80pt}\asymp(\tilde{\bm{\mu}}_1-\tilde{\bm{\mu}}_0)^T\left(\tilde{\bm{\Sigma}}+\frac{1}{\delta({\tilde{\bm{\Sigma}}})}\textbf{I}_p\right)^{-1}\left(\bm{\mu}_1-\frac{\tilde{\bm{\mu}}_0+\tilde{\bm{\mu}}_1}{2}\right)+\text{ln}\frac{\pi_1}{\pi_0}\\
     &(\tilde{\bm{\mu}}_1-\tilde{\bm{\mu}}_0)^T\mathbb{E}_{\textbf{R}}\left[\textbf{R}^T(\textbf{R}\tilde{\Sigma}\textbf{R}^T+\gamma\textbf{I}_p)^{-1}\textbf{R}\right]\bm{\Sigma}\mathbb{E}_{\textbf{R}}\left[\textbf{R}^T(\textbf{R}\tilde{\Sigma}\textbf{R}^T+\gamma\textbf{I}_p)^{-1}\textbf{R}\right](\tilde{\bm{\mu}}_1-\tilde{\bm{\mu}}_0)\\
     &\hspace{80pt}\asymp(\tilde{\bm{\mu}}_1-\tilde{\bm{\mu}}_0)^T\left(\tilde{\bm{\Sigma}}+\frac{1}{\delta({\tilde{\bm{\Sigma}}})}\textbf{I}_p\right)^{-1}\bm{\Sigma}\left(\tilde{\bm{\Sigma}}+\frac{1}{\delta({\tilde{\bm{\Sigma}}})}\textbf{I}_p\right)^{-1}(\tilde{\bm{\mu}}_1-\tilde{\bm{\mu}}_0)
 \end{align}
 where the first relation corresponds to \eqref{m0}, the second to \eqref{m1}, and the final relation to \eqref{sigma}. For the initial result concerning \eqref{m0}, we have for the term $\text{ln}\frac{\tilde{\pi}_1}{\tilde{\pi}_0}$  the convergence $\text{ln}\frac{\tilde{\pi}_1}{\tilde{\pi}_0}\xrightarrow{\text{a.s.}}\text{ln}\frac{\pi_1}{\pi_0}$. The asymptotic expression for the remaining term can be obtained by first inserting $\gamma\textbf{I}_p$ to match Lemma 4, resulting in the expression
\begin{equation}
    (\tilde{\bm{\mu}}_1-\tilde{\bm{\mu}}_0)^T\mathbb{E}_{\textbf{R}}\left[\textbf{R}^T(\textbf{R}\tilde{\Sigma}\textbf{R}^T+\gamma\textbf{I}_p)^{-1}\textbf{R}\right]\left(\bm{\mu}_0-\frac{\tilde{\bm{\mu}}_0+\tilde{\bm{\mu}}_1}{2}\right)
\end{equation}
This is followed by a direct application of the result in Lemma 4 with $\textbf{a}=(\tilde{\bm{\mu}}_1-\tilde{\bm{\mu}}_0)$ and $\textbf{b}=\left(\bm{\mu}_0-\frac{\tilde{\bm{\mu}}_0+\tilde{\bm{\mu}}_1}{2}\right)$. The designated $\textbf{a}$ and $\textbf{b}$ vectors are bounded under assumption (e). The initial result concerning \eqref{m1} can be obtained in the same way.

The initial result concerning \eqref{sigma} can be obtained in a similar fashion by first adding in the $\gamma\textbf{I}_p$ term to both expectations yielding the expression
\begin{equation}
 (\tilde{\bm{\mu}}_1-\tilde{\bm{\mu}}_0)^T\mathbb{E}_{\textbf{R}}\left[\textbf{R}^T(\textbf{R}\tilde{\Sigma}\textbf{R}^T+\gamma\textbf{I}_p)^{-1}\textbf{R}\right]\bm{\Sigma}\mathbb{E}_{\textbf{R}}\left[\textbf{R}^T(\textbf{R}\tilde{\Sigma}\textbf{R}^T+\gamma\textbf{I}_p)^{-1}\textbf{R}\right](\tilde{\bm{\mu}}_1-\tilde{\bm{\mu}}_0)
\end{equation}
This is followed by successive applications of Lemma 4. In the first stage, \\ $\textbf{a}=(\tilde{\bm{\mu}}_1-\tilde{\bm{\mu}}_0)^T\mathbb{E}_{\textbf{R}}\left[\textbf{R}^T(\textbf{R}\tilde{\Sigma}\textbf{R}^T+\gamma\textbf{I}_p)^{-1}\textbf{R}\right]\bm{\Sigma}$ and $\textbf{b}=(\tilde{\bm{\mu}}_1-\tilde{\bm{\mu}}_0)$. After applying Lemma 4, we obtain
\begin{align}
  &(\tilde{\bm{\mu}}_1-\tilde{\bm{\mu}}_0)^T\mathbb{E}_{\textbf{R}}\left[\textbf{R}^T(\textbf{R}\tilde{\Sigma}\textbf{R}^T+\gamma\textbf{I}_p)^{-1}\textbf{R}\right]\bm{\Sigma}\mathbb{E}_{\textbf{R}}\left[\textbf{R}^T(\textbf{R}\tilde{\Sigma}\textbf{R}^T+\gamma\textbf{I}_p)^{-1}\textbf{R}\right](\tilde{\bm{\mu}}_1-\tilde{\bm{\mu}}_0)\\ &\hspace{70px}\asymp(\tilde{\bm{\mu}}_1-\tilde{\bm{\mu}}_0)^T\mathbb{E}_{\textbf{R}}\left[\textbf{R}^T(\textbf{R}\tilde{\Sigma}\textbf{R}^T+\gamma\textbf{I}_p)^{-1}\textbf{R}\right]\bm{\Sigma}\left(\tilde{\bm{\Sigma}}+\frac{1}{\delta({\tilde{\bm{\Sigma}}})}\textbf{I}_p\right)^{-1}(\tilde{\bm{\mu}}_1-\tilde{\bm{\mu}}_0)
\end{align}
In the second stage, $\textbf{a}=(\tilde{\bm{\mu}}_1-\tilde{\bm{\mu}}_0)$ and $\textbf{b}=\bm{\Sigma}\left(\tilde{\bm{\Sigma}}+\frac{1}{\delta({\tilde{\bm{\Sigma}}})}\textbf{I}_p\right)^{-1}(\tilde{\bm{\mu}}_1-\tilde{\bm{\mu}}_0)$. The designated $\textbf{a}$ and $\textbf{b}$ vectors at each stage are bounded under assumptions (e) and (f).
To recover the intermediate convergence relations involving \eqref{m0}, \eqref{m1}, and \eqref{sigma}, we take the limit as $\gamma\rightarrow 0$.   For the limits of the left-hand side of each of the relations to be defined, we introduce condition (g) of the growth regime, $\liminf\limits_{p} \lambda_{\text{min}}\left(\bm{\Sigma}\right)>0$. This ensures that every member of the sequence $\lim\limits_{\gamma\rightarrow 0}\textbf{R}\tilde{\Sigma}\textbf{R}^T+\gamma\textbf{I}_p$ is nonsingular. The right-hand side convergence results hold for $\gamma>0$. Since $\textbf{R}\tilde{\Sigma}\textbf{R}^T$ is nonsingular and its smallest eigenvalue is bounded uniformly away from zero, the convergence still holds for $\gamma$ in an open set containing zero. We now have the following convergence relations
\begin{align}
     m_0(\tilde{\bm{\mu}}_0,\tilde{\bm{\mu}}_1,\tilde{\bm{\Sigma}},\tilde{\pi}_0,\tilde{\pi}_1)-(\tilde{\bm{\mu}}_1-\tilde{\bm{\mu}}_0)^T\left(\tilde{\bm{\Sigma}}+\frac{1}{\lim\limits_{\gamma\rightarrow 0}\delta({\tilde{\bm{\Sigma}}})}\textbf{I}_p\right)^{-1}\left(\bm{\mu}_0-\frac{\tilde{\bm{\mu}}_0+\tilde{\bm{\mu}}_1}{2}\right)-\text{ln}\frac{\pi_1}{\pi_0}&\xrightarrow{\text{a.s.}}0\label{inter1}\\
     m_1(\tilde{\bm{\mu}}_0,\tilde{\bm{\mu}}_1,\tilde{\bm{\Sigma}},\tilde{\pi}_0,\tilde{\pi}_1)-(\tilde{\bm{\mu}}_1-\tilde{\bm{\mu}}_0)^T\left(\tilde{\bm{\Sigma}}+\frac{1}{\lim\limits_{\gamma\rightarrow 0}\delta({\tilde{\bm{\Sigma}}})}\textbf{I}_p\right)^{-1}\left(\bm{\mu}_1-\frac{\tilde{\bm{\mu}}_0+\tilde{\bm{\mu}}_1}{2}\right)-\text{ln}\frac{\pi_1}{\pi_0}&\xrightarrow{\text{a.s.}}0\label{inter2}\\
     \sigma^2(\tilde{\bm{\mu}}_0,\tilde{\bm{\mu}}_1,\tilde{\bm{\Sigma}})-(\tilde{\bm{\mu}}_1-\tilde{\bm{\mu}}_0)^T\left(\tilde{\bm{\Sigma}}+\frac{1}{\lim\limits_{\gamma\rightarrow 0}\delta({\tilde{\bm{\Sigma}}})}\textbf{I}_p\right)^{-1}\bm{\Sigma}\left(\tilde{\bm{\Sigma}}+\frac{1}{\lim\limits_{\gamma\rightarrow 0}\delta({\tilde{\bm{\Sigma}}})}\textbf{I}_p\right)^{-1}(\tilde{\bm{\mu}}_1-\tilde{\bm{\mu}}_0)&\xrightarrow{\text{a.s.}}0\label{inter3}
 \end{align}
 What remains to be determined is the term $\lim\limits_{\gamma\rightarrow 0}\delta({\tilde{\bm{\Sigma}}})$ as a function of the true covariance $\bm{\Sigma}$ for each of the cases ${\tilde{\bm{\Sigma}}}={{\bm{\Sigma}}}$ and ${\tilde{\bm{\Sigma}}}=\hat{{\bm{\Sigma}}}$.

 \subsubsection{Known covariance}\label{knownCov}
 First consider the case ${\tilde{\bm{\Sigma}}}={{\bm{\Sigma}}}$. Using the definitions of $\delta(\bm{\Sigma})$ and $\tilde{\delta}(\bm{\Sigma})$ in Lemma 4 and the eigendecomposition of $\bm{\Sigma}$ as ${\bm{\Sigma}}={\textbf{V}}\textbf{D}_{\bm{\Sigma}}{\textbf{V}}^T$, we obtain
\begin{equation}
    \delta(\bm{\Sigma})=\frac{1}{\gamma+\frac{1}{d}\text{tr}\left\{\textbf{D}_{\bm{\Sigma}}\left(\delta(\bm{\Sigma})\textbf{D}_{\bm{\Sigma}}+\textbf{I}_p\right)^{-1}\right\}}
\end{equation}
Taking $\lim\limits_{\gamma\rightarrow 0}$ of both sides and rearranging, we obtain
\begin{equation}
    1-\frac{1}{d}\text{tr}\left\{\textbf{D}_{\bm{\Sigma}}\left(\textbf{D}_{\bm{\Sigma}}+\frac{1}{\lim\limits_{\gamma\rightarrow 0}\delta(\bm{\Sigma})}\textbf{I}_p\right)^{-1}\right\}=0\label{fitme}
\end{equation}
The sequence $\delta(\bm{\Sigma})$ can be shown to be bounded for $\gamma\in[0,\infty)$. As a result, the sequence $\lim\limits_{\gamma\rightarrow 0}\delta(\bm{\Sigma})$ is also bounded. By the Bolzano-Weierstrass theorem, there exists a convergent subsequence of $\lim\limits_{\gamma\rightarrow 0}\delta(\bm{\Sigma})$. Additionally, any subsequence of $\lim\limits_{\gamma\rightarrow 0}\delta(\bm{\Sigma})$ and its limit, if it exists, should satisfy \eqref{fitme}.
Based on \eqref{fitme}, define the polynomial $g(x)$ as
\begin{align}
    g(x)&=1-\frac{1}{d}\text{tr}\left\{\textbf{D}_{\bm{\Sigma}}\left(\textbf{D}_{\bm{\Sigma}}+\frac{1}{x}\textbf{I}_p\right)^{-1}\right\}\nonumber\\
    &=1-\frac{1}{d}\sum_{i=1}^p\frac{\lambda_i(\bm{\Sigma})}{\lambda_i(\bm{\Sigma})+\frac{1}{x}}
\end{align}
where $\lambda_i(\bm{\Sigma})$ is the $i^{\text{th}}$ eigenvalue of $\bm{\Sigma}$. We observe that $g(x)$ is a monotonically decreasing function of $x$, $\lim\limits_{x\rightarrow 0}g(x)=1$, and $\lim\limits_{x\rightarrow \infty}g(x)=1-\frac{p}{d}$. If $p>d$ so that $1-\frac{p}{d}$ is negative, then $g(x)$ has a unique root over $x>0$. This is ensured by condition (c) of the growth regime. Since every subsequence of $\lim\limits_{\gamma\rightarrow 0}\delta(\bm{\Sigma})$ satisfies $g(x)$, every subsequence converges to the unique root of $g(x)$. Thus the limit of the sequence $\lim\limits_{\gamma\rightarrow 0}\delta(\bm{\Sigma})$ is the root of $g(x)$ over $x>0$. We denote this root by $\zeta_{\bm{\Sigma}}(\bm{\Sigma})$, where the subscript $\bm{\Sigma}$ indicates that $\tilde{\bm{\Sigma}}=\bm{\Sigma}$.  We can express $g(x)$ in terms of $\bm{\Sigma}$ by the multiplication
\begin{align}
     &1-\frac{1}{d}\text{tr}\left\{\textbf{V}^T\textbf{V}\textbf{D}_{\bm{\Sigma}}\textbf{V}^T\textbf{V}\left(\textbf{D}_{\bm{\Sigma}}+\frac{1}{x}\textbf{I}_p\right)^{-1}\right\}
\end{align}
after which the cyclic property of the trace results in
\begin{align}
     &1-\frac{1}{d}\text{tr}\left\{\bm{\Sigma}\left(\bm{\Sigma}+\frac{1}{x}\textbf{I}_p\right)^{-1}\right\}
\end{align}
so that we obtain the alternate form of $g(x)$ in \eqref{zetaSigma}.

\subsubsection{Unknown covariance}\label{theGREATunknown}
Now consider the case ${\tilde{\bm{\Sigma}}}=\hat{{\bm{\Sigma}}}$. We must derive $\lim\limits_{\gamma\rightarrow 0}\delta({\hat{\bm{\Sigma}}})$ in such a way that it no longer depends on $\hat{\bm{\Sigma}}$ and instead depends on the true covariance $\bm{\Sigma}$. Using the definitions of $\delta(\hat{\bm{\Sigma}})$ and $\tilde{\delta}(\hat{\bm{\Sigma}})$ in Lemma 4 and the eigendecomposition of $\hat{\bm{\Sigma}}$ as ${\bm{\Sigma}}={\textbf{U}}\textbf{D}{\textbf{U}}^T$, we obtain
\begin{equation}
    \delta(\hat{\bm{\Sigma}})=\frac{1}{\gamma+\frac{1}{d}\text{tr}\left\{\textbf{D}\left(\delta(\hat{\bm{\Sigma}})\textbf{D}+\textbf{I}_p\right)^{-1}\right\}}
\end{equation}
 Taking $\lim\limits_{\gamma\rightarrow 0}$ of both sides and rearranging results in
\begin{equation}
    1-\frac{1}{d}\text{tr}\left\{\textbf{D}\left(\textbf{D}+\frac{1}{\lim\limits_{\gamma\rightarrow 0}\delta(\hat{\bm{\Sigma}})}\textbf{I}_p\right)^{-1}\right\}=0\label{a6}
\end{equation}
The second term can be further manipulated as
\begin{align}
    \frac{1}{d}\text{tr}\left\{\textbf{D}\left(\textbf{D}+\frac{1}{\lim\limits_{\gamma\rightarrow 0}\delta(\hat{\bm{\Sigma}})}\textbf{I}_p\right)^{-1}\right\}
    &=\frac{1}{d}\text{tr}\left\{\textbf{U}^T\textbf{U}\textbf{D}\textbf{U}^T\textbf{U}\left(\textbf{D}+\frac{1}{\lim\limits_{\gamma\rightarrow 0}\delta(\hat{\bm{\Sigma}})}\textbf{I}_p\right)^{-1}\right\}\nonumber\\
    &=\frac{1}{d}\text{tr}\left\{\hat{\bm{\Sigma}}\left(\hat{\bm{\Sigma}}+\frac{1}{\lim\limits_{\gamma\rightarrow 0}\delta(\hat{\bm{\Sigma}})}\textbf{I}_p\right)^{-1}\right\}\nonumber\\
    &=\frac{1}{d}\text{tr}\left\{\left(\hat{\bm{\Sigma}}+\frac{1}{\lim\limits_{\gamma\rightarrow 0}\delta(\hat{\bm{\Sigma}})}\textbf{I}_p-\frac{1}{\lim\limits_{\gamma\rightarrow 0}\delta(\hat{\bm{\Sigma}})}\textbf{I}_p\right)\left(\hat{\bm{\Sigma}}+\frac{1}{\lim\limits_{\gamma\rightarrow 0}\delta(\hat{\bm{\Sigma}})}\textbf{I}_p\right)^{-1}\right\}\nonumber\\
    &=\frac{p}{d}-\frac{1}{\lim\limits_{\gamma\rightarrow 0}\delta(\hat{\bm{\Sigma}})}\frac{1}{d}\text{tr}\left\{\left(\hat{\bm{\Sigma}}+\frac{1}{\lim\limits_{\gamma\rightarrow 0}\delta(\hat{\bm{\Sigma}})}\textbf{I}_p\right)^{-1}\right\}\label{a7}
\end{align}
Now we must deal with the randomness coming from the sample covariance. The sample covariance $\hat{\bm{\Sigma}}$ can be expressed exactly as $\hat{\bm{\Sigma}}=\frac{1}{n-2}\bm{\Sigma}^{1/2}\bar{\textbf{Y}}\bar{\textbf{Y}}^T\bm{\Sigma}^{1/2}$ for some $\bar{\textbf{Y}}\in\mathbb{R}^{p\times (n-2)}$ which has i.i.d. columns distributed as $\mathcal{N}(\textbf{0},\textbf{I}_p)$. To do this, first it can be shown that
\begin{equation}
    \hat{\bm{\Sigma}}=\frac{1}{n-2}\bm{\Sigma}^{1/2}\textbf{Y}_0\left(\textbf{I}_{n_0}-\frac{\textbf{1}_{n_0}\textbf{1}_{n_0}^T}{n_0}\right)\textbf{Y}_0^T\bm{\Sigma}^{1/2}+\frac{1}{n-2}\bm{\Sigma}^{1/2}\textbf{Y}_1\left(\textbf{I}_{n_1}-\frac{\textbf{1}_{n_1}\textbf{1}_{n_1}^T}{n_1}\right)\textbf{Y}_1^T\bm{\Sigma}^{1/2}
\end{equation}
for some $\textbf{Y}_0\in\mathbb{R}^{p\times n_0}$ such that $\textbf{X}_0=\bm{\mu}_0\textbf{1}_{n_0}^T+\bm{\Sigma}^{1/2}\textbf{Y}_0$ and some $\textbf{Y}_1\in\mathbb{R}^{p\times n_1}$ such that $\textbf{X}_1=\bm{\mu}_1\textbf{1}_{n_1}^T+\bm{\Sigma}^{1/2}\textbf{Y}_1$. The columns of both $\textbf{Y}_0$ and $\textbf{Y}_1$ are distributed as $\mathcal{N}(\textbf{0},\textbf{I}_p)$. Since the terms $\frac{\textbf{1}_{n_0}\textbf{1}_{n_0}^T}{n_0}$ and $\frac{\textbf{1}_{n_1}\textbf{1}_{n_1}^T}{n_1}$ each have one eigenvalue which is equal to $1$ in both cases, their eigendecompositions can be represented as
\begin{equation}
\frac{\textbf{1}_{n_0}\textbf{1}_{n_0}^T}{n_0}=\textbf{U}_0\begin{bmatrix}
1 &  &  & \\
 & 0 &  & \\
 &  & \ddots & \\
 &  &  & 0
\end{bmatrix}\textbf{U}_0^T\hspace{15pt}
\text{and} \hspace{15pt}\frac{\textbf{1}_{n_1}\textbf{1}_{n_1}^T}{n_1}=\textbf{U}_1\begin{bmatrix}
1 &  &  & \\
 & 0 &  & \\
 &  & \ddots & \\
 &  &  & 0
\end{bmatrix}\textbf{U}_1^T\label{wow}
\end{equation}
where $\textbf{U}_0$ and $\textbf{U}_1$ have as their first columns the vectors $\frac{\textbf{1}_{n_0}}{\sqrt{n_0}}$ and $\frac{\textbf{1}_{n_1}}{\sqrt{n_1}}$ respectively. By using these same bases to eigendecompose $\textbf{I}_{n_0}$ and $\textbf{I}_{n_1}$ in \eqref{wow}, we obtain
\begin{align}
   \hat{\bm{\Sigma}}&=\frac{1}{n-2}\bm{\Sigma}^{1/2}{\textbf{Y}}_0\textbf{U}_0\begin{bmatrix}
0 &  &  & \\
 & 1 &  & \\
 &  & \ddots & \\
 &  &  & 1
\end{bmatrix}\textbf{U}_0^T{\textbf{Y}}_0^T\bm{\Sigma}^{1/2}+\frac{1}{n-2}\bm{\Sigma}^{1/2}{\textbf{Y}}_1\textbf{U}_1\begin{bmatrix}
0 &  &  & \\
 & 1 &  & \\
 &  & \ddots & \\
 &  &  & 1
\end{bmatrix}\textbf{U}_1^T{\textbf{Y}}_1^T\bm{\Sigma}^{1/2}\\
&\sim\frac{1}{n-2}\bm{\Sigma}^{1/2}{\textbf{Y}}_0\begin{bmatrix}
0 &  &  & \\
 & 1 &  & \\
 &  & \ddots & \\
 &  &  & 1
\end{bmatrix}{\textbf{Y}}_0^T\bm{\Sigma}^{1/2}+\frac{1}{n-2}\bm{\Sigma}^{1/2}{\textbf{Y}}_1\begin{bmatrix}
0 &  &  & \\
 & 1 &  & \\
 &  & \ddots & \\
 &  &  & 1
\end{bmatrix}\textbf{U}_1^T{\textbf{Y}}_1^T\bm{\Sigma}^{1/2}\\
&=\frac{1}{n-2}\bm{\Sigma}^{1/2}\begin{bmatrix}
\bar{\textbf{Y}}_0 & \bar{\textbf{Y}}_1
\end{bmatrix}\begin{bmatrix}
\bar{\textbf{Y}}_0^T\\
\bar{\textbf{Y}}_1^T
\end{bmatrix}\bm{\Sigma}^{1/2}
\end{align}
where $\bar{\textbf{Y}}_0^T\in\mathbb{R}^{p\times (n_0-1)}$ is the submatrix of ${\textbf{Y}}_0$ obtained by removing its first column and $\bar{\textbf{Y}}_1^T\in\mathbb{R}^{p\times (n_1-1)}$ is the submatrix of ${\textbf{Y}}_1$ obtained by removing its first column. Denoting $\bar{\textbf{Y}}=\begin{bmatrix}
\bar{\textbf{Y}}_0 & \bar{\textbf{Y}}_1
\end{bmatrix}$, we obtain $\hat{\bm{\Sigma}}=\frac{1}{n-2}\bm{\Sigma}^{1/2}\bar{\textbf{Y}}\bar{\textbf{Y}}^T\bm{\Sigma}^{1/2}$ where $\bar{\textbf{Y}}\in\mathbb{R}^{p\times (n-2)}$ has i.i.d. columns distributed as $\mathcal{N}(\textbf{0},\textbf{I}_p)$.

By substituting $\frac{1}{n-2}\bm{\Sigma}^{1/2}{\bar{\textbf{Y}}}{\bar{\textbf{Y}}}^T\bm{\Sigma}^{1/2}$ for $\hat{\bm{\Sigma}}$ and making use of the eigendecomposition $\bm{\Sigma}=\textbf{V}\textbf{D}_{\bm{\Sigma}}\textbf{V}^T$ along with the cyclic property of the trace, we obtain
\begin{align}
    1-\frac{p}{d}+\frac{1}{\lim\limits_{\gamma\rightarrow 0}\delta(\hat{\bm{\Sigma}})}\frac{1}{d}\text{tr}\left\{\left(\frac{1}{n-2}\textbf{D}_{\bm{\Sigma}}^{1/2}\textbf{W}\textbf{W}^T\textbf{D}_{\bm{\Sigma}}^{1/2}+\frac{1}{\lim\limits_{\gamma\rightarrow 0}\delta(\hat{\bm{\Sigma}})}\textbf{I}_p\right)^{-1}\right\}= 0\label{a8}
\end{align}
 where $\textbf{W}=\textbf{V}^T\bar{\textbf{Y}}\in\mathbb{R}^{p\times (n-2)}$ also has i.i.d columns distributed as $\mathcal{N}(\textbf{0},\textbf{I})$.
Equation \eqref{a8} involves the normalized trace of the resolvent of the matrix $\frac{1}{n-2}\textbf{D}_{\bm{\Sigma}}^{1/2}\textbf{W}\textbf{W}^T\textbf{D}_{\bm{\Sigma}}^{1/2}$ with separable variance profile. We can apply the result in \cite{hachem2013bilinear} to obtain its deterministic equivalent. Applying this result, we have
\begin{equation}
    1-\frac{p}{d}+\frac{1}{d}\text{tr}\left\{\left(\textbf{I}_p+\tilde{e}\textbf{D}_{\bm{\Sigma}}\right)^{-1}\right\}\asymp 0\label{a9}
\end{equation}
where
 \begin{align}
    &e=\lim\limits_{\gamma\rightarrow 0}\delta(\hat{\bm{\Sigma}})\frac{1}{n}\text{tr}\left\{\bm{\Sigma}\left(\tilde{e}\bm{\Sigma}+\textbf{I}_p\right)^{-1}\right\}\nonumber\\
    &\tilde{e}=\frac{\lim\limits_{\gamma\rightarrow 0}\delta(\hat{\bm{\Sigma}})}{1+e}\label{a10}
\end{align}
We will solve for $\tilde{e}$ using \eqref{a9} and then use the system of equations \eqref{a10} to solve for the limit of $\lim\limits_{\gamma\rightarrow 0}\delta(\hat{\bm{\Sigma}})$ following the same argument in Section \ref{knownCov}.

Equation \eqref{a9} suggests that asymptotically $\tilde{e}$ is the root of the polynomial
 \begin{align}
     h(x)&=1-\frac{p}{d}+\frac{1}{d}\text{tr}\left\{\left(x\textbf{D}_{\bm{\Sigma}}+\textbf{I}_p\right)^{-1}\right\}\nonumber\\
     &=1-\frac{p}{d}+\frac{1}{d}\sum_{i=1}^p\frac{1}{1+x\lambda_i\left(\bm{\Sigma}\right)}\label{h2}
 \end{align}
which is monotonically decreasing. As $x\rightarrow 0$, $h(x)$ tends to $1$ and as $x\rightarrow \infty$, $h(x)$ tends to $1-\frac{p}{d}$, which is negative when $p>d$, condition (c) of the growth regime. Thus $h(x)$ has a unique root $\tilde{e}$ over $x>0$ which we denote $x^*$ in Theorem 3. To prove that $\tilde{e}\asymp x^*$, first subtract \eqref{a9} from \eqref{h2} at its root to obtain the relation
\begin{align}
    \frac{1}{d}\sum_{i=1}^p\frac{1}{1+x^*\lambda_i(\bm{\Sigma})}-\frac{1}{d}\sum_{i=1}^p\frac{1}{1+\tilde{e}\lambda_i(\bm{\Sigma})}\asymp 0
\end{align}
which can be rewritten as
\begin{align}
     (\tilde{e}-x^*)\frac{1}{d}\sum_{i=1}^p\frac{\lambda_i(\bm{\Sigma})}{1+x^*\lambda_i(\bm{\Sigma})+\tilde{e}\lambda_i(\bm{\Sigma})+x^*\tilde{e}\lambda_i^2(\bm{\Sigma})}\asymp 0 \label{yup}
\end{align}
The sum can be bounded from below as
\begin{align}
    \frac{1}{d}\sum_{i=1}^p\frac{\lambda_i(\bm{\Sigma})}{1+x^*\lambda_i(\bm{\Sigma})+\tilde{e}\lambda_i(\bm{\Sigma})+x^*\tilde{e}\lambda_i(\bm{\Sigma})}&\ge\frac{p}{d}\frac{\lambda_{\text{min}}(\bm{\Sigma})}{1+x^*\lambda_{\text{max}}(\bm{\Sigma})+\tilde{e}\lambda_{\text{max}}(\bm{\Sigma})+x^*\tilde{e}\lambda_{\text{max}}^2(\bm{\Sigma})}\\
    &>0
\end{align}
Therefore \eqref{yup} implies $\tilde{e}\asymp x^*$. Multiplication by $\textbf{V}^T\textbf{V}$ puts $h(x)$ in terms of $\bm{\Sigma}$ as presented in Theorem 3. Using \eqref{a10} to solve for the limit of $\lim\limits_{\gamma\rightarrow 0}\delta(\hat{\bm{\Sigma}})$ in terms of $\tilde{e}$ which we denote by $\zeta_{\hat{\bm{\Sigma}}}(\bm{\Sigma})$, we obtain
 \begin{equation*}
    \zeta_{\hat{\bm{\Sigma}}}(\bm{\Sigma})=\frac{x^*}{1-x^*\frac{1}{n}\text{tr}\left\{\bm{\Sigma}\left(x^*\bm{\Sigma}+\textbf{I}_p\right)^{-1}\right\}}
\end{equation*}
where $\lim\limits_{\gamma\rightarrow 0}\delta(\hat{\bm{\Sigma}}) \asymp\zeta_{\hat{\bm{\Sigma}}}(\bm{\Sigma})$ follows by a similar argument to that in Section \ref{knownCov} using the Bolzano-Weierstrass theorem.

\subsection{Proof of Theorem 2}\label{appA1}
By setting $\tilde{\bm{\mu}}_0=\bm{\mu}_0$, $\tilde{\bm{\mu}}_1=\bm{\mu}_1$,and $\tilde{\bm{\Sigma}}=\bm{\Sigma}$ in Theorem 1, we directly obtain the expressions for the DEs stated in Theorem 2. The corresponding error DE $\bar{\varepsilon}_{{\bm{\mu}}_0,{\bm{\mu}}_1,{\bm{\Sigma}}}$ follows directly from Lemma 2.

\subsection{Proof of Theorem 3}\label{appA2}
By setting $\tilde{\bm{\mu}}_0=\hat{\bm{\mu}}_0$, $\tilde{\bm{\mu}}_1=\hat{\bm{\mu}}_1$, and $\tilde{\bm{\Sigma}}=\bm{\Sigma}$ in Theorem 1, we obtain the following intermediate convergence relations with respect to $\textbf{R}$
 \begin{align}
     m_0(\hat{\bm{\mu}}_0,\hat{\bm{\mu}}_1,{\bm{\Sigma}},\tilde{\pi}_0,\tilde{\pi}_1)-(\hat{\bm{\mu}}_1-\hat{\bm{\mu}}_0)^T\left({\bm{\Sigma}}+\frac{1}{\zeta_{\bm{\Sigma}}(\bm{\Sigma})}\textbf{I}_p\right)^{-1}\left(\bm{\mu}_0-\frac{\hat{\bm{\mu}}_0+\hat{\bm{\mu}}_1}{2}\right)-\text{ln}\frac{\pi_1}{\pi_0}&\xrightarrow{\text{a.s.}}0\label{first2}\\
     m_1(\hat{\bm{\mu}}_0,\hat{\bm{\mu}}_1,{\bm{\Sigma}},\tilde{\pi}_0,\tilde{\pi}_1)-(\hat{\bm{\mu}}_1-\hat{\bm{\mu}}_0)^T\left({\bm{\Sigma}}+\frac{1}{\zeta_{\bm{\Sigma}}(\bm{\Sigma})}\textbf{I}_p\right)^{-1}\left(\bm{\mu}_1-\frac{\hat{\bm{\mu}}_0+\hat{\bm{\mu}}_1}{2}\right)-\text{ln}\frac{\pi_1}{\pi_0}&\xrightarrow{\text{a.s.}}0\label{second2}\\
     \sigma^2(\hat{\bm{\mu}}_0,\hat{\bm{\mu}}_1,{\bm{\Sigma}})-(\hat{\bm{\mu}}_1-\hat{\bm{\mu}}_0)^T\left({\bm{\Sigma}}+\frac{1}{\zeta_{\bm{\Sigma}}(\bm{\Sigma})}\textbf{I}_p\right)^{-1}\bm{\Sigma}\left({\bm{\Sigma}}+\frac{1}{\zeta_{\bm{\Sigma}}(\bm{\Sigma})}\textbf{I}_p\right)^{-1}(\hat{\bm{\mu}}_1-\hat{\bm{\mu}}_0)&\xrightarrow{\text{a.s.}}0\label{third2}
 \end{align}
 The randomness in $\hat{\bm{\mu}}_0$ and $\hat{\bm{\mu}}_1$ can be expressed through random matrices $\textbf{Z}_0\in\mathbb{R}^{p\times n_0}$ and $\textbf{Z}_1\in\mathbb{R}^{p\times n_1}$ respectively, each defined as having i.i.d. Gaussian zero-mean and unit variance entries,
\begin{equation}
    \hat{\bm{\mu}}_0={\bm{\mu}}_0+\frac{\bm{\Sigma}^{1/2}\textbf{Z}_0\textbf{1}}{n_0}
\end{equation}
\begin{equation}
    \hat{\bm{\mu}}_1={\bm{\mu}}_1+\frac{\bm{\Sigma}^{1/2}\textbf{Z}_1\textbf{1}}{n_1}
\end{equation}
where $\frac{\textbf{Z}_0\textbf{1}}{n_0}\sim\mathcal{N}\left(\textbf{0}_p,\frac{1}{n_0}\textbf{I}_p\right)$ and $\frac{\textbf{Z}_1\textbf{1}}{n_1}\sim\mathcal{N}\left(\textbf{0}_p,\frac{1}{n_1}\textbf{I}_p\right)$. Substituting these into each of the convergence relations above and taking the expectation over $\textbf{Z}_0\textbf{1}$ and $\textbf{Z}_1\textbf{1}$ for each yields the deterministic equivalents $\bar{m}_{0,\hat{\bm{\mu}}_0,\hat{\bm{\mu}}_1,{\bm{\Sigma}}}$, $\bar{m}_{1,\hat{\bm{\mu}}_0,\hat{\bm{\mu}}_1,{\bm{\Sigma}}}$, and $\bar{\sigma}^2_{\hat{\bm{\mu}}_0,\hat{\bm{\mu}}_1,{\bm{\Sigma}}}$ respectively. The corresponding error DE $\bar{\varepsilon}_{\hat{\bm{\mu}}_0,\hat{\bm{\mu}}_1,{\bm{\Sigma}}}$ follows directly from Lemma 2.

\subsection{Proof of Theorem 4}\label{appA3}
By setting $\tilde{\bm{\mu}}_0={\bm{\mu}}_0$, $\tilde{\bm{\mu}}_1={\bm{\mu}}_1$, and $\tilde{\bm{\Sigma}}=\hat{\bm{\Sigma}}$ in Theorem 1, we obtain the following intermediate convergence relations with respect to $\textbf{R}$
 \begin{align}
     m_0({\bm{\mu}}_0,{\bm{\mu}}_1,\hat{\bm{\Sigma}},\tilde{\pi}_0,\tilde{\pi}_1)-({\bm{\mu}}_1-{\bm{\mu}}_0)^T\left(\hat{\bm{\Sigma}}+\frac{1}{\zeta_{\hat{\bm{\Sigma}}}(\bm{\Sigma})}\textbf{I}_p\right)^{-1}\left(\bm{\mu}_0-\frac{{\bm{\mu}}_0+{\bm{\mu}}_1}{2}\right)-\text{ln}\frac{\pi_1}{\pi_0}&\xrightarrow{\text{a.s.}}0\label{first3}\\
     m_1({\bm{\mu}}_0,{\bm{\mu}}_1,\hat{\bm{\Sigma}},\tilde{\pi}_0,\tilde{\pi}_1)-(\hat{\bm{\mu}}_1-\hat{\bm{\mu}}_0)^T\left(\hat{\bm{\Sigma}}+\frac{1}{\zeta_{\hat{\bm{\Sigma}}}(\bm{\Sigma})}\textbf{I}_p\right)^{-1}\left(\bm{\mu}_1-\frac{{\bm{\mu}}_0+{\bm{\mu}}_1}{2}\right)-\text{ln}\frac{\pi_1}{\pi_0}&\xrightarrow{\text{a.s.}}0\label{second3}\\
     \sigma^2({\bm{\mu}}_0,{\bm{\mu}}_1,\hat{\bm{\Sigma}})-({\bm{\mu}}_1-{\bm{\mu}}_0)^T\left(\hat{\bm{\Sigma}}+\frac{1}{\zeta_{\hat{\bm{\Sigma}}}(\bm{\Sigma})}\textbf{I}_p\right)^{-1}\bm{\Sigma}\left(\hat{\bm{\Sigma}}+\frac{1}{\zeta_{\hat{\bm{\Sigma}}}(\bm{\Sigma})}\textbf{I}_p\right)^{-1}({\bm{\mu}}_1-{\bm{\mu}}_0)&\xrightarrow{\text{a.s.}}0\label{third3}
 \end{align}
We deal with the random $\hat{\bm{\Sigma}}$ that occurs in the intermediate convergence relations using standard random matrix theory results. For the first two relations, substituting $\frac{1}{n}\textbf{D}_{\bm{\Sigma}}^{1/2}\textbf{W}\textbf{W}^T\textbf{D}_{\bm{\Sigma}}^{1/2}$ (as defined in Section \ref{theGREATunknown}) for the sample covariance will result in the same resolvent to which the results in \cite{hachem2013bilinear} can be applied to obtain the final forms in Theorem 4. The same can be done for the third relation, except that the expression involves a double resolvent. Applying the result in \cite{benaych2016spectral} for deterministic equivalents of double resolvents leads to the final form presented in Theorem 4. The double resolvent introduces the multiplicative factor $\kappa$. The error DE $\bar{\varepsilon}_{{\bm{\mu}}_0,{\bm{\mu}}_1,\hat{\bm{\Sigma}}}$ then follows directly from Lemma 2.
\subsection{Proof of Theorem 5}\label{appA4}
By setting $\tilde{\bm{\mu}}_0=\hat{\bm{\mu}}_0$, $\tilde{\bm{\mu}}_1=\hat{\bm{\mu}}_1$, and $\tilde{\bm{\Sigma}}=\hat{\bm{\Sigma}}$ in Theorem 1, we obtain the following intermediate convergence relations with respect to $\textbf{R}$
 \begin{align}
     m_0(\hat{\bm{\mu}}_0,\hat{\bm{\mu}}_1,\hat{\bm{\Sigma}},\tilde{\pi}_0,\tilde{\pi}_1)-(\hat{\bm{\mu}}_1-\hat{\bm{\mu}}_0)^T\left(\hat{\bm{\Sigma}}+\frac{1}{\zeta_{\hat{\bm{\Sigma}}}(\bm{\Sigma})}\textbf{I}_p\right)^{-1}\left(\bm{\mu}_0-\frac{\hat{\bm{\mu}}_0+\hat{\bm{\mu}}_1}{2}\right)-\text{ln}\frac{\pi_1}{\pi_0}&\xrightarrow{\text{a.s.}}0\label{first4}\\
     m_1(\hat{\bm{\mu}}_0,\hat{\bm{\mu}}_1,\hat{\bm{\Sigma}},\tilde{\pi}_0,\tilde{\pi}_1)-(\hat{\bm{\mu}}_1-\hat{\bm{\mu}}_0)^T\left(\hat{\bm{\Sigma}}+\frac{1}{\zeta_{\hat{\bm{\Sigma}}}(\bm{\Sigma})}\textbf{I}_p\right)^{-1}\left(\bm{\mu}_1-\frac{\hat{\bm{\mu}}_0+\hat{\bm{\mu}}_1}{2}\right)-\text{ln}\frac{\pi_1}{\pi_0}&\xrightarrow{\text{a.s.}}0\label{second4}\\
     \sigma^2(\hat{\bm{\mu}}_0,\hat{\bm{\mu}}_1,\hat{\bm{\Sigma}})-(\hat{\bm{\mu}}_1-\hat{\bm{\mu}}_0)^T\left(\hat{\bm{\Sigma}}+\frac{1}{\zeta_{\hat{\bm{\Sigma}}}(\bm{\Sigma})}\textbf{I}_p\right)^{-1}\bm{\Sigma}\left(\hat{\bm{\Sigma}}+\frac{1}{\zeta_{\hat{\bm{\Sigma}}}(\bm{\Sigma})}\textbf{I}_p\right)^{-1}(\hat{\bm{\mu}}_1-\hat{\bm{\mu}}_0)&\xrightarrow{\text{a.s.}}0\label{third4}
 \end{align}
 The sample means in each of the intermediate convergence expressions can be substituted by
 \begin{equation}
    \hat{\bm{\mu}}_0={\bm{\mu}}_0+\frac{\bm{\Sigma}^{1/2}\textbf{Z}_0\textbf{1}}{n_0}
\end{equation}
\begin{equation}
    \hat{\bm{\mu}}_1={\bm{\mu}}_1+\frac{\bm{\Sigma}^{1/2}\textbf{Z}_1\textbf{1}}{n_1}
\end{equation}
and the expectation over $\textbf{Z}_0\textbf{1}$ and $\textbf{Z}_1\textbf{1}$ taken as in Section \ref{appA2}. By using the fact that  $\textbf{Z}_0\textbf{1}$ and $\textbf{Z}_1\textbf{1}$ are independent of $\hat{\bm{\Sigma}}$, this step results in the same expressions as in Theorem 3 except that they contain the sample covariance instead of the true covariance. To these expressions, we then apply the same steps as in Section \ref{appA3} to remove the randomness coming from $\hat{\bm{\Sigma}}$. This yields the final DEs for the class-conditional discriminant statistics presented in Theorem 5. The error DE $\bar{\varepsilon}_{\hat{\bm{\mu}}_0,\hat{\bm{\mu}}_1,\hat{\bm{\Sigma}}}$ then follows directly from Lemma 2.

\section{Proof of Theorem 6}\label{app:appB}
To construct $\hat{\varepsilon}$, we construct the G-estimators $\hat{m}_0$, $\hat{m}_1$, and $\hat{\sigma}^2$ as outlined in Section \ref{gest}, Lemma 2. To do this, we make use of the intermediate covergence relations for each of the discriminant statistics with respect to the random projection given by \eqref{inter1}, \eqref{inter2}, and \eqref{inter3} in Section \ref{firstDE}. These converge to their respective DEs. These are convenient to work with because the expectation term involving $\textbf{R}$ has already been dealt with, but at the same time, the sample statistics are still intact. We manipulate these intermediate expressions so that they are functions of the sample statistics only. Section \ref{mean} and Section \ref{variance} present the derivations of the G-estimators of the class-conditional discriminant means and variance respectively.
\subsection{G-estimator of the class-conditional discriminant means}\label{mean}
To derive the G-estimator $\hat{m}_0$ of $m_0(\hat{\bm{\mu}}_0,\hat{\bm{\mu}}_1,\hat{\bm{\Sigma}},\tilde{\pi}_0,\tilde{\pi}_1)$, we make use of \eqref{inter1} with $\tilde{\bm{\mu}}_0=\hat{\bm{\mu}}_0$, $\tilde{\bm{\mu}}_1=\hat{\bm{\mu}}_1$, and $\tilde{\bm{\Sigma}}=\hat{\bm{\Sigma}}$, which is an intermediate stage within the convergence from $m_0(\hat{\bm{\mu}}_0,\hat{\bm{\mu}}_1,\hat{\bm{\Sigma}},\tilde{\pi}_0,\tilde{\pi}_1)$ to $\bar{m}_{0,\hat{\bm{\mu}}_0,\hat{\bm{\mu}}_1,\hat{\bm{\Sigma}}}$ and so satisfies
\begin{equation}
   (\hat{\bm{\mu}}_1-\hat{\bm{\mu}}_0)^T\left(\hat{\bm{\Sigma}}+\frac{1}{\lim\limits_{\gamma\rightarrow 0}\delta({\hat{\bm{\Sigma}}})}\textbf{I}_p\right)^{-1}\left(\bm{\mu}_0-\frac{\hat{\bm{\mu}}_0+\hat{\bm{\mu}}_1}{2}\right)+\text{ln}\frac{\pi_1}{\pi_0}- \bar{m}_{0,\hat{\bm{\mu}}_0,\hat{\bm{\mu}}_1,\hat{\bm{\Sigma}}}\xrightarrow{\text{a.s.}}0\label{gm0}
\end{equation}
First, we derive the quantity $\lim\limits_{\gamma\rightarrow 0}\delta({\hat{\bm{\Sigma}}})$ in terms of the sample covariance.

 Using the definitions of $\delta(\hat{\bm{\Sigma}})$ and $\tilde{\delta}(\hat{\bm{\Sigma}})$ in Lemma 4 and the eigendecomposition of $\hat{\bm{\Sigma}}$ as ${\hat{\bm{\Sigma}}}={\textbf{U}}\textbf{D}{\textbf{U}}^T$, we obtain
\begin{equation}
    \delta(\hat{\bm{\Sigma}})=\frac{1}{\gamma+\frac{1}{d}\text{tr}\left\{\textbf{D}\left(\delta(\hat{\bm{\Sigma}})\textbf{D}+\textbf{I}_p\right)^{-1}\right\}}
\end{equation}
 Taking $\lim\limits_{\gamma\rightarrow 0}$ of both sides and rearranging results in
\begin{equation}
    1-\frac{1}{d}\text{tr}\left\{\textbf{D}\left(\textbf{D}+\frac{1}{\lim\limits_{\gamma\rightarrow 0}\delta(\hat{\bm{\Sigma}})}\textbf{I}_p\right)^{-1}\right\}=0\label{a11}
\end{equation}
Denote  $\lim\limits_{\gamma\rightarrow 0}\delta({\hat{\bm{\Sigma}}})$ by $\zeta_{\hat{\bm{\Sigma}}}(\hat{\bm{\Sigma}})$, where the subscript refers to the fact that we have set $\tilde{\bm{\Sigma}}=\hat{\bm{\Sigma}}$ and $(\hat{\bm{\Sigma}})$ refers to the fact that this quantity is derived in such a way that it is a function of the sample covariance. Define the function $f(x)$ as
\begin{align}
    f(x)&=1-\frac{1}{d}\text{tr}\left\{\textbf{D}\left(\textbf{D}+\frac{1}{x}\textbf{I}_p\right)^{-1}\right\}\label{changeme}\\
    &=1-\frac{1}{d}\sum_{i=1}^{\text{rank}(\hat{\bm{\Sigma}})}\frac{\lambda_i(\hat{\bm{\Sigma}})}{\frac{1}{x}+\lambda_i(\hat{\bm{\Sigma}})}\label{looky}
\end{align}
From \eqref{looky}, as $x\rightarrow 0$, $f(x)$ tends to $1$ and as $x\rightarrow \infty$, $f(x)$ tends to $1-\frac{\text{rank}(\hat{\bm{\Sigma}})}{d}$, which is negative when $d<\text{rank}(\hat{\bm{\Sigma}})$. Thus $\zeta_{\hat{\bm{\Sigma}}}(\hat{\bm{\Sigma}})$ is the unique root of $f(x)$. Strategic multiplication by $\textbf{U}^T\textbf{U}$ in \eqref{changeme} puts $f(x)$ in terms of $\hat{\bm{\Sigma}}$ as presented in Theorem 5. The estimator of the term $\text{ln}\frac{\pi_1}{\pi_0}$ is simply $\text{ln}\frac{\hat{\pi}_1}{\hat{\pi}_0}$. Thus we have
\begin{equation}
   (\hat{\bm{\mu}}_1-\hat{\bm{\mu}}_0)^T\left(\hat{\bm{\Sigma}}+\frac{1}{\zeta_{\hat{\bm{\Sigma}}}(\hat{\bm{\Sigma}})}\textbf{I}_p\right)^{-1}\left(\bm{\mu}_0-\frac{\hat{\bm{\mu}}_0+\hat{\bm{\mu}}_1}{2}\right)+\text{ln}\frac{\hat{\pi}_1}{\hat{\pi}_0}- \bar{m}_{0,\hat{\bm{\mu}}_0,\hat{\bm{\mu}}_1,\hat{\bm{\Sigma}}}\xrightarrow{\text{a.s.}}0\label{g2}
\end{equation}
We now find the G-estimator of the first term in \eqref{g2} by expressing it as follows
\begin{align}
    (\hat{\bm{\mu}}_1-\hat{\bm{\mu}}_0)^T\left(\frac{1}{\zeta_{\hat{\bm{\Sigma}}}(\hat{\bm{\Sigma}})}\textbf{I}_p+\hat{\bm{\Sigma}}\right)^{-1}\left(\bm{\mu}_0-\frac{\hat{\bm{\mu}}_0+\hat{\bm{\mu}}_1}{2}\right)
    &=(\hat{\bm{\mu}}_1-\hat{\bm{\mu}}_0)^T\left(\frac{1}{\zeta_{\hat{\bm{\Sigma}}}(\hat{\bm{\Sigma}})}\textbf{I}_p+\hat{\bm{\Sigma}}\right)^{-1}\left(\hat{\bm{\mu}}_0-\frac{\hat{\bm{\mu}}_0+\hat{\bm{\mu}}_1}{2}\right)\nonumber\\&\hspace{50px}+(\hat{\bm{\mu}}_1-\hat{\bm{\mu}}_0)^T\left(\frac{1}{\zeta_{\hat{\bm{\Sigma}}}(\hat{\bm{\Sigma}})}\textbf{I}_p+\hat{\bm{\Sigma}}\right)^{-1}\left({\bm{\mu}}_0-\hat{\bm{\mu}}_0\right)\label{g3}
\end{align}
The first term on the right-hand side is the plugin estimator. The second term is its correction. It involves $\bm{\mu}_0$. Our aim now is to find the G-estimator of this correction term.

By substituting $\hat{\bm{\mu}}_0={\bm{\mu}}_0+\frac{\bm{\Sigma}^{1/2}\textbf{Z}_0\textbf{1}}{n_0}$ and $\hat{\bm{\mu}}_1={\bm{\mu}}_1+\frac{\bm{\Sigma}^{1/2}\textbf{Z}_1\textbf{1}}{n_1}$, where $\textbf{Z}_0\in\mathbb{R}^{p\times n_0}$ and $\textbf{Z}_1\in\mathbb{R}^{p\times n_1}$ have i.i.d. Gaussian zero-mean and unit variance entries, taking the expectation over $\textbf{Z}_0\textbf{1}$ and $\textbf{Z}_1\textbf{1}$, and making use of the fact that $\hat{\bm{\Sigma}}$ is independent of $\textbf{Z}_0$ and $\textbf{Z}_1$ and that $\frac{\textbf{Z}_0\textbf{1}}{n_0}\sim\mathcal{N}\left(\textbf{0}_p,\frac{1}{n_0}\textbf{I}_p\right)$ and $\frac{\textbf{Z}_1\textbf{1}}{n_1}\sim\mathcal{N}\left(\textbf{0}_p,\frac{1}{n_1}\textbf{I}_p\right)$, this simplifies as follows
\begin{align}
    \mathbb{E}_{\textbf{Z}_0\textbf{1},\textbf{Z}_1\textbf{1}}\left[(\hat{\bm{\mu}}_1-\hat{\bm{\mu}}_0)^T\left(\frac{1}{\zeta_{\hat{\bm{\Sigma}}}(\hat{\bm{\Sigma}})}\textbf{I}_p+\hat{\bm{\Sigma}}\right)^{-1}\left({\bm{\mu}}_0-\hat{\bm{\mu}}_0\right)\right]
    &=\frac{1}{n_0}\text{tr}\left\{\bm{\Sigma}\left(\frac{1}{\zeta_{\hat{\bm{\Sigma}}}(\hat{\bm{\Sigma}})}\textbf{I}_p+\hat{\bm{\Sigma}}\right)^{-1}\right\}\label{party}
\end{align}
by which we claim the convergence
\begin{equation}
    (\hat{\bm{\mu}}_1-\hat{\bm{\mu}}_0)^T\left(\frac{1}{\zeta_{\hat{\bm{\Sigma}}}(\hat{\bm{\Sigma}})}\textbf{I}_p+\hat{\bm{\Sigma}}\right)^{-1}\left({\bm{\mu}}_0-\hat{\bm{\mu}}_0\right)\asymp \frac{1}{n_0}\text{tr}\left\{\bm{\Sigma}\left(\frac{1}{\zeta_{\hat{\bm{\Sigma}}}(\hat{\bm{\Sigma}})}\textbf{I}_p+\hat{\bm{\Sigma}}\right)^{-1}\right\}\label{theOG}
\end{equation}
To find the G-estimator of the right-hand side of \eqref{theOG}, the strategy we use is to substitute the sample covariance $\hat{\bm{\Sigma}}$ for the true covariance $\bm{\Sigma}$ and express this asymptotically in terms of the original quantity. We then replace every occurrence of $\hat{\bm{\Sigma}}$ by the identically distributed quantity $\frac{1}{n-2}\bm{\Sigma}^{1/2}\bar{\textbf{Y}}\bar{\textbf{Y}}^T\bm{\Sigma}^{1/2}$ where $\bar{\textbf{Y}}\in\mathbb{R}^{p\times (n-2)}$ has i.i.d. columns distributed as $\mathcal{N}(\textbf{0},\textbf{I})$.

Note that $\frac{1}{n-2}\bm{\Sigma}^{1/2}\bar{\textbf{Y}}\bar{\textbf{Y}}^T\bm{\Sigma}^{1/2}$ can be expressed as $\frac{1}{n-2}\bm{\Sigma}^{1/2}\bar{\textbf{Y}}\bar{\textbf{Y}}^T\bm{\Sigma}^{1/2}=\frac{1}{n-2}\sum_{j=1}^{n-2}\tilde{\textbf{y}}_j\tilde{\textbf{y}}_j^T$ where $\tilde{\textbf{Y}}:=\bm{\Sigma}^{1/2}\bar{\textbf{Y}}$ and $\tilde{\textbf{y}}_j$ is the $j^{\text{th}}$ column of $\tilde{\textbf{Y}}$. Now define \begin{equation}\textbf{Q}:=\left(\frac{1}{\zeta_{\hat{\bm{\Sigma}}}(\hat{\bm{\Sigma}})}\textbf{I}_p+\frac{1}{n-2}\sum_{j=1}^{n-2}\tilde{\textbf{y}}_j\tilde{\textbf{y}}_j^T\right)^{-1}
\end{equation}
We have
\begin{align}
    \frac{1}{n_0}\text{tr}\left\{\hat{\bm{\Sigma}}\left(\frac{1}{\zeta_{\hat{\bm{\Sigma}}}(\hat{\bm{\Sigma}})}\textbf{I}_p+\hat{\bm{\Sigma}}\right)^{-1}\right\}=\frac{1}{n}\sum_{i=1}^n\frac{1}{n_0}\text{tr}\left\{\tilde{\textbf{y}}_i^T\textbf{Q}_i\tilde{\textbf{y}}_i\right\}\\
    =\frac{1}{n}\sum_{i=1}^n\frac{\frac{1}{n_0}\text{tr}\left\{\tilde{\textbf{y}}_i^T\textbf{Q}\tilde{\textbf{y}}_i\right\}}{1+\frac{1}{n}\tilde{\textbf{y}}_i^T\textbf{Q}\tilde{\textbf{y}}_i}\label{OG2}
\end{align}
where ${\textbf{Q}}_i$ is defined as ${\textbf{Q}}_i:=\left(\frac{1}{\zeta_{\hat{\bm{\Sigma}}}(\hat{\bm{\Sigma}})}\textbf{I}_p+\frac{1}{n}\sum_{j\ne i}\tilde{\textbf{y}}_j\tilde{\textbf{y}}_j^T\right)^{-1}$ and the last line follows from applying the matrix inversion lemma (see \cite{muller2016random}) to either $\tilde{\textbf{y}}_i^T{\textbf{Q}}_i$ or ${\textbf{Q}}_i\tilde{\textbf{y}}_i$. Noting that $\mathbb{E}_{\tilde{\textbf{y}}_i}[\frac{1}{n}\tilde{\textbf{y}}_i^T{\textbf{Q}}_i\tilde{\textbf{y}}_i]=\frac{1}{n}\text{tr}\left\{\bm{\Sigma}{\textbf{Q}}_i\right\}$, it can be shown that
\begin{equation}
    \frac{1}{n}\sum_{i=1}^n\frac{\frac{1}{n_0}\text{tr}\left\{\tilde{\textbf{y}}_i^T\textbf{Q}\tilde{\textbf{y}}_i\right\}}{1+\frac{1}{n}\tilde{\textbf{y}}_i^T\textbf{Q}\tilde{\textbf{y}}_i}\asymp \frac{1}{n}\sum_{i=1}^n\frac{\frac{1}{n_0}\text{tr}\left\{\tilde{\textbf{y}}_i^T\textbf{Q}\tilde{\textbf{y}}_i\right\}}{1+\frac{1}{n}\text{tr}\left\{\bm{\Sigma}{\textbf{Q}}_i\right\}}
\end{equation}
by bounding the error in making this substitution by a decaying function of $d$ as before. By taking the expectation over $\tilde{\textbf{y}}_i$ followed by the rank-one perturbation lemma (see \cite{muller2016random}) applied to each term involving $\textbf{Q}_i$, we have the convergence
\begin{equation}
\frac{1}{n}\sum_{i=1}^n\frac{\frac{1}{n_0}\text{tr}\left\{\tilde{\textbf{y}}_i^T\textbf{Q}\tilde{\textbf{y}}_i\right\}}{1+\frac{1}{n}\text{tr}\left\{\bm{\Sigma}{\textbf{Q}}_i\right\}}\asymp \frac{\frac{1}{n_0}\text{tr}\left\{\bm{\Sigma}\textbf{Q}\right\}}{1+\frac{1}{n}\text{tr}\left\{\bm{\Sigma}\textbf{Q}\right\}}
\end{equation}
We can thus claim
\begin{equation}
     \frac{1}{n_0}\text{tr}\left\{\hat{\bm{\Sigma}}\left(\frac{1}{\zeta_{\hat{\bm{\Sigma}}}(\hat{\bm{\Sigma}})}\textbf{I}_p+\hat{\bm{\Sigma}}\right)^{-1}\right\}\asymp\frac{\frac{1}{n_0}\text{tr}\left\{\bm{\Sigma}\textbf{Q}\right\}}{1+\frac{1}{n}\text{tr}\left\{\bm{\Sigma}\textbf{Q}\right\}}
\end{equation}
which after rearranging yields the G-estimator
\begin{equation}
    \frac{\frac{1}{n_0}\text{tr}\left\{\hat{\bm{\Sigma}}\left(\hat{\bm{\Sigma}}+\frac{1}{\zeta_{\hat{\bm{\Sigma}}}(\hat{\bm{\Sigma}})}\textbf{I}_p\right)^{-1}\right\}}{1-\frac{1}{n}\text{tr}\left\{\hat{\bm{\Sigma}}\left(\hat{\bm{\Sigma}}+\frac{1}{\zeta_{\hat{\bm{\Sigma}}}(\hat{\bm{\Sigma}})}\textbf{I}_p\right)^{-1}\right\}}\asymp\frac{1}{n_0}\text{tr}\left\{\bm{\Sigma}\textbf{Q}\right\}\label{veryIMP}
\end{equation}
From \eqref{theOG} we can see that this is the G-estimator of the bias correction term and thus the expression for $\hat{m}_0$ follows from \eqref{g3}. A similar derivation yields the expression for $\hat{m}_1$.

\subsection{G-estimator of the class-conditional discriminant variance}\label{variance}
To derive the G-estimator $\hat{\sigma}^2$ of $\sigma^2(\hat{\bm{\mu}}_0,\hat{\bm{\mu}}_1,\hat{\bm{\Sigma}})$, we make use of \eqref{inter3} with $\tilde{\bm{\mu}}_0=\hat{\bm{\mu}}_0$, $\tilde{\bm{\mu}}_1=\hat{\bm{\mu}}_1$, and $\tilde{\bm{\Sigma}}=\hat{\bm{\Sigma}}$, which is an intermediate stage within the convergence from $\sigma^2(\hat{\bm{\mu}}_0,\hat{\bm{\mu}}_1,\hat{\bm{\Sigma}})$ to $\bar{\sigma}^2_{\hat{\bm{\mu}}_0,\hat{\bm{\mu}}_1,\hat{\bm{\Sigma}}}$ and so satisfies
\begin{equation}
  (\hat{\bm{\mu}}_1-\hat{\bm{\mu}}_0)^T\left(\hat{\bm{\Sigma}}+\frac{1}{\lim\limits_{\gamma\rightarrow 0}\delta({\hat{\bm{\Sigma}}})}\textbf{I}_p\right)^{-1}\bm{\Sigma}\left(\hat{\bm{\Sigma}}+\frac{1}{\lim\limits_{\gamma\rightarrow 0}\delta({\hat{\bm{\Sigma}}})}\textbf{I}_p\right)^{-1}(\hat{\bm{\mu}}_1-\hat{\bm{\mu}}_0)- \bar{\sigma}^2_{\hat{\bm{\mu}}_0,\hat{\bm{\mu}}_1,\hat{\bm{\Sigma}}}\xrightarrow{\text{a.s.}}0
\end{equation}
The $\lim\limits_{\gamma\rightarrow 0}\delta({\hat{\bm{\Sigma}}})$ term is derived as in Section \ref{mean} and is denoted by $\zeta_{\hat{\bm{\Sigma}}}(\hat{\bm{\Sigma}})$. Now we find the G-estimator of the first term. Again, the strategy we use is to substitute the sample covariance $\hat{\bm{\Sigma}}$ for the true covariance $\bm{\Sigma}$ and try to express this asymptotically in terms of the original quantity. Using the notation defined in Section \ref{mean}, we have
\begin{align}
     (\hat{\bm{\mu}}_1-\hat{\bm{\mu}}_0)^T\left(\hat{\bm{\Sigma}}+\frac{1}{\zeta_{\hat{\bm{\Sigma}}}(\hat{\bm{\Sigma}})}\textbf{I}_p\right)^{-1}\hat{\bm{\Sigma}}\left(\hat{\bm{\Sigma}}+\frac{1}{\zeta_{\hat{\bm{\Sigma}}}(\hat{\bm{\Sigma}})}\textbf{I}_p\right)^{-1}(\hat{\bm{\mu}}_1-\hat{\bm{\mu}}_0)  &\asymp\frac{1}{n}\sum_{i=1}^n(\hat{\bm{\mu}}_1-\hat{\bm{\mu}}_0)^T{\textbf{Q}}\tilde{\textbf{y}}_i\tilde{\textbf{y}}_i^T{\textbf{Q}}(\hat{\bm{\mu}}_1-\hat{\bm{\mu}}_0)\nonumber\\
     &=\frac{1}{n}\sum_{i=1}^n\frac{(\hat{\bm{\mu}}_1-\hat{\bm{\mu}}_0)^T{\textbf{Q}}_i\tilde{\textbf{y}}_i\tilde{\textbf{y}}_i^T{\textbf{Q}}_i(\hat{\bm{\mu}}_1-\hat{\bm{\mu}}_0)}{\left(1+\frac{1}{n}\tilde{\textbf{y}}_i^T\textbf{Q}_i\tilde{\textbf{y}}_i\right)^2}\label{letmego}
\end{align}
where the last line follows from applying the matrix inversion lemma (see \cite{muller2016random}) to each of ${\textbf{Q}}_i\tilde{\textbf{y}}_i$ and $\tilde{\textbf{y}}_i^T{\textbf{Q}}_i$. It can be shown that
\begin{equation}
    \frac{1}{n}\sum_{i=1}^n\frac{(\hat{\bm{\mu}}_1-\hat{\bm{\mu}}_0)^T{\textbf{Q}}_i\tilde{\textbf{y}}_i\tilde{\textbf{y}}_i^T{\textbf{Q}}_i(\hat{\bm{\mu}}_1-\hat{\bm{\mu}}_0)}{\left(1+\frac{1}{n}\tilde{\textbf{y}}_i^T\textbf{Q}_i\tilde{\textbf{y}}_i\right)^2}\asymp \frac{1}{n}\sum_{i=1}^n\frac{(\hat{\bm{\mu}}_1-\hat{\bm{\mu}}_0)^T{\textbf{Q}}_i\tilde{\textbf{y}}_i\tilde{\textbf{y}}_i^T{\textbf{Q}}_i(\hat{\bm{\mu}}_1-\hat{\bm{\mu}}_0)}{\left(1+\frac{1}{n}\text{tr}\left\{\bm{\Sigma}{\textbf{Q}}_i\right\}\right)^2}
\end{equation}
due to the convergence $\frac{1}{n}\tilde{\textbf{y}}_i^T\textbf{Q}_i\tilde{\textbf{y}}_i\asymp\frac{1}{n}\text{tr}\left\{\bm{\Sigma}{\textbf{Q}}_i\right\}$.
By taking the expectation over $\tilde{\textbf{y}}_i$ followed by the rank-one perturbation lemma (see \cite{muller2016random}) applied to the $\frac{1}{n}\text{tr}\left\{\bm{\Sigma}{\textbf{Q}}_i\right\}$ term in the denominator, we have the convergence
\begin{equation}
 \frac{1}{n}\sum_{i=1}^n\frac{(\hat{\bm{\mu}}_1-\hat{\bm{\mu}}_0)^T{\textbf{Q}}_i\tilde{\textbf{y}}_i\tilde{\textbf{y}}_i^T{\textbf{Q}}_i(\hat{\bm{\mu}}_1-\hat{\bm{\mu}}_0)}{\left(1+\frac{1}{n}\text{tr}\left\{\bm{\Sigma}{\textbf{Q}}_i\right\}\right)^2}\asymp \frac{1}{n}\sum_{i=1}^n\frac{(\hat{\bm{\mu}}_1-\hat{\bm{\mu}}_0)^T{\textbf{Q}}_i\bm{\Sigma}{\textbf{Q}}_i(\hat{\bm{\mu}}_1-\hat{\bm{\mu}}_0)}{\left(1+\frac{1}{n}\text{tr}\left\{\bm{\Sigma}{\textbf{Q}}\right\}\right)^2}
\end{equation}
This can be expressed as
\begin{align}
    \frac{1}{n}\sum_{i=1}^n\frac{(\hat{\bm{\mu}}_1-\hat{\bm{\mu}}_0)^T{\textbf{Q}}_i\bm{\Sigma}{\textbf{Q}}_i(\hat{\bm{\mu}}_1-\hat{\bm{\mu}}_0)}{\left(1+\frac{1}{n}\text{tr}\left\{\bm{\Sigma}{\textbf{Q}}\right\}\right)^2}
    &=\frac{1}{n}\sum_{i=1}^n\frac{(\hat{\bm{\mu}}_1-\hat{\bm{\mu}}_0)^T\left({\textbf{Q}}_i-{\textbf{Q}}\right)\bm{\Sigma}{\textbf{Q}}_i(\hat{\bm{\mu}}_1-\hat{\bm{\mu}}_0)}{\left(1+\frac{1}{n}\text{tr}\left\{\bm{\Sigma}{\textbf{Q}}\right\}\right)^2}\nonumber\\
    &\hspace{15px}+\frac{1}{n}\sum_{i=1}^n\frac{(\hat{\bm{\mu}}_1-\hat{\bm{\mu}}_0)^T{\textbf{Q}}\bm{\Sigma}{\textbf{Q}}(\hat{\bm{\mu}}_1-\hat{\bm{\mu}}_0)}{\left(1+\frac{1}{n}\text{tr}\left\{\bm{\Sigma}{\textbf{Q}}\right\}\right)^2}\nonumber\\
    &\hspace{15px}+\frac{1}{n}\sum_{i=1}^n\frac{(\hat{\bm{\mu}}_1-\hat{\bm{\mu}}_0)^T{\textbf{Q}}\bm{\Sigma}\left({\textbf{Q}}_i-{\textbf{Q}}\right)(\hat{\bm{\mu}}_1-\hat{\bm{\mu}}_0)}{\left(1+\frac{1}{n}\text{tr}\left\{\bm{\Sigma}{\textbf{Q}}\right\}\right)^2}
\end{align}
It can be shown that the terms involving $\left({\textbf{Q}}_i-{\textbf{Q}}\right)$ go to zero and so we have, in conjunction with \eqref{letmego},
\begin{align}
    (\hat{\bm{\mu}}_1-\hat{\bm{\mu}}_0)^T\left(\hat{\bm{\Sigma}}+\frac{1}{\zeta_{\hat{\bm{\Sigma}}}(\hat{\bm{\Sigma}})}\textbf{I}_p\right)^{-1}\hat{\bm{\Sigma}}\left(\hat{\bm{\Sigma}}+\frac{1}{\zeta_{\hat{\bm{\Sigma}}}(\hat{\bm{\Sigma}})}\textbf{I}_p\right)^{-1}(\hat{\bm{\mu}}_1-\hat{\bm{\mu}}_0)  &\asymp\frac{(\hat{\bm{\mu}}_1-\hat{\bm{\mu}}_0)^T{\textbf{Q}}\bm{\Sigma}{\textbf{Q}}(\hat{\bm{\mu}}_1-\hat{\bm{\mu}}_0)}{\left(1+\frac{1}{n}\text{tr}\left\{\bm{\Sigma}{\textbf{Q}}\right\}\right)^2}
\end{align}
from which we obtain
\begin{align}
    \left(1+\frac{1}{n}\text{tr}\left\{\bm{\Sigma}\left(\hat{\bm{\Sigma}}+\frac{1}{\zeta_{\hat{\bm{\Sigma}}}(\hat{\bm{\Sigma}})}\textbf{I}_p\right)^{-1}\right\}\right)^2&(\hat{\bm{\mu}}_1-\hat{\bm{\mu}}_0)^T\left(\hat{\bm{\Sigma}}+\frac{1}{\zeta_{\hat{\bm{\Sigma}}}(\hat{\bm{\Sigma}})}\textbf{I}_p\right)^{-1}\hat{\bm{\Sigma}}\left(\hat{\bm{\Sigma}}+\frac{1}{\zeta_{\hat{\bm{\Sigma}}}(\hat{\bm{\Sigma}})}\textbf{I}_p\right)^{-1}(\hat{\bm{\mu}}_1-\hat{\bm{\mu}}_0)\nonumber\\  &\asymp(\hat{\bm{\mu}}_1-\hat{\bm{\mu}}_0)^T\left(\hat{\bm{\Sigma}}+\frac{1}{\zeta_{\hat{\bm{\Sigma}}}(\hat{\bm{\Sigma}})}\textbf{I}_p\right)^{-1}\bm{\Sigma}\left(\hat{\bm{\Sigma}}+\frac{1}{\zeta_{\hat{\bm{\Sigma}}}(\hat{\bm{\Sigma}})}\textbf{I}_p\right)^{-1}(\hat{\bm{\mu}}_1-\hat{\bm{\mu}}_0)
\end{align}
The expression for $\hat{\sigma}^2$ in Theorem 5 then follows from substituting the G-estimator of $\frac{1}{n}\text{tr}\left\{\bm{\Sigma}\left(\hat{\bm{\Sigma}}+\frac{1}{\zeta_{\hat{\bm{\Sigma}}}(\hat{\bm{\Sigma}})}\textbf{I}_p\right)^{-1}\right\}$ which can be deduced from \eqref{veryIMP}.

\bibliographystyle{IEEEtran}
\bibliography{references}

\begin{thebibliography}{10}
\providecommand{\url}[1]{#1}
\csname url@samestyle\endcsname
\providecommand{\newblock}{\relax}
\providecommand{\bibinfo}[2]{#2}
\providecommand{\BIBentrySTDinterwordspacing}{\spaceskip=0pt\relax}
\providecommand{\BIBentryALTinterwordstretchfactor}{4}
\providecommand{\BIBentryALTinterwordspacing}{\spaceskip=\fontdimen2\font plus
\BIBentryALTinterwordstretchfactor\fontdimen3\font minus
  \fontdimen4\font\relax}
\providecommand{\BIBforeignlanguage}[2]{{%
\expandafter\ifx\csname l@#1\endcsname\relax
\typeout{** WARNING: IEEEtran.bst: No hyphenation pattern has been}%
\typeout{** loaded for the language `#1'. Using the pattern for}%
\typeout{** the default language instead.}%
\else
\language=\csname l@#1\endcsname
\fi
#2}}
\providecommand{\BIBdecl}{\relax}
\BIBdecl

\bibitem{durrant2013random}
\BIBentryALTinterwordspacing
R.~J. Durrant and A.~Kab{\'a}n, ``Random projections as regularizers: Learning
  a linear discriminant ensemble from fewer observations than dimensions,'' in
  \emph{Proceedings of the Asian Conference on Machine Learning},
  vol.~29.\hskip 1em plus 0.5em minus 0.4em\relax JMLR, 2013, pp. 17--32.
  [Online]. Available:
  \url{http://jmlr.org/proceedings/papers/v29/Durrant13.html Google Scholar}
\BIBentrySTDinterwordspacing

\bibitem{lim2000comparison}
T.-S. Lim, W.-Y. Loh, and Y.-S. Shih, ``A comparison of prediction accuracy,
  complexity, and training time of thirty-three old and new classification
  algorithms,'' \emph{Machine learning}, vol.~40, no.~3, pp. 203--228, 2000.

\bibitem{giansante2003classification}
L.~Giansante, D.~Di~Vincenzo, and G.~Bianchi, ``Classification of monovarietal
  italian olive oils by unsupervised ({PCA}) and supervised ({LDA})
  chemometrics,'' \emph{Journal of the Science of Food and Agriculture},
  vol.~83, no.~9, pp. 905--911, 2003.

\bibitem{skrobot2007use}
V.~L. Skrobot, E.~V. Castro, R.~C. Pereira, V.~M. Pasa, and I.~C. Fortes, ``Use
  of principal component analysis ({PCA}) and linear discriminant analysis
  ({LDA}) in gas chromatographic ({GC}) data in the investigation of gasoline
  adulteration,'' \emph{Energy \& Fuels}, vol.~21, no.~6, pp. 3394--3400, 2007.

\bibitem{azcarate2017chemometric}
S.~M. Azcarate, R.~Gil, P.~Smichowski, M.~Savio, and J.~M. Cami{\~n}a,
  ``Chemometric application in foodomics: Nutritional quality parameters
  evaluation in milk-based infant formula,'' \emph{Microchemical Journal}, vol.
  130, pp. 1--6, 2017.

\bibitem{melucci2019checking}
D.~Melucci, A.~Zappi, L.~Bolelli, F.~Corvucci, G.~Serra, M.~Boi, F.-V.
  Grillenzoni, G.~Fedrizzi, S.~Menotta, and S.~Girotti, ``Checking syrup
  adulteration of honey using bioluminescent bacteria and chemometrics,''
  \emph{European Food Research and Technology}, vol. 245, no.~2, pp. 315--324,
  2019.

\bibitem{zhao2003face}
W.~Zhao, R.~Chellappa, P.~J. Phillips, and A.~Rosenfeld, ``Face recognition: A
  literature survey,'' \emph{ACM computing surveys (CSUR)}, vol.~35, no.~4, pp.
  399--458, 2003.

\bibitem{lu2003face}
J.~Lu, K.~N. Plataniotis, and A.~N. Venetsanopoulos, ``Face recognition using
  {LDA}-based algorithms,'' \emph{IEEE Transactions on Neural networks},
  vol.~14, no.~1, pp. 195--200, 2003.

\bibitem{kramer2018understanding}
R.~S. Kramer, A.~W. Young, and A.~M. Burton, ``Understanding face
  familiarity,'' \emph{Cognition}, vol. 172, pp. 46--58, 2018.

\bibitem{portillo2018view}
J.~Portillo-Portillo, R.~Leyva, V.~Sanchez, G.~Sanchez-Perez, H.~Perez-Meana,
  J.~Olivares-Mercado, K.~Toscano-Medina, and M.~Nakano-Miyatake, ``A
  view-invariant gait recognition algorithm based on a joint-direct linear
  discriminant analysis,'' \emph{Applied Intelligence}, vol.~48, no.~5, pp.
  1200--1217, 2018.

\bibitem{sharma2008cancer}
A.~Sharma and K.~K. Paliwal, ``Cancer classification by gradient {LDA}
  technique using microarray gene expression data,'' \emph{Data \& Knowledge
  Engineering}, vol.~66, no.~2, pp. 338--347, 2008.

\bibitem{paliwal2010improved}
K.~K. Paliwal and A.~Sharma, ``Improved direct {LDA} and its application to
  {DNA} microarray gene expression data,'' \emph{Pattern Recognition Letters},
  vol.~31, no.~16, pp. 2489--2492, 2010.

\bibitem{huerta2010hybrid}
E.~B. Huerta, B.~Duval, and J.-K. Hao, ``A hybrid {LDA} and genetic algorithm
  for gene selection and classification of microarray data,''
  \emph{Neurocomputing}, vol.~73, no. 13-15, pp. 2375--2383, 2010.

\bibitem{sharma2012filter}
A.~Sharma, S.~Imoto, and S.~Miyano, ``A filter based feature selection
  algorithm using null space of covariance matrix for {DNA} microarray gene
  expression data,'' \emph{Current Bioinformatics}, vol.~7, no.~3, pp.
  289--294, 2012.

\bibitem{li2018fisher}
W.~Li, B.~Liao, W.~Zhu, M.~Chen, Z.~Li, X.~Wei, L.~Peng, G.~Huang, L.~Cai, and
  H.~Chen, ``Fisher discrimination regularized robust coding based on a local
  center for tumor classification,'' \emph{Scientific reports}, vol.~8, no.~1,
  p. 9152, 2018.

\bibitem{dudoit2002comparison}
S.~Dudoit, J.~Fridlyand, and T.~P. Speed, ``Comparison of discrimination
  methods for the classification of tumors using gene expression data,''
  \emph{Journal of the American statistical association}, vol.~97, no. 457, pp.
  77--87, 2002.

\bibitem{mai2013review}
Q.~Mai, ``A review of discriminant analysis in high dimensions,'' \emph{Wiley
  Interdisciplinary Reviews: Computational Statistics}, vol.~5, no.~3, pp.
  190--197, 2013.

\bibitem{RAUDYS1998385}
\BIBentryALTinterwordspacing
S.~Raudys and R.~P. Duin, ``Expected classification error of the fisher linear
  classifier with pseudo-inverse covariance matrix,'' \emph{Pattern Recognition
  Letters}, vol.~19, no.~5, pp. 385 -- 392, 1998. [Online]. Available:
  \url{http://www.sciencedirect.com/science/article/pii/S0167865598000166}
\BIBentrySTDinterwordspacing

\bibitem{Sharma2015}
\BIBentryALTinterwordspacing
A.~Sharma and K.~K. Paliwal, ``Linear discriminant analysis for the small
  sample size problem: An overview,'' \emph{International Journal of Machine
  Learning and Cybernetics}, vol.~6, no.~3, pp. 443--454, Jun 2015. [Online].
  Available: \url{https://doi.org/10.1007/s13042-013-0226-9}
\BIBentrySTDinterwordspacing

\bibitem{Bingham:2001:RPD:502512.502546}
\BIBentryALTinterwordspacing
E.~Bingham and H.~Mannila, ``Random projection in dimensionality reduction:
  Applications to image and text data,'' in \emph{Proceedings of the Seventh
  ACM SIGKDD International Conference on Knowledge Discovery and Data Mining},
  ser. KDD '01.\hskip 1em plus 0.5em minus 0.4em\relax New York, NY, USA: ACM,
  2001, pp. 245--250. [Online]. Available:
  \url{http://doi.acm.org/10.1145/502512.502546}
\BIBentrySTDinterwordspacing

\bibitem{5597392}
R.~J. Durrant and A.~Kab{\'a}n, ``A bound on the performance of {LDA} in
  randomly projected data spaces,'' in \emph{2010 20th International Conference
  on Pattern Recognition}, Aug 2010, pp. 4044--4047.

\bibitem{cannings2017random}
T.~I. Cannings and R.~J. Samworth, ``Random-projection ensemble
  classification,'' \emph{Journal of the Royal Statistical Society: Series B
  (Statistical Methodology)}, vol.~79, no.~4, pp. 959--1035, 2017.

\bibitem{Sammut:2017:EML:3153490}
C.~Sammut and G.~I. Webb, \emph{Encyclopedia of Machine Learning and Data
  Mining}, 2nd~ed.\hskip 1em plus 0.5em minus 0.4em\relax Springer Publishing
  Company, Incorporated, 2017.

\bibitem{bengio2004no}
Y.~Bengio and Y.~Grandvalet, ``No unbiased estimator of the variance of k-fold
  cross-validation,'' \emph{Journal of machine learning research}, vol.~5, no.
  Sep, pp. 1089--1105, 2004.

\bibitem{dougherty2001small}
E.~R. Dougherty, ``Small sample issues for microarray-based classification,''
  \emph{Comparative and Functional Genomics}, vol.~2, no.~1, pp. 28--34, 2001.

\bibitem{domingos2012few}
P.~M. Domingos, ``A few useful things to know about machine learning.''
  \emph{Commun. {ACM}}, vol.~55, no.~10, pp. 78--87, 2012.

\bibitem{zollanvari2015generalized}
A.~Zollanvari and E.~R. Dougherty, ``Generalized consistent error estimator of
  linear discriminant analysis,'' \emph{IEEE transactions on signal
  processing}, vol.~63, no.~11, pp. 2804--2814, 2015.

\bibitem{elkhalil2017large}
K.~Elkhalil, A.~Kammoun, R.~Couillet, T.~Y. Al-Naffouri, and M.-S. Alouini, ``A
  large dimensional study of regularized discriminant analysis classifiers,''
  \emph{arXiv preprint arXiv:1711.00382}, 2017.

\bibitem{yang2018regularized}
X.~Yang, K.~Elkhalil, A.~Kammoun, T.~Y. Al-Naffouri, and M.-S. Alouini,
  ``Regularized discriminant analysis: A large dimensional study,'' in
  \emph{2018 IEEE International Symposium on Information Theory (ISIT)}.\hskip
  1em plus 0.5em minus 0.4em\relax IEEE, 2018, pp. 536--540.

\bibitem{8683386}
K.~Elkhalil, A.~Kammoun, R.~Calderbank, T.~Y. Al-Naffouri, and M.-S. Alouini,
  ``Asymptotic performance of linear discriminant analysis with random
  projections,'' in \emph{ICASSP 2019 - 2019 IEEE International Conference on
  Acoustics, Speech and Signal Processing (ICASSP)}, May 2019, pp. 3472--3476.

\bibitem{durrant2013learning}
R.~J. Durrant, ``Learning in high dimensions with projected linear
  discriminants,'' Ph.D. dissertation, University of Birmingham, 2013.

\bibitem{wang2018dimension}
C.~Wang and B.~Jiang, ``On the dimension effect of regularized linear
  discriminant analysis,'' \emph{Electronic Journal of Statistics}, vol.~12,
  no.~2, pp. 2709--2742, 2018.

\bibitem{hastie1995penalized}
T.~Hastie, A.~Buja, and R.~Tibshirani, ``Penalized discriminant analysis,''
  \emph{The Annals of Statistics}, pp. 73--102, 1995.

\bibitem{mesejo2016computer}
P.~Mesejo, D.~Pizarro, A.~Abergel, O.~Rouquette, S.~Beorchia, L.~Poincloux, and
  A.~Bartoli, ``Computer-aided classification of gastrointestinal lesions in
  regular colonoscopy,'' \emph{IEEE transactions on medical imaging}, vol.~35,
  no.~9, pp. 2051--2063, 2016.

\bibitem{singh2002gene}
D.~Singh \emph{et~al.}, ``Gene expression correlates of clinical prostate
  cancer behavior,'' \emph{Cancer cell}, vol.~1, no.~2, pp. 203--209, 2002.

\bibitem{muller2016random}
A.~Muller and M.~Debbah, ``Random matrix theory tutorial-{I}ntroduction to
  deterministic equivalents,'' \emph{Traitement du signal}, vol.~33, no. 2-3,
  pp. 223--248, 2016.

\bibitem{hachem2013bilinear}
W.~Hachem, P.~Loubaton, J.~Najim, P.~Vallet \emph{et~al.}, ``On bilinear forms
  based on the resolvent of large random matrices,'' in \emph{Annales de
  l'Institut Henri Poincar{\'e}, Probabilit{\'e}s et Statistiques}, vol.~49,
  no.~1.\hskip 1em plus 0.5em minus 0.4em\relax Institut Henri Poincar{\'e},
  2013, pp. 36--63.

\bibitem{benaych2016spectral}
F.~Benaych-Georges and R.~Couillet, ``Spectral analysis of the gram matrix of
  mixture models,'' \emph{ESAIM: Probability and Statistics}, vol.~20, pp.
  217--237, 2016.

\end{thebibliography}

\end{document}